\documentclass{article}
\PassOptionsToPackage{dvipsnames,table}{xcolor}
\usepackage{fix-cm}
\usepackage{graphicx} 
\usepackage{booktabs}
\usepackage{wrapfig}
\usepackage{tikz}
\usetikzlibrary{arrows.meta,decorations.pathreplacing,patterns}
\usepackage{natbib}
\usepackage{algorithm}
\usepackage{algpseudocode}
\usepackage{tcolorbox}
\tcbuselibrary{breakable}
\usepackage{fvextra}
\usepackage{microtype}
\definecolor{highlightBlue}{HTML}{2362B3}
\newtcolorbox{highlightbox}{colback=highlightBlue!7, colframe=highlightBlue!65,
  boxrule=0.6pt, arc=2.5pt, left=5pt, right=5pt, top=3pt, bottom=3pt,
  before skip=6pt, after skip=8pt}
\usepackage{hyperref}
\let\cgRealAddContentsLine\addcontentsline
\usepackage{iclr2027_conference,times}
\let\addcontentsline\cgRealAddContentsLine
\hypersetup{
    final,
    colorlinks,
    linkcolor=ourred,
    citecolor=ourblue,
    urlcolor=url,
}
\usepackage{amsmath,amsfonts,amssymb,amsthm,bm,bbm}
\usepackage{graphicx}
\usepackage{caption}  

\newcommand{\method}{\text{Clock Diffusion}}

\newcommand{\x}{\mathbf{x}}

\newcommand{\z}{\mathbf{z}}

\newcommand{\e}{\mathbf{e}}

\newcommand{\cat}{\mathrm{Cat}}

\newcommand{\vocabsize}{V}
\newcommand{\ats}{\alpha_{t|s}}
\newcommand{\at}{\alpha_{t}}
\newcommand{\as}{\alpha_{s}}

\newcommand{\sts}{\sigma_{t|s}}
\newcommand{\st}{\sigma_{t}}
\renewcommand{\ss}{\sigma_{s}}

\newcommand{\snr}{\mathrm{SNR}}
\newcommand{\snrl}{\snr^\ell}
\newcommand{\qt}{q_t}
\newcommand{\qs }{q_s}
\newcommand{\qst}{q_{s|t}}
\newcommand{\qts}{q_{t|s}}

\def\lossdiffdisc{{\mathcal{L}^T_\text{diffusion}}}
\def\lossrecons{{\mathcal{L}_{\text{recons}}}}
\def\lossprior{{\mathcal{L}_\text{prior}}}

\newcommand{\reals}{\mathbb{R}}
\newcommand{\N}{\mathcal{N}}

\newcommand{\Ast}{\mathcal{A}_{s|t}}
\newcommand{\Zst}{\mathcal{Z}_{s|t}}
\newcommand{\Ct}{\mathcal{C}_{t}}
\newcommand{\Cs}{\mathcal{C}_{s}}
\newcommand{\Cst}{\mathcal{C}_{s|t}}
\newcommand{\Gv}{G_\varphi}

\usepackage[dvipsnames,table]{xcolor}
\definecolor{ourblue}{rgb}{0.368,0.507,0.71}
\definecolor{ourorange}{rgb}{0.881,0.611,0.142}
\definecolor{ourgreen}{rgb}{0.56,0.692,0.195}
\definecolor{ourred}{rgb}{0.923,0.386,0.209}
\definecolor{ourviolet}{rgb}{0.528,0.471,0.701}
\definecolor{ourbrown}{rgb}{0.772,0.432,0.102}
\definecolor{ourlightblue}{rgb}{0.364,0.619,0.782}
\definecolor{ourdarkgreen}{rgb}{0.572,0.586,0.}
\definecolor{url}{HTML}{d95225}
\definecolor{bloodred}{HTML}{B00000}
\definecolor{ourgray}{HTML}{36454F}

\definecolor{ourcyan2}{rgb}{0.125,0.722,0.804}
\definecolor{ourred2}{rgb}{0.863,0.184,0.047}
\definecolor{ouryellow2}{cmyk}{0,0.16,1.0,0.07}
\definecolor{ourviolet2}{cmyk}{0.55,0.56,0,0.47}
\definecolor{ourorange2}{cmyk}{0,0.46,0.89,0.11}

\definecolor{figblue}{HTML}{4169E1}
\definecolor{figgreen}{HTML}{33CD32}
\definecolor{figorange}{HTML}{FFA503}
\definecolor{figred}{HTML}{FF0000}
\definecolor{figgray}{HTML}{808080}

\def\eqref#1{equation~\ref{#1}}

\def\1{\bm{1}}

\DeclareMathAlphabet{\mathsfit}{\encodingdefault}{\sfdefault}{m}{sl}
\SetMathAlphabet{\mathsfit}{bold}{\encodingdefault}{\sfdefault}{bx}{n}

\def\gA{{\mathcal{A}}}

\def\gC{{\mathcal{C}}}

\def\gL{{\mathcal{L}}}
\def\gM{{\mathcal{M}}}
\def\gN{{\mathcal{N}}}

\def\gW{{\mathcal{W}}}

\newcommand{\E}{\mathbb{E}}

\newcommand{\softmax}{\mathrm{softmax}}

\newcommand{\KL}{\mathrm{KL}}

\DeclareMathOperator*{\argmax}{arg\,max}

\DeclareFontFamily{OT1}{pzc}{}
\DeclareFontShape{OT1}{pzc}{m}{it}{<-> s * [1.10] pzcmi7t}{}
\DeclareMathAlphabet{\mathpzc}{OT1}{pzc}{m}{it}

\newcommand{\xL}{\x^{1:L}}

\newcommand{\xCt}{\x^{\Ct}}
\newcommand{\xl}{\x^{\ell}}

\newcommand{\eL}{\e^{1:L}}
\newcommand{\el}{\e^{\ell}}
\newcommand{\zt}{\z_t}
\newcommand{\ztL}{\zt^{1:L}}
\newcommand{\ztl}{\zt^{\ell}}
\newcommand{\ztAst}{\zt^{\Ast}}
\newcommand{\zsL}{\zs^{1:L}}
\newcommand{\zsl}{\zs^{\ell}}
\newcommand{\zsAst}{\zs^{\Ast}}
\newcommand{\zti}{\z_{t(i)}}

\newcommand{\ztiL}{\zti^{1:L}}
\newcommand{\zs}{\z_s}
\newcommand{\zsi}{\z_{s(i)}}

\newcommand{\zsiL}{\zsi^{1:L}}
\newcommand{\zzero}{\z_0}
\newcommand{\zzeroL}{\zzero^{1:L}}

\newcommand{\zone}{\z_1}
\newcommand{\zoneL}{\zone^{1:L}}

\newcommand{\zzeroone}{\z_{0:1}}
\newcommand{\zzerooneL}{\zzeroone^{1:L}}

\newcommand{\ellp}{\ell^\prime}
\newcommand{\ul}{u^{\ell}}
\newcommand{\uls}{\ul(s)}
\newcommand{\ult}{\ul(t)}
\newcommand{\uAst}{u^{\Ast}}
\newcommand{\uAstt}{\uAst(t)}

\newcommand{\alt}{\at^\ell}
\newcommand{\slt}{\st^\ell}

\def\KV{\mathbf{KV}}

\author{Yair~Schiff$^{\,1,2,}$\hypersetup{linkcolor=black}\thanks{Corresponding author: \href{mailto:yschiff@nvidia.com}{\textcolor{black}{\textls[-50]\texttt{yschiff@nvidia.com}}}~/~\href{mailto:yzs2@cornell.edu}{\textcolor{black}{\textls[-50]\texttt{yzs2@cornell.edu}}}. Work done while at NVIDIA. \\ \hbox to 1.8em{\hss$^\dagger$}Joint senior authorship.}\hphantom{$^{*}$}, Omer~Belhasin$^{1}$, Roy~Uziel$^{1}$, Matan~Rusanovsky$^{1}$, Ran~Zilberstein$^{1}$, \\
\textbf{Marianne~Arriola$^{2}$, Gilad~Turok$^{2}$, Guanghan~Wang$^{2}$, Volodymyr~Kuleshov$^{2,\dagger}$, Michael~Elad$^{1,\dagger}$} \\[8pt]
\multicolumn{1}{c}{$^{1}$NVIDIA \qquad $^{2}$Cornell University}
}

\iclrpreprint 
\title{Clock Diffusion: Efficient \mbox{Semi-Autoregressive} Continuous Diffusion Language Models}
\usepackage{float}
\begin{document}
\maketitle
\ificlrfinal\vspace{-0.1in}\vspace{-2pt}\fi 

\begin{abstract}
Recent works on continuous diffusion for discrete data have demonstrated performance on par with comparable discrete diffusion models. However, these continuous counterparts lack key features that are essential to practical use as language models, namely variable-length generation and support for a key-value cache, and they still lag behind the frontier of autoregressive and discrete diffusion quality. In this work, we address these limitations. We do so by introducing a model parameterization that uses position-dependent noise schedules to define semi-autoregressive (SAR) continuous diffusion language models (DLMs). Together with efficient training and sampling algorithms, we call this framework \textbf{Clock Diffusion}, and we present two special cases of our method: block and sliding window generation. We then define \textbf{ClockDLMs}, a family of Gaussian DLMs based on sliding window Clock Diffusion that attain state-of-the-art diffusion likelihood bounds on OpenWebText, even beating the performant block SAR discrete diffusion models. ClockDLMs trained on TinyGSM also substantially outperform continuous baselines on the GSM8K benchmark and match and exceed comparable SAR discrete diffusion models. Finally, building on our parameterization, we propose more efficient samplers that we dub \textbf{Cache Grab}, which adapt techniques from accelerated inference in discrete diffusion, such as committing tokens whose probabilities exceed a confidence threshold and self-speculative decoding, further improving our models' quality and efficiency.
\end{abstract}

{\ificlrfinal\setlength{\intextsep}{8pt}\fi
\begin{figure}[h!]
    \ificlrfinal\vspace{-1.4em}\else\vspace{-1.8em}\fi
    \centering
    \input{figures/graphical_abstract}
    \ificlrfinal\vspace{0.0em}\else\vspace{0.3em}\fi
    \caption{
    (\textit{Left}) \textbf{Clock Diffusion} formulates semi-autoregressive continuous diffusion for discrete data, unlocking key drivers of practical LMs.
    (\textit{Right}) \textbf{ClockDLMs} achieve SoTA diffusion language modeling, beating block discrete diffusion and setting a new standard for continuous LMs.
    Our \textbf{Cache Grab} sampler further pushes the frontier, exceeding AR accuracy on GSM8K.
    }
    \label{fig:graphical_abstract}
\end{figure}}

\section{Introduction}\label{sec:intro}

Efforts to adapt diffusion models \citep{sohl2015deep, song2019generative, ho2020denoising, song2020score} to discrete domains, such as text, can be split into two streams of work.
The first relies on embedding the data in a continuous representation and then applying standard continuous diffusion on this space \citep{li2022diffusion, gulrajani2023likelihood}.
The second and historically more successful approach thus far has been to directly define the corruption and denoising processes on the discrete representations \citep{austin2021structured, campbell2022continuous, loudiscrete, sahoo2024simple, schiff2024simple}.
A recent resurgence of continuous models has challenged this paradigm \citep{chen2026langflow, davis2026scaling, lee2026flow, potaptchik2026discrete, roos2026categorical, yang2026continuous}, showing competitive performance with full sequence discrete diffusion.
However, these latest full-sequence continuous models cannot be used in practice, as they do not support variable-length generation with KV caching.
Moreover, their performance lags behind autoregressive (AR) language models (LMs) and semi-autoregressive (SAR) discrete diffusion.

In this work, we therefore present SAR continuous diffusion language models (DLMs),
unlocking arbitrary length generation and KV caching for continuous models while substantially improving quality.
We propose a framework consisting of variable noise schedules along the sequence \citep{ruhe2024rolling} and efficient training and sampling algorithms that we call \textbf{Clock Diffusion}, owing to our use of position-dependent local clocks that reparameterize the global time of the diffusion process.
We use our parameterization to formulate block and sliding window SAR generation.

Combining Gaussian diffusion over token embeddings with sliding window Clock Diffusion, we introduce ClockDLMs.
We train and evaluate these models on widely used language datasets.
Our model is the first continuous DLM to match block discrete diffusion on perplexity and sets a new state of the art among continuous models.

For the sliding window model, our sampling algorithm enables KV cache updates at every iteration.
Building on this, we propose accelerated inference methods that we dub \textbf{Cache Grab} samplers, which adapt techniques from discrete diffusion, such as committing tokens with probabilities above a confidence threshold \citep{wu2025fast} and self-speculative decoding \citep{fu2026nemotron}.

\begin{highlightbox}
\textbf{Contributions}~
In summary, our contributions are as follows:
    1) We introduce Clock Diffusion, a framework for variable-length generation and KV caching for continuous DLMs that consists of a probabilistic model and efficient training and sampling algorithms.
    2) We define ClockDLMs, a new family of Gaussian DLMs based on sliding window Clock Diffusion.
    3) We develop samplers, which accelerate inference with ClockDLMs by adapting successful methods from discrete diffusion.
    We name these Cache Grab samplers.
    4) ClockDLMs achieve SoTA among all continuous DLMs on language modeling (as measured by PPL on OWT), even beating discrete block DLMs.
    When combined with our Cache Grab samplers, we also achieve the highest GSM8K accuracy among comparably sized models trained on TinyGSM.
\end{highlightbox}

\section{Background}\label{sec:background}
\textbf{Diffusion Processes}~
Diffusion uses a forward corruption process that progressively corrupts samples $\x \sim q_{\text{data}}$ through a Markov chain of time-indexed latent variables $\zt$, for $t \in [0,1]$, so that the terminal distribution approximates a tractable prior $q_{\text{prior}}$.
The \textit{forward} corruption process $\qts(\zt \mid \zs)$, where for  $s < t$, latent variables become progressively noisier, can be applied recursively to yield tractable noising kernels $\qt(\zt \mid \x)$, which in turn can be used to form tractable \textit{reverse} posteriors $\qst(\zs \mid \zt, \x) = \qts(\zt \mid \zs)\tfrac{\qs(\zs \mid \x)}{\qt(\zt \mid \x)}$.
Diffusion generative models are defined by learned reverse distributions, which are commonly parameterized as $p_\theta(\zs \mid \zt) := \qst(\zs \mid \zt, \x = \x_\theta)$, where $\x_\theta$ is a neural network prediction of the clean signal $\x$ given the noisy input $\zt$.
For multivariate data $\x, \zt \in \reals^d$, both the forward and reverse processes are applied independently across dimensions.
For sequence data, we model $\xL,  \ztL \in \reals^{L \times d}$, where superscripts denote the sequence length dimension, and we apply the diffusion processes independently along this dimension as well.

\textbf{Continuous Diffusion for Text}~
To enable Gaussian corruption for discrete data, we embed tokens into a continuous space where the diffusion process is applied.
Let $\xl \in \{0, 1\}^\vocabsize \cap \Delta^\vocabsize$ denote a one-hot vector, where $\vocabsize$ is the vocabulary size and $\Delta^\vocabsize$ is the simplex over the vocabulary.
An embedder $E: \Delta^\vocabsize \rightarrow \reals^d$ produces embeddings $\el := E[\xl] \in \reals^{d}$.
We denote a sequence of $L$ embeddings as $\eL := E[\xL] \in \reals^{L \times d}$.
The diffusion process is then defined as follows:
\begin{align}
    q(\ztl \mid \xl) = \N(\at \el, \st^2\mathbf{I}) \quad \text{and} \quad q(\zsl \mid \ztl, \xl) = \gN\left( \tfrac{\ats\ss^2}{\st^2}\ztl + \tfrac{\as\sts^2}{\st^2}\el, \tfrac{\sts^2\ss^2}{\st^2}\mathbf{I} \right),
\end{align}
where $\ats := \at / \as, \sts^2 := \st^2 - \ats^2\ss^2$, and $\at := \alpha(t)$ and $\st := \sigma(t)$ are monotonically decreasing and increasing functions in $t$, respectively, that dictate the amount of signal and noise, respectively, present in a given latent variable.
For the learned reverse process, often, we let $\xl_\theta \in \Delta^\vocabsize$ be the softmax output of a neural network that predicts logits over the vocabulary at position $\ell$ and our parameterized prediction of the clean token embeddings is $\eL_\theta := E[\xL_\theta]$.

Previous continuous LMs can be categorized by the embedder $E$ that they use, with some notable examples being: \textit{One-hot embedding}: $E := \mathrm{Id}: \Delta^\vocabsize \rightarrow \reals^\vocabsize$ \citep{lee2026flow, potaptchik2026discrete, roos2026categorical}, \textit{Token embeddings}: $E$ is the embedding matrix $\in \reals^{\vocabsize \times d}$ \citep{li2022diffusion, dieleman2022continuous, gulrajani2023likelihood, chen2026langflow, yang2026continuous}, and \textit{Contextual embeddings}: $E$ is a pre-trained encoder \citep{hu2026elf}.

\textbf{Semi-Autoregressive Discrete Diffusion}~
In the context of discrete diffusion, \citet{arriola2025block} decompose the distribution over sequences into sequential blocks and perform diffusion over each block.
This SAR approach tightens the likelihood bound, improves sample quality, and most notably unlocks KV cache sampling when using Transformer architectures \citep{vaswani2017attention}.

\section{\method}\label{sec:clock}
In this section, we present our framework, which consists of a position-dependent noising parameterization (Section \ref{subsec:clock_param}), a KV cache sampling algorithm (Section \ref{subsec:clock_sampling}), and efficient training (Section \ref{subsec:clock_training}).
In Section \ref{subsec:special_clocks}, we provide block and sliding window instantiations of this framework, and in Section \ref{subsec:clock_dlms}, we define ClockDLMs, a family of Gaussian DLMs built on our sliding window parameterization.
Finally, in Section \ref{subsec:sampling_acceleration}, we develop efficient samplers for ClockDLMs. 

\subsection{Parameterization}\label{subsec:clock_param}
\textbf{Forward Process}~
We build on previous works that proposed position-dependent noise schedules in the context of continuous data \citep{chen2024diffusion, ruhe2024rolling}, such as video frame prediction or robotic motion planning, and adapt these methods to language.
Starting from a continuous diffusion for discrete data noising process, as described in Section \ref{sec:background}, the key insight is that we can have each position $\ell \in \{1, \ldots, L\}$ convert global diffusion time $t$ via a local clock $\ul: [0, 1] \rightarrow [0, 1]$, which is non-decreasing and must satisfy $\ul(0) = 0$ and $\ul(1) = 1$.
This defines per-position noising schedules: $\qts(\ztL \mid \zsL) = \prod_{\ell=1}^L q_{\ult|\uls}(\ztl \mid \zsl)$.
Importantly, we also impose that for $\ult = \uls = u$, i.e., for a global diffusion time progression that leads to no change in a token's local clock, the noising kernel is the Dirac measure:
\begin{align}\label{eq:no_change_fwd}
    q_{u|u}(\ztl \mid \zsl) := \delta(\ztl - \zsl) \rightarrow \text{\textit{positions where local time has not advanced remain fixed}}.
\end{align}

\textbf{Reverse Process}~
In the denoising direction, using local clocks, we create two non-overlapping sets:
\begin{align}
\text{`}\textit{awake}\text{'}~~\Ast := \{\ell \mid \uls < \ult\}, \quad\text{`}\textit{snoozed}\text{'}~~\Zst := \{\ell \mid \uls = \ult\} = (\Ast)^c.
\end{align}
\textit{Awake} positions are actively being denoised, as they undergo a local clock change.
\textit{Snoozed} positions correspond to latent variable components that can be safely ignored, as they consist of tokens that either have not begun ($\uls = \ult = 1$) or have completed denoising ($\uls = \ult = 0$).

The kernel from (\ref{eq:no_change_fwd}) lets us derive a Dirac measure for the posterior of tokens in the \textit{snoozed} set:
\begin{align}\label{eq:clock_true_joint}
    q_{\uls|\ult}(\zsl \mid \ztl, \xl) = q_{u|u}(\ztl \mid \zsl)\tfrac{q_u(\zsl \mid \xl)}{q_u(\ztl \mid \xl)} = \delta(\ztl - \zsl), \quad \text{if } \ell \in \Zst,
\end{align}
where in the last equality we used the fact that for any function we have $\delta(\ztl - \zsl)f(\zsl) = \delta(\ztl - \zsl)f(\ztl)$ applied to $f(\zsl) = q_u(\zsl \mid \xl) / q_u(\ztl \mid \xl)$.
We can then characterize the full reverse posterior as factorizing over awake and snoozed positions:
\begin{align}\label{eq:clock_rev}
    \qst(\zsL \mid \ztL, \xL) &= \textstyle\prod_{\ell \in \Ast} q_{\uls \mid \ult}(\zsl \mid \ztl, \xl) \textstyle\prod_{\ell \in \Zst} \delta(\ztl - \zsl).
\end{align}

\textbf{Generative Model}~
For the learned process, define the set of completed positions at global time $t$ as $\Ct := \{\ell \mid \ult = 0\}$ and the newly completed positions for times $s < t$ as $\Cst := \Cs \setminus \Ct$.
Let $\zzerooneL := (\z_{t(i)}^{1:L})_{i=0}^{T-1}$ denote the latent trajectory.
Then we posit a generative model whose joint distribution over data and latent variable sequences factorizes as follows:
\begin{align}\label{eq:clock_learned_joint}
\begin{split}
    p{_\theta}&(\xL, \zzerooneL) = p_\theta(\zoneL) \textstyle\prod_{i=1}^{T-1} p_\theta\!\left( \zsL,\x^{\Cst} \mid \ztL, \xCt \right) \\
    &:= p_\theta(\zoneL) \textstyle\prod_{i=1}^{T-1} p_\theta\!\left(\zsL \mid \ztL, \xCt \right)p_\theta\!\left(\x^{\Cst} \mid \zsL, \xCt \right).
\end{split}
\end{align}
This factorization dictates that at every timestep, we model latent sequence denoising $\ztL \to \zsL$ and then account for updates to the clean sequence given these changes to the denoised latent.
We define the latent-transition factor in (\ref{eq:clock_learned_joint}) so that it factorizes over snoozed and awake positions:
\begin{align}
    p_\theta\!\left(\zsL \mid \ztL, \xCt \right) := p_\theta(\zs^{\Zst} \mid \zt^{\Zst})p_\theta(\zs^{\Ast} \mid \zt^{\Ast}, \xCt).
\end{align}
Following the true reverse posterior, we can ignore snoozed positions by defining for any $\ell \in \Zst$: $p_\theta(\zsl \mid \ztl) := \delta(\ztl - \zsl)$.
We then need only account for the awake positions, which depend on their previous state at time $t$ and any already revealed tokens $\x^{\Ct}$.

Finally, to model updates to the data sequence, we output distributions over the clean data given the updated awake positions and already revealed tokens:
\begin{align}
    p_\theta(\x^{\Cst} \mid \zsL, \xCt) := p_\theta(\x^{\Cst} \mid \zs^{\Ast}, \xCt) := \textstyle\prod_{\ell \in \Cst} \cat(\xl; \xl_\theta(\zs^{\Ast}, \xCt)), 
\end{align}
where $\cat(\cdot ;p)$ denotes a categorical distribution with probability vector $p \in \Delta^\vocabsize$.

\textbf{Semi-Autoregressive Local Clocks}~
In our work, we study local clocks for which every position runs the same schedule but is offset by a per-position \emph{phase} $\varphi(\ell)$:
\begin{align}\label{eq:phase}
    \ult := \operatorname{clip}_{[0,1]}\big(r\,t - \varphi(\ell)\big), \qquad r := \max_{\ellp \in \{1, \ldots, L\}}\varphi(\ellp) + 1,
\end{align}
where $\varphi: \{1, \ldots, L\} \rightarrow \reals_{\geq 0}$ and $\operatorname{clip}_{[0,1]}(\cdot)$ clips its argument to $[0,1]$.
As shown in Section \ref{subsec:special_clocks}, different choices of $\varphi$ lead to different instantiations of Clock Diffusion.
Intuitively, $\varphi(\ell)$ is the `\emph{head start}' of position $\ell$.
Each position experiences rescaled time $t \rightarrow r\,t$ and is awake precisely for all $t \in [0, 1]$ such that $\varphi(\ell) < r\,t < \varphi(\ell) + 1$, i.e., every position denoises over a window that lasts $1/r$ units of global diffusion time, and $\varphi(\ell)$ sets where that window falls.
This parameterization induces SAR decoding: positions with a larger phase finish denoising earlier.

Since all token windows are of equal duration, for any two positions $\ell, \ellp$ the offset between their windows is determined by
$\Gv(\ell, \ellp) := \varphi(\ellp) - \varphi(\ell).$
When $\Gv \geq 1$, windows are disjoint, so $\ellp$ finishes denoising before or just as $\ell$ begins; when $\Gv = 0$, the two positions denoise in lockstep; and when $0 < \Gv < 1$, they overlap, so $\ellp$ leads $\ell$ but the two are denoised together for part of their trajectories.
Negative $\Gv$ gives the same three cases with the roles of $\ell$ and $\ellp$ exchanged.

\textbf{Variational Bound}~
Clock Diffusion admits a principled likelihood bound (see Appendix \ref{appsec:clock_nelbo}):
\begin{align}\label{eq:clock_nelbo_main}
    &\gL(\theta) := \underbrace{\KL[q(\zoneL \mid \xL) || p_\theta(\zoneL)]}_{\lossprior} + \underbrace{\textstyle\sum_{i=1}^{T-1}\E_{q(\zs^{\Ast} \mid \xL)} \textstyle\sum_{\ell \in \Cst} -\log p_\theta(\xl \mid \zsAst, \xCt)}_{\gL_{\text{recons}}} \nonumber \\
    &~~~~~~~~~~~~~~~~+ \underbrace{\textstyle\sum_{i=1}^{T-1}\E_{q(\ztL \mid \xL)} \textstyle\sum_{\ell \in \Ast} \KL[q(\zsl \mid \ztl, \xl) || p_\theta(\zsl \mid \ztAst, \xCt)]}_{\gL_{\text{diffusion}}}.
\end{align}

\subsection{Key-Value Cache Sampling}\label{subsec:clock_sampling}
At any global time $t$, the local clocks split the sequence into disjoint sets: positions in $\Ct$ have finished denoising, awake positions are being actively denoised, and the rest have not yet begun.
Positions in $\Ct$ never change again and therefore their KV entries are written once, when the position completes, and reused unchanged thereafter.
Every awake clock advances at rate $r$, so evaluating $t$ at $T$ equally spaced values moves each awake clock by $r/T$ per model call.
Giving every position a budget of $N$ reverse updates therefore fixes the number of model calls at $T := r\,N$.

\begin{wrapfigure}{r}{0.56\textwidth}
\vspace{-2.1em}
\begin{minipage}[t]{0.56\textwidth}
    \begin{algorithm}[H]
        \small
        \caption{\method\ Sampling}
        \label{alg:clock_sampling}
        \begin{algorithmic}[1]
        \Require prompt, length $L$, per-position budget $N$, clocks $\ul$, prior $q_{\mathrm{prior}}$, \textsc{Denoiser}, sampling routine $\textsc{Sampler}$
        \State $T \gets rN$ \Comment{\texttt{Total budget}}
        \State $\KV\gets\textsc{Cache}(\text{prompt},\emptyset)$
        \State $\zoneL \sim q_{\mathrm{prior}}$
        \For{$t = 1, \tfrac{T-1}{T}, \dots, \tfrac{1}{T}$}
            \State $s \gets t - \tfrac{1}{T}$
            \State $\Ast \gets \{\ell \mid \uls < \ult \}$
            \State $\textsc{logits}^{\Ast}\gets\textsc{Denoiser}(\ztAst,\uAstt,\KV)$
            \State $\zsAst\gets\textsc{Sampler}(\textsc{logits}^{\Ast})$
            \State $\Cst \gets\{\ell\in\Ast \mid \uls=0\}$ 
            \State $\x^{\Cst}\gets\argmax\textsc{logits}^{\Cst}$
            \State $\KV\gets\textsc{Cache}(\x^{\Cst},\KV)$
        \EndFor
        \State \Return $\xL$
        \end{algorithmic}
    \end{algorithm}
\end{minipage}
\vspace{-3.8em}
\end{wrapfigure}
Algorithm \ref{alg:clock_sampling} is the resulting sampler.
At each step, we advance the entire set $\Ast$ with a single call to $\textsc{Denoiser}(\ztAst,\uAstt,\KV)$.
Positions whose clocks just reached zero are collected into $\Cst$, decoded via $\argmax$, and appended to the cache with a clean-token model forward, written as $\textsc{Cache}(\x^{\Cst},\KV)$.
In practice, we avoid the separate model forward $\textsc{Cache}$ call by fusing this with the subsequent denoiser step, requiring a custom attention mask, depicted in Appendix \ref{appsec:sampling}.

\subsection{Efficient Training}\label{subsec:clock_training}
The diffusion loss term in (\ref{eq:clock_nelbo_main}) introduces a complexity for training: since the conditioning set varies by position, we cannot naively model a reverse diffusion step at each position with a single function evaluation.
To account for this, \cite{arriola2025block} propose a vectorized implementation where we train with sequences of length $2L$: the first part of the sequence corresponds to clean tokens $\xL$ and the second part to noisy tokens $\ztL$.
For each sample, we partition the noisy tokens into disjoint groups; each group represents a separate awake set.
The attention mask $\gM \in \{0, 1\}^{2L \times 2L}$ can be broken into four $L\times L$ sub-masks:
$
    \gM =
    \begin{pmatrix}
        \gM_{\textsc{cc}} & \mathbf{0}\\
        \gM_{\textsc{nc}} & \gM_{\textsc{nn}}
    \end{pmatrix}
$.

\textbf{Clean-to-clean}~ $\gM_{\textsc{cc}}[\ell, \ellp] := \1\!\left[\Gv(\ell, \ellp) \geq 0\right]$.
A decoded token's KV entry is written once and reused for the remainder of generation, so it must be computable from what was available at the moment it was written; permitting $\Gv < 0$ would allow dependence on tokens that did not yet exist.

\textbf{Noisy-to-clean}~ For a reverse step from $t$ to $s$, $\gM_{\textsc{nc}}[\ell,\ellp] := \1\![\ellp\in\Ct]$ for $\ell\in\Ast$.
In our vectorized block and sliding window training layouts, each noisy group attends to the clean positions strictly preceding that group.

\begin{wrapfigure}{r}{0.45\textwidth}
    \vspace{-2.3em}
    \centering
%
%
%
%
%

\begingroup
\definecolor{cgMaskBlue}{HTML}{A8C6EE}
\definecolor{cgMaskGreen}{HTML}{BBDDA8}
\definecolor{cgMaskOrange}{HTML}{F7C89B}
\definecolor{cgMaskCut}{HTML}{C76835}
\definecolor{cgMaskEmpty}{HTML}{ECECEC}
\definecolor{cgMaskRule}{HTML}{55606E}
\definecolor{cgMaskInk}{HTML}{25232A}
\definecolor{cgMaskMuted}{HTML}{667085}

\newcommand{\cgL}{6}
\newcommand{\cgb}{2}

\begin{tikzpicture}[
  x=0.30cm,
  y=0.30cm,
  font=\small,
  mask label/.style={font=\tiny, text=cgMaskInk, inner sep=0.8pt},
  axis label/.style={font=\tiny\itshape, text=cgMaskMuted, inner sep=1.0pt},
  legend text/.style={font=\tiny, text=cgMaskInk, inner sep=0pt, align=left},
  cut outline/.style={draw=cgMaskCut, densely dashed, line width=0.9pt},
]

\pgfmathtruncatemacro{\cgN}{2*\cgL}

\foreach \cgq in {1,...,\cgN}{%
  \foreach \cgk in {1,...,\cgN}{%
    \pgfmathtruncatemacro{\cgnq}{\cgq>\cgL ? 1 : 0}%
    \pgfmathtruncatemacro{\cgnk}{\cgk>\cgL ? 1 : 0}%
    \pgfmathtruncatemacro{\cgpq}{\cgq-\cgnq*\cgL}%
    \pgfmathtruncatemacro{\cgpk}{\cgk-\cgnk*\cgL}%
    \pgfmathtruncatemacro{\cgbq}{floor((\cgpq-1)/\cgb)}%
    \pgfmathtruncatemacro{\cgbk}{floor((\cgpk-1)/\cgb)}%
    \pgfmathtruncatemacro{\cgcc}{(\cgnq==0 && \cgnk==0 && \cgbk<=\cgbq) ? 1 : 0}%
    \pgfmathtruncatemacro{\cgnc}{(\cgnq==1 && \cgnk==0 && \cgbk<\cgbq) ? 1 : 0}%
    \pgfmathtruncatemacro{\cgnn}{(\cgnq==1 && \cgnk==1 && \cgbk==\cgbq) ? 1 : 0}%
    \pgfmathtruncatemacro{\cgst}%
      {\cgcc==1 ? 1 : (\cgnc==1 ? 2 : (\cgnn==1 ? 3 : 0))}%
    \ifcase\cgst\relax
      \colorlet{cgMaskCur}{cgMaskEmpty}%
    \or
      \colorlet{cgMaskCur}{cgMaskBlue}%
    \or
      \colorlet{cgMaskCur}{cgMaskGreen}%
    \or
      \colorlet{cgMaskCur}{cgMaskOrange}%
    \fi
    \path[fill=cgMaskCur, draw=white, line width=0.35pt]
      ({\cgk-1},{-\cgq+1}) rectangle ++(1,-1);%
  }%
}

\draw[draw=cgMaskRule, line width=0.7pt] (\cgL,0) -- (\cgL,-\cgN);
\draw[draw=cgMaskRule, line width=0.7pt] (0,-\cgL) -- (\cgN,-\cgL);
\draw[draw=cgMaskRule, line width=0.7pt] (0,0) rectangle (\cgN,-\cgN);

\foreach \cgq in {1,...,\cgL}{%
  \foreach \cgk in {1,...,\cgL}{%
    \pgfmathtruncatemacro{\cgbq}{floor((\cgq-1)/\cgb)}%
    \pgfmathtruncatemacro{\cgbk}{floor((\cgk-1)/\cgb)}%
    \ifnum\cgbk=\cgbq
      \ifnum\cgk>\cgq
        \draw[cut outline] ({\cgk-1},{-\cgq+1}) rectangle ++(1,-1);%
      \fi
    \fi
  }%
}

\foreach \cgi in {1,...,\cgL}{%
  \node[mask label, anchor=east] at (-0.2,{-\cgi+0.5}) {$\mathbf{x}^{\cgi}$};
  \node[mask label, anchor=east]
    at (-0.2,{-\cgi-\cgL+0.5}) {$\mathbf{z}_{t}^{\cgi}$};
  \node[mask label, anchor=north] at ({\cgi-0.5},-\cgN-0.1) {$\mathbf{x}^{\cgi}$};
  \node[mask label, anchor=north]
    at ({\cgi+\cgL-0.5},-\cgN-0.1) {$\mathbf{z}_{t}^{\cgi}$};
}
\node[axis label, rotate=90] at (-2.2,{-\cgN/2}) {Queries};
\node[axis label] at ({\cgN/2},-13.8) {Keys};

\def\cgpitch{2.2}
\newcommand{\cgmasklegend}[3]{%
  \path[fill=#2, draw=cgMaskRule, line width=0.3pt]
    (12.7,{-0.8-#1*\cgpitch}) rectangle ++(1,-1);
  \node[legend text, anchor=west] at (14.1,{-1.3-#1*\cgpitch}) {#3};
}

\cgmasklegend{0}{cgMaskBlue}{Clean-to-clean\\(block) causal}
\cgmasklegend{1}{cgMaskGreen}{Noisy-to-clean\\block causal}
\cgmasklegend{2}{cgMaskOrange}{Noisy-to-noisy\\block diagonal}
\cgmasklegend{3}{cgMaskBlue}{Removed for\\Sliding Window}
\draw[cut outline] (12.7,{-0.8-3*\cgpitch}) rectangle ++(1,-1);
\node[axis label, anchor=north west] at (13,{-2.7-3*\cgpitch}) {$L = 6,~~S=2$};

\end{tikzpicture}
\endgroup
    \vspace{-1.5em}
    \caption{Clock Diffusion training masks.
    \textit{Block}: clean-to-clean is block causal, a noisy query attends to clean keys strictly preceding its own block, and noisy queries attend bidirectionally within a block.
    \textit{Sliding Window}: we remove the outlined clean-to-clean entries, making that quadrant causal.
    }
    \label{fig:attention_masks}
\vspace{-4.5em}
\end{wrapfigure}
\textbf{Noisy-to-noisy}~ $\gM_{\textsc{nn}}[\ell, \ellp] := 1$ if $\ell, \ellp$ belong to the same denoising group, and $0$ otherwise.

\subsection{Special Cases of Clock Diffusion}\label{subsec:special_clocks}
\textbf{Full Sequence}~
We first note that simply letting $\varphi(\ell) = 0 $, for all $\ell$, will yield $\ult = t$ at every position, which recovers full sequence DLMs.

\textbf{Block}~
Clock Diffusion can also admit a continuous version of the block discrete diffusion method proposed in \citet{arriola2025block}.
In Appendix \ref{appsubsec:block_clock_dlms}, we show how a specific choice of per-position phase defines local clocks, training, and sampling that correspond to block continuous diffusion.

\textbf{Sliding Window}~
Block diffusion is limited by the fact that we must wait for an entire block to complete denoising before updating the KV cache.
To mitigate this, we propose Sliding Window Clock Diffusion, where we maintain an active window of at most $S$ tokens that slides whenever the leftmost position transitions to $\ult = 0$.
We achieve this with the following per-position phase:
\begin{align}\label{eq:window_clock}
    \varphi_{\textsc{SW}}(\ell) := \frac{L - \ell}{S} \implies \ul_{\textsc{SW}}(t) := \operatorname{clip}_{[0,1]}\left(\frac{(L+S-1)t+\ell-L}{S}\right).
\end{align}
The phase gap now measures the distance between two positions in units of the window width: $\Gv(\ell, \ellp) = \tfrac{L-\ellp}{S} - \tfrac{L-\ell}{S} = \tfrac{\ell - \ellp}{S}.$
This translates to a straightforward modification of the block diffusion mask: $\gM_{\textsc{cc}}$ reduces to pure causal attention.
During training, we partition noisy positions into disjoint groups of $S$ tokens, and each group attends to its own noisy
positions and its preceding clean prefix (see Figure \ref{fig:attention_masks}).
In Appendix \ref{appsubsec:training_sw}, we show how to compute an unbiased estimate of the variational bound, which we derive in Appendix \ref{appsubsec:sw_clock_dlms}.
Training with causal attention on clean tokens allows us, at generation time, to commit and cache any leftmost prefix of an active window without waiting for the rest of the window to complete denoising, since we can always commit and cache these tokens with the causal AR mask.
We leverage this fact in Section \ref{subsec:sampling_acceleration}, below.

\subsection{Clock Diffusion Language Models}\label{subsec:clock_dlms}
Building on our framework, we combine \textit{Sliding Window Clock Diffusion} with \textit{Gaussian noise on token embeddings}, i.e., we use a learnable embedding matrix $E \in \reals^{\vocabsize \times d}$ as the embedder and perform diffusion over clean embeddings $\e := E^\top\x$, and we parameterize our denoising model as $\e_\theta := E^\top\x_\theta$, for $\x_\theta \in \Delta^\vocabsize$.
We also adopt several of the training techniques, such as learned noise schedules, that were used in a recent successful full sequence continuous DLM from \citet{yang2026continuous}.
Finally, we optimize a continuous-time formulation of the variational Sliding Window Clock Diffusion objective from (\ref{eq:clock_nelbo_main}).
Together these elements form a new continuous DLM family that we call Clock Diffusion Language Models (\textbf{ClockDLMs}).
As an ablation, below, we also train a block version, B-ClockDLMs.
See Appendices \ref{appsec:clock_dlms} and \ref{appsec:training} for full details.

\begin{figure}[t]
    \centering
    \begin{minipage}[t]{0.48\textwidth}
        \vspace{0pt}
        \centering
%
%
%
%
\begingroup%
\definecolor{ecCache}{HTML}{3F8A5C}%
\definecolor{ecWin}{HTML}{2362B3}%
\definecolor{ecGhost}{HTML}{667085}%
\definecolor{ecThresh}{HTML}{728FB6}%
\definecolor{ecInk}{HTML}{25232A}%
\def\ecP{0.92}
\def\ecBW{0.185}
\def\ecT{0.95}
%
\newcommand{\ecpos}[6]{%
 \pgfmathsetmacro{\ecx}{\ecP*(#1-1)}%
 \ifnum#6=1\colorlet{echi}{ecCache}\else\colorlet{echi}{ecWin}\fi%
 \foreach \eck/\ech in {1/#2,2/#3,3/#4}{%
   \pgfmathsetmacro{\ecbx}{\ecx+0.11+0.235*(\eck-1)}%
   \ifnum\eck=#5%
     \ifnum#6=1\def\ecfill{echi!85}\else\def\ecfill{echi!22}\fi%
     \path[fill=\ecfill,draw=echi,line width=0.7pt]%
       (\ecbx,0) rectangle ++(\ecBW,\ech);%
   \else%
     \path[fill=ecGhost!18,draw=ecGhost!55,line width=0.4pt]%
       (\ecbx,0) rectangle ++(\ecBW,\ech);%
   \fi}}%
%
\newcommand{\eccell}[3]{%
 \pgfmathsetmacro{\ecx}{\ecP*(#1-1)}%
 \ifnum#3=1\colorlet{ecc}{ecCache}\else\colorlet{ecc}{ecGhost}\fi%
 \path[fill=ecc!8,draw=ecc!60,line width=0.5pt,rounded corners=1.5pt]%
   (\ecx+0.02,-0.52) rectangle ++(0.84,0.40);%
 \node[font=\scriptsize,text=ecInk,inner sep=0] at (\ecx+0.44,-0.325) {#2};}%
\begin{tikzpicture}[x=1cm,y=1cm,>=Stealth]
\draw[ecGhost!70,line width=0.5pt,->] (-0.12,0) -- (-0.12,1.52);
\draw[ecGhost!70,line width=0.5pt] (-0.12,0) -- (5.60,0);
\node[font=\scriptsize,text=ecGhost!85!black,rotate=90,inner sep=0]
  at (-0.42,0.76) {softmax};

\ecpos{1}{1.28}{0.45}{0.60}{1}{1}
\ecpos{2}{0.40}{0.52}{1.22}{3}{1}
\ecpos{3}{1.18}{0.42}{0.55}{1}{1}
\ecpos{4}{0.78}{0.62}{0.70}{1}{0}
\ecpos{5}{0.60}{0.72}{0.55}{2}{0}
\ecpos{6}{1.15}{0.40}{0.50}{1}{0}

\draw[ecThresh,densely dashed,line width=0.9pt] (-0.12,\ecT) -- (5.56,\ecT);

\eccell{1}{the}{1}\eccell{2}{cat}{1}\eccell{3}{sat}{1}
\eccell{4}{by}{0}\eccell{5}{the}{0}\eccell{6}{mat}{0}

\draw[ecCache,densely dashed,line width=0.9pt,rounded corners=2.5pt]
  (-0.04,-0.58) rectangle ({\ecP*2+0.90},-0.06);
\node[font=\scriptsize,text=ecCache,anchor=west,inner sep=0]
  at (0.0,-0.76) {grabbed early};

\draw[ecInk!75,line width=0.7pt,->]
  ({\ecP*3+0.44},1.44) -- ({\ecP*3+0.44},0.86);
\node[font=\scriptsize,text=ecInk!75,anchor=south,inner sep=1.5pt]
  at ({\ecP*3+0.44},1.46) {\textit{first below confidence threshold: run stops}};

\end{tikzpicture}%
\endgroup%
    \end{minipage}\hfill
    \begin{minipage}[t]{0.48\textwidth}
        \vspace{0pt}
        \centering
%
%
%
%
\begingroup%
\definecolor{ssCache}{HTML}{3F8A5C}%
\definecolor{ssWin}{HTML}{2362B3}%
\definecolor{ssGhost}{HTML}{667085}%
\definecolor{ssRej}{HTML}{C76835}%
\definecolor{ssInk}{HTML}{25232A}%
\def\ssP{0.92}
\def\ssW{0.84}
\def\ssH{0.40}
%
\newcommand{\sscell}[4]{%
 \pgfmathsetmacro{\ssx}{\ssP*(#1-1)}%
 \path[fill=#4!8,draw=#4!60,line width=0.5pt,rounded corners=1.5pt]%
   (\ssx+0.02,#2) rectangle ++(\ssW,\ssH);%
 \node[font=\scriptsize,text=ssInk,inner sep=0]%
   at (\ssx+0.02+0.5*\ssW,#2+0.5*\ssH) {#3};}%
%
\newcommand{\sscorr}[3]{%
 \pgfmathsetmacro{\ssx}{\ssP*(#1-1)}%
 \path[fill=ssCache!8,draw=ssCache!60,line width=0.5pt,rounded corners=1.5pt]%
   (\ssx+0.02,#2) rectangle ++(\ssW,\ssH);%
 \node[font=\scriptsize,text=ssInk,inner sep=0]%
   at (\ssx+0.02+0.5*\ssW,#2+0.5*\ssH) {#3};}%
%
\newcommand{\sstok}[4]{%
 \pgfmathsetmacro{\ssx}{\ssP*(#1-1)+0.02}%
 \pgfmathsetmacro{\ssu}{#4/6}%
 \path[fill=ssWin!13] (\ssx,#2) rectangle ++(\ssW,\ssH);%
 \pgfmathsetseed{\numexpr 1013+617*#1\relax}%
 \pgfmathsetmacro{\sssw}{\ssW/12}%
 \pgfmathsetmacro{\sssh}{\ssH/6}%
 \begin{scope}%
 \clip (\ssx,#2) rectangle ++(\ssW,\ssH);%
 \foreach \si in {0,...,11}{\foreach \sj in {0,...,5}{%
   \pgfmathsetmacro{\ssg}{(rand+rand+rand)*0.62}%
   \pgfmathsetmacro{\sso}{min(0.92,abs(\ssg)*\ssu)}%
   \pgfmathtruncatemacro{\ssup}{\ssg>0 ? 1 : 0}%
   \ifnum\ssup=1%
     \path[fill=ssWin!88!black,opacity=\sso]%
       (\ssx+\sssw*\si,#2+\sssh*\sj) rectangle ++(\sssw,\sssh);%
   \else%
     \path[fill=white,opacity=\sso]%
       (\ssx+\sssw*\si,#2+\sssh*\sj) rectangle ++(\sssw,\sssh);%
   \fi}}%
 \end{scope}%
 \pgfmathsetmacro{\ssfo}{#4<2 ? 0 : 0.60}%
 \node[font=\scriptsize,text=ssInk,inner sep=0.7pt,rounded corners=0.5pt,%
       fill=white,fill opacity=\ssfo]%
   at (\ssx+0.5*\ssW,#2+0.5*\ssH) {#3};%
 \draw[ssWin!55,line width=0.35pt,rounded corners=1.5pt]%
   (\ssx,#2) rectangle ++(\ssW,\ssH);}%
\begin{tikzpicture}[x=1cm,y=1cm,>=Stealth]
\def\sstop{1.16}\def\ssmid{0.44}\def\ssbot{-0.26}
\def\sslab{-0.10}
\node[font=\scriptsize,text=ssWin,anchor=east,align=right,inner sep=0]
  at (\sslab,\sstop+0.5*\ssH) {diffusion\\draft};
\sscell{1}{\sstop}{the}{ssWin}\sscell{2}{\sstop}{cat}{ssWin}
\sscell{3}{\sstop}{sat}{ssWin}\sscell{4}{\sstop}{by}{ssWin}
\sscell{5}{\sstop}{the}{ssWin}\sscell{6}{\sstop}{mat}{ssWin}
\node[font=\scriptsize,text=ssGhost!85!black,anchor=east,align=right,inner sep=0]
  at (\sslab,\ssmid+0.5*\ssH) {AR\\verify};
\sscell{1}{\ssmid}{the}{ssGhost}\sscell{2}{\ssmid}{cat}{ssGhost}
\sscell{3}{\ssmid}{sat}{ssGhost}\sscell{4}{\ssmid}{on}{ssGhost}
\pgfmathsetmacro{\ssgap}{0.5*(\sstop+\ssmid+\ssH)}
\foreach \sm in {1,2,3}{\node[font=\tiny,text=ssCache,inner sep=0]
  at ({\ssP*(\sm-1)+0.44},\ssgap) {$\checkmark$};}
\node[font=\scriptsize,text=ssRej,inner sep=0] at ({\ssP*3+0.44},\ssgap) {$\bm{\times}$};
\node[font=\scriptsize,text=ssInk!75,anchor=east,align=right,inner sep=0]
  at (\sslab,\ssbot+0.5*\ssH) {after};
\sscell{1}{\ssbot}{the}{ssCache}\sscell{2}{\ssbot}{cat}{ssCache}
\sscell{3}{\ssbot}{sat}{ssCache}\sscorr{4}{\ssbot}{on}
\sstok{5}{\ssbot}{the}{3}\sstok{6}{\ssbot}{mat}{4}
\draw[ssCache!75,line width=0.6pt,->,densely dotted]
  ({\ssP*3+0.44},\ssmid-0.03) -- ({\ssP*3+0.44},\ssbot+\ssH+0.05);
\draw[ssCache,densely dashed,line width=0.9pt,rounded corners=2.5pt]
  (-0.02,\ssbot-0.05) rectangle ({\ssP*3+0.88},\ssbot+\ssH+0.05);
\node[font=\scriptsize,text=ssCache,anchor=north west,inner sep=0]
  at (-0.02,\ssbot-0.11) {grabbed early};
\node[font=\scriptsize,text=ssWin,anchor=north east,inner sep=0]
  at ({\ssP*5+0.86},\ssbot-0.11) {keep denoising};
\end{tikzpicture}%
\endgroup%
    \end{minipage}
    \vspace{-0.25em}
    \caption{
    (\textit{Left}) Cache Grab-Confidence.
    (\textit{Right}) Cache Grab-SSD.
    }
    \label{fig:cache_grab_sampling}
    \vspace{-0.75em}
\end{figure}
\subsection{Cache Grab Samplers: Accelerated Inference for ClockDLMs}\label{subsec:sampling_acceleration}
In our sampling algorithm, any tokens whose local clocks transition to $\ul(s) = 0$ are committed and added to the cache (lines 9--11 of Algorithm \ref{alg:clock_sampling}).
Here we introduce samplers that enlarge this set by optionally grabbing additional token positions for the cache.
These \textbf{Cache Grab} samplers can thus accelerate ClockDLM inference by skipping function evaluations for early-committed tokens.

\textbf{Cache Grab-Confidence}~
In discrete diffusion, several works accelerate generation with little to no cost in quality by utilizing heuristics to select which token positions to generate.
For example, \citet{wu2025fast} propose to greedily unmask all positions for which the softmax output associated with the $\argmax$ token exceeds some threshold.
We can adapt this heuristic for ClockDLMs.
Specifically, after every denoising step, we examine the probability associated with the $\argmax$ token at every position, and we \textit{early commit} any contiguous left-to-right positions that exceed our desired threshold, saving any remaining function evaluations in these tokens' denoising budgets.
The accepted tokens can then be added to the KV cache and all remaining tokens from the active window shift left, continuing their denoising progress from where they left off, as new noisy tokens are added to the rightmost slots of the window.
This sampler is depicted in Figure \ref{fig:cache_grab_sampling} and Algorithm \ref{alg:cache_grab_conf} in Appendix \ref{appsubsec:cache_grab_conf}.
In Appendix \ref{appsubsec:cache_grab_conf}, we present an adapted version of this sampler for B-ClockDLMs.

\textbf{Cache Grab-Self-Speculative Decoding}~
Notably, our sliding window attention mask coincides with the one proposed in hybrid AR + diffusion training \citep{gat2025set, fu2026nemotron}.
In particular, \citet{fu2026nemotron} use this mask to enable a `self-speculative decoding' (SSD) mode, a variant of speculative decoding \citep{chen2023accelerating, leviathan2023fast} where the diffusion drafter and AR verifier share all parameters.
Inspired by this work, we similarly propose a Cache Grab-SSD sampler.
As in \citet{fu2026nemotron}, we require a jointly trained AR prediction on the first half of the $2L$ training sequences, and at inference, we perform two function evaluations, one in diffusion mode to produce token drafts, and one in AR mode to verify and accept a prefix of the active window.
Unlike the linear self-speculative sampler of \citet{fu2026nemotron}, which restarts drafting from all mask tokens, Cache Grab-SSD retains the noisy states and denoising progress of uncommitted positions.
Figure \ref{fig:cache_grab_sampling} illustrates a single step of this sampler, and in Appendix \ref{appsubsec:cache_grab_ssd}, we provide more details.

\begin{table}[t]
\vspace{-1em}
\centering
\begin{minipage}[t]{0.415\textwidth}
\centering
\setlength{\tabcolsep}{3pt}%
    \scriptsize
    \caption{Language modeling evaluation (perplexity; PPL $\downarrow$) of models trained on OWT.
    Diffusion values represent variational bounds (PPL$_\leq$).
    \scriptsize{$^\dagger$ Reported in \citet{sahoo2024simple}. $^\ddag$ Reported in \citet{sahoo2025diffusion}. $^\$$ Reported in \citet{arriola2025block}. $^\mathsection$ Reported in \citet{arriola2026set}. $^\mathparagraph$ Reported in \citet{zheng2026continuously}. $^*$ Reported in \citet{yang2026continuous}.  $^\textsc{r}$ Retrained from scratch.}
    }
    \vspace{-1.1em}
    \begin{tabular}{lllc}
        \toprule
        &&& PPL ($\downarrow$) \\
        \midrule
        \multicolumn{3}{l}{\textit{Autoregressive Baseline}$^\dagger$} & 17.5 \\
        \midrule
        \multicolumn{4}{l}{\textit{Discrete Diffusion Baselines}} \\
        & \multicolumn{2}{l}{UDLM$^{\ddag}$ {\tiny \citep{schiff2024simple}}} & 27.4 \\
        & \multicolumn{2}{l}{Duo$^{\ddag}$ {\tiny \citep{sahoo2025diffusion}}} & 25.2 \\
        & \multicolumn{2}{l}{SEDD Absorb$^{\dagger}$ {\tiny \citep{loudiscrete}}} & 24.1 \\
        & \multicolumn{2}{l}{MDLM$^*$ {\tiny \citep{sahoo2024simple}}} & 23.1 \\
        & \multicolumn{1}{l}{BD3LM {\tiny \citep{arriola2025block}}} & \tiny{($S=16$)}$^{\$}$ & 22.3 \\
        & & \tiny{($S=4$)}$^{\$}$ & 20.7 \\
        & \multicolumn{1}{l}{SW-SetDLM {\tiny \citep{arriola2026set}}} & \tiny{($S \leq 8$)}$^\mathsection$ & 22.4 \\
        \midrule
        \multicolumn{4}{l}{\textit{Continuous Diffusion Baselines}} \\
        & \multicolumn{2}{l}{CADD$^\mathparagraph$ {\tiny \citep{zheng2026continuously}}} & 23.1 \\
        & \multicolumn{2}{l}{LangFlow$^*$ {\tiny \citep{chen2026langflow}}} & 32.2 \\
        & \multicolumn{2}{l}{Plaid$^*$ {\tiny \citep{gulrajani2023likelihood}}} & 24.4 \\
        & \multicolumn{2}{l}{RePlaid$^\textsc{r}$ {\tiny \citep{yang2026continuous}}} & 21.3 \\
        \multicolumn{4}{l}{\textit{Clock Diffusion (Ours)}} \\
        & B-ClockDLM & \tiny($S=64$) & 20.9 \\
        &  & \tiny($S=16$) & 20.8 \\
        &  & \tiny($S=4$) & 20.2 \\
        & ClockDLM & \tiny($S=64$) & 21.1 \\
        &  & \tiny($S=16$) & 20.6 \\
        &  & \tiny($S=4$) & \textbf{19.5} \\
        \bottomrule
    \end{tabular}
    \label{tab:owt_ppl}
\end{minipage}%
\hfill%
\begin{minipage}[t]{0.555\textwidth}
\centering
    \scriptsize
    \caption{
    Zero-shot, pass@$1$ accuracy (Acc \%; $\uparrow$) on the GSM8K test set for models trained on the TinyGSM dataset.
    NFE values are means $\pm$ sample standard deviation.
    \scriptsize{$^\dagger$ Models and code from \citet{deschenaux2026language}; $\mathbb{S}$-FLM accuracy quoted from that work. $^\ddag$ Models and code from \citet{agarwal2026posterior}. $^\textsc{r}$ Retrained from scratch.}
    \vspace{-0.8em}
    }
    \setlength{\tabcolsep}{3pt}
    \begin{tabular}{lllcc}
        \toprule
         & & & NFEs/seq. ($\downarrow$) & Acc. \% ($\uparrow$) \\
         & & & \tiny{Mean$_{\pm\text{SD}}$} & \tiny{$0$-shot, pass@$1$} \\
         \midrule
         \multicolumn{3}{l}{\textit{Autoregressive Baseline$^\textsc{r}$}}
         & 158.4$_{\pm\text{48.7}}$ & 64.3 \\
         \midrule
         \multicolumn{5}{l}{\textit{Discrete Diffusion Baselines}} \\
         & \multicolumn{2}{l}{Duo$^\dagger$ {\tiny \citep{sahoo2025diffusion}}} & 1024$_{\pm\text{0.0}}$ & 19.0 \\
         & \multicolumn{2}{l}{~~~w/ softmax temp = 0.1$^\dagger$} & 1024$_{\pm\text{0.0}}$ & 34.9 \\
         & \multicolumn{2}{l}{MDLM$^\textsc{r}$ {\tiny \citep{sahoo2024simple}}} & 112.4$_{\pm\text{5.6}}$ & 40.4 \\
         & \multicolumn{1}{l}{~~~+ SAR decoding} & \tiny($S=32$)$^\textsc{r}$ & 184.1$_{\pm\text{72.1}}$ & 53.1 \\
         & \multicolumn{1}{l}{BD3LM {\tiny \citep{arriola2025block}}} & \tiny($S= 32$)$^\textsc{r}$ & 177.7$_{\pm\text{48.2}}$ & 61.8 \\
         & \multicolumn{1}{l}{SW-SetDLM {\tiny \citep{arriola2026set}}} & \tiny($S\leq 64$)$^\textsc{r}$ & 174.1$_{\pm\text{53.6}}$ & 52.0 \\
         \midrule
         \multicolumn{4}{l}{\textit{Continuous Diffusion Baselines}} \\
         & \multicolumn{2}{l}{CANDI$^\dagger$ {\tiny \citep{pynadath2025candi}}} & 1022.0$_{\pm\text{2.8}}$ & 0.4 \\
         & \multicolumn{2}{l}{FLM$^\dagger$ {\tiny \citep{lee2026flow}}} & 1024$_{\pm\text{0.0}}$ & 0.8 \\
         & \multicolumn{2}{l}{$\mathbb{S}$-FLM$^\dagger$ {\tiny \citep{deschenaux2026language}}} & 1024$_{\pm\text{0.0}}$ & 18.0 \\
         & \multicolumn{2}{l}{FMLM$+^\ddag$ {\tiny \citep{agarwal2026posterior}}} & 45.1$_{\pm\text{33.3}}$ & 21.0 \\
         & \multicolumn{2}{l}{~~~MDLM Distill$^\ddag$} & 42.4$_{\pm\text{37.0}}$ & 26.1 \\
         & \multicolumn{2}{l}{~~~MDLM Init$^\ddag$} & 39.2$_{\pm\text{23.8}}$ & 34.7 \\
         & \multicolumn{2}{l}{RePlaid$^\textsc{r}$ {\tiny \citep{yang2026continuous}}} & 1025.0$_{\pm\text{0.0}}$ & 44.5 \\
         \multicolumn{5}{l}{\textit{Clock Diffusion (Ours)}} \\
         & B-ClockDLM & \tiny($S=32$) & 190.1$_{\pm\text{47.3}}$ & 54.3 \\
         & ClockDLM & \tiny($S=32$) & 188.1$_{\pm\text{44.8}}$ & 59.6 \\
         & \multicolumn{2}{l}{~~~+ Cache Grab-Conf.} & 59.5$_{\pm\text{27.6}}$ & 60.0 \\
         & \multicolumn{2}{l}{~~~+ Cache Grab-SSD} & \textbf{35.5}$_{\pm\text{21.6}}$ & \textbf{67.6} \\
         \bottomrule
    \end{tabular}
    \label{tab:gsm8k_accuracy}
\end{minipage}
\vspace{-1em}
\end{table}

\section{Experiments}\label{sec:exp}
\begin{highlightbox}
\textbf{Takeaway:} ClockDLMs outperform previous continuous models while matching and exceeding the best SAR discrete models, achieving SoTA diffusion likelihood on standard benchmarks.
\end{highlightbox}

\textbf{Baselines}~
We compare to autoregressive (AR) and discrete diffusion models which include uniform diffusion (UDLM \citep{schiff2024simple}, Duo \citep{sahoo2025diffusion}), masked diffusion \citep{sahoo2024simple}, and SAR masked diffusion models \citep{arriola2025block, arriola2026set}.
We also compare against several notable continuous flow language models: LangFlow \citep{chen2026langflow}, FLM \citep{lee2026flow}, $\mathbb{S}$-FLM \citep{deschenaux2026language}, and FMLM+ \citep{agarwal2026posterior}.
The most direct comparisons to our work are Plaid \citep{gulrajani2023likelihood} / RePlaid \citep{yang2026continuous}, which are the full sequence continuous DLM analogs of our ClockDLMs.

\subsection{Likelihood Evaluation}
\textbf{Experimental Setup}~
Following previous works \citep{sahoo2024simple, arriola2025block}, we train models on OpenWebText \citep{Gokaslan2019OpenWeb} for 1M gradient update steps.
We measure language modeling with perplexity (PPL $\downarrow$) and use variational lower bounds for diffusion models.
In Appendix \ref{appsubsec:exp_results_mauve}, we also evaluate generative quality using MAUVE \citep{pillutla2021mauve}.

\textbf{Results}~
In Table \ref{tab:owt_ppl}, we demonstrate SoTA likelihoods compared to all diffusion models, including both continuous DLMs and full sequence and SAR discrete diffusion.
Relative to BD3LM with $S=4$, we reduce the PPL gap to the AR baseline by approximately 38\%.
Notably, both ClockDLM and B-ClockDLM outperform comparable SAR discrete diffusion.

\begin{figure}[t!]
    \centering
    \includegraphics[width=\textwidth]{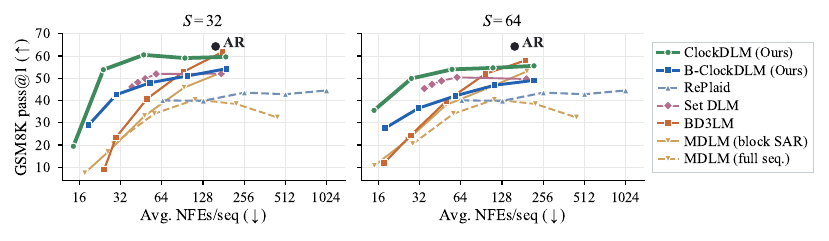}
    \vspace{-2em}
    \caption{Accuracy--efficiency trade-offs on GSM8K.
    Zero-shot test pass@$1$ versus average NFEs/seq for models trained on TinyGSM.
    \textit{(Left)} Active window size $S=32$.
    \textit{(Right)} $S=64$.}
    \label{fig:nfe_pareto}
\end{figure}

\subsection{Mathematical Reasoning}

\textbf{Experimental Setup}~
We train relatively lightweight models on the TinyGSM dataset \citep{liu2023tinygsm}, which consists of math questions and Python code responses that execute a reasoning trace to produce the answer.
We then evaluate on the GSM8K \citep{cobbe2021training} test set.

\textbf{Main Results}~
In Table \ref{tab:gsm8k_accuracy}, we show that ClockDLM significantly outperforms all previous continuous DLMs, with up to a $\sim$33\% increase in accuracy, and we match the performance of SAR discrete diffusion.
When utilizing Cache Grab techniques we increase performance and throughput.
Cache Grab-Confidence preserves accuracy with nearly 3$\times$ speedups; Cache Grab-SSD exceeds AR accuracy with the highest throughput among Table \ref{tab:gsm8k_accuracy} configurations.
\begin{wrapfigure}{l}{0.38\textwidth}
    \centering
    \vspace{-1em}
    \includegraphics[width=0.95\linewidth]{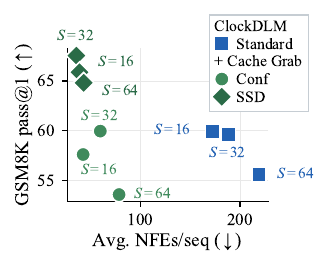}
    \vspace{-1em}
    \caption{GSM8K Cache Grab. Cache Grab-Conf. achieves training-free speedups with minimal to no quality degradation.
    SSD unlocks improved performance \& throughput.}
    \label{fig:gsm8k_cache_grab}
    \vspace{-1.3em}
\end{wrapfigure}

\textbf{Pareto Frontiers}~
Figure \ref{fig:nfe_pareto} evaluates the performance--efficiency trade-off by sweeping over denoising budgets.
ClockDLMs significantly push the frontier relative to the full sequence continuous diffusion baseline and match or exceed the performance of SAR discrete diffusion methods.
\begin{wraptable}{r}{0.57\textwidth}
\vspace{-4.3em}
\centering
\scriptsize
    \caption{Summarization metrics (ROUGE-1/2/L; $\uparrow$) on CNN/DailyMail.
    NFE values are means $\pm$ sample standard deviation.
    \scriptsize{All models trained from scratch.}}
    \setlength{\tabcolsep}{4pt}
    \begin{tabular}{lllcccc}
        \toprule
        & & & NFEs/seq. ($\downarrow$) & \multicolumn{3}{c}{ROUGE ($\uparrow$)} \\
        \cmidrule(lr){5-7}
        & & & \tiny{Mean$_{\pm\text{SD}}$} & 1 & 2 & L \\
        \midrule    
         \multicolumn{3}{l}{\textit{Autoregressive}} & 73.1$_{\pm\text{13.4}}$ & 35.9 & 14.8 & 25.0 \\
         \midrule
         \multicolumn{7}{l}{\textit{Discrete Diffusion Baselines}} \\
         & MDLM & \tiny(full seq.) & 180.0$_{\pm\text{0.0}}$ & 39.5 & 17.1 & 26.1 \\
         & ~~~\tiny{\citep{sahoo2024simple}} & \tiny($S=64$) & 128.0$_{\pm\text{0.0}}$ & 40.1 & 18.1 & 27.2 \\
         & & \tiny($S=32$) & 96.0$_{\pm\text{0.0}}$ & 40.0 & 18.0 & 27.1 \\
         & BD3LM & \tiny($S=64$) & 126.9$_{\pm\text{19.4}}$ & 39.8 & 17.7 & 27.2 \\
         & ~~~\tiny{\citep{arriola2025block}} & \tiny($S=32$) & 110.7$_{\pm\text{9.3}}$ & 39.4 & 17.5 & 27.0 \\
         & SW-SetDLM & \tiny($S \leq 128$) & 74.9$_{\pm\text{14.2}}$ & 38.8 & 17.1 & 27.1 \\
         & ~~~\tiny{\citep{arriola2026set}} & \tiny($S \leq 64$) & 72.2$_{\pm\text{16.2}}$ & 39.7 & 17.9 & 27.8 \\
         \midrule
         \multicolumn{7}{l}{\textit{Continuous Diffusion Baseline}} \\
         & \multicolumn{1}{l}{RePlaid} & \tiny{($T=1024$)} & 1025.0$_{\pm\text{0.0}}$ & 32.2 & 12.7 & 22.1 \\
         & ~~~\tiny{\citep{yang2026continuous}} & \tiny{($T=128$)} & 129$_{\pm\text{0.0}}$ & 30.4 & 11.1 & 20.7 \\
         \multicolumn{7}{l}{\textit{Clock Diffusion (Ours)}} \\
         & B-ClockDLM & \tiny($S=64$) & 120.2$_{\pm\text{32.0}}$ & 34.3 & 12.7 & 22.9 \\
         & & \tiny($S=32$) & 91.3$_{\pm\text{18.3}}$ & 33.6 & 11.3 & 21.8 \\
         & ClockDLM & \tiny($S=64$) & 152.4$_{\pm\text{0.6}}$ & 34.0 & 11.7 & 21.7 \\
         & & \tiny($S=32$) & 120.3$_{\pm\text{0.8}}$ & 36.0 & 14.1 & 23.8 \\
        \bottomrule
    \end{tabular}
    \label{tab:cnndm}
\vspace{-5.2em}
\end{wraptable}

\textbf{Cache Grab Improvements}~
In Figure \ref{fig:gsm8k_cache_grab}, we further explore the benefits of the Cache Grab samplers.
Evaluating models trained at various active window sizes, $S \in \{16, 32, 64\}$, we show how Cache Grab-Conf. leads to minor to no accuracy drop-off with $\sim$3--4$\times$ fewer NFEs/seq across window sizes and that Cache Grab-SSD unlocks improved performance with up to $\sim$5$\times$ fewer NFEs/seq.
Notably, consistent with previous works that developed SSD for discrete diffusion, we also find that joint AR and diffusion loss training is complementary, increasing the performance of both AR and diffusion-only decoding (see Table \ref{tab:gsm8k_ssd_ablation} in Appendix \ref{appsec:exp_results}).

\subsection{Summarization}
\textbf{Experimental Setup}~
We train summarization models on CNN/DailyMail \citep{DBLP:conf/nips/HermannKGEKSB15, see-etal-2017-get}, following \citet{arriola2026encoder, arriola2026set}.

\textbf{Results}~
Table \ref{tab:cnndm} reports NFEs/seq alongside ROUGE-1/2/L.
ClockDLM ($S=32$) improves all three ROUGE scores over RePlaid ($T=1024$) with $\sim8.5\times$ fewer NFEs.
In Figure \ref{fig:cnn_cache_grab}, we again see the benefits of the accelerated Cache Grab-Confidence sampler.
Reducing the confidence threshold from 0.99 to 0.7 requires $\sim$1.1--1.2$\times$ fewer NFEs/seq with at most a 1.3\% reduction in ROUGE-1.

\subsection{Ablations}\label{subsec:exp_ablations}

\textbf{Block vs.\ Sliding Window}~
We often find that ClockDLMs, which are based on sliding window clocks, outperform their block counterparts, B-ClockDLMs.
This may reflect the additional flexibility of ClockDLMs, which have a more left-to-right decoding scheme even within active windows compared to block diffusion.

\textbf{Effect of Window Size}~
We find the expected pattern that in almost all cases moving to smaller blocks / window sizes improves performance as generation becomes more like AR.
We note that our results are robust to window size; e.g., at matched blocks for BD3LM, we consistently improve language modeling perplexity (Table \ref{tab:owt_ppl}) and the Pareto frontier of quality vs. efficiency (Figure \ref{fig:nfe_pareto}).

\begin{wrapfigure}{r}{0.36\textwidth}
\vspace{-6.7em}
\centering
\includegraphics[width=\linewidth]{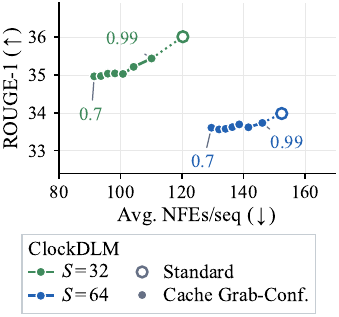}
\vspace{-1.5em}
\caption{Cache Grab-Confidence on CNN/DailyMail.
ROUGE-1 vs.\ NFEs/seq for ClockDLMs, sweeping the early-commit confidence threshold.
Hollow markers use no early commit, i.e., rows in Table \ref{tab:cnndm}.}
\label{fig:cnn_cache_grab}
\vspace{-1.3em}
\end{wrapfigure}

\textbf{Varying Confidence Threshold}~
In Figure \ref{fig:cnn_cache_grab}, we see that speed gains grow as the confidence threshold is lowered, as expected, and we see a fairly monotonic degradation in quality.
In Table \ref{tab:gsm8k_accuracy_cache_grab_conf} in Appendix \ref{appsubsec:exp_results_conf_gsm8k}, this monotonic relationship is less pronounced.

\section{Related Work}
Our work is most related to the recent continuous DLMs that revived token embedding diffusion with modernized architectures and training recipes \citep{chen2026langflow, yang2026continuous}.
We also leverage vectorized training methods introduced in block discrete diffusion \citep{arriola2025block}.
\citet{han2022ssd} used SAR decoding with diffusion over logits, but lagged in quality relative to discrete diffusion and did not propose our efficient training and KV cache sampling.

\textbf{Position-dependent Noise Schedules}~
Previous diffusion works have explored position-dependent noise.
AR Diffusion \citep{wu2023ar} biased noise across the full sequence to have leftmost tokens be less noisy, and Diffusion Forcing \citep{chen2024diffusion} generalized this method.
Most closely related is Rolling Diffusion \citep{ruhe2024rolling}, which formalized a process akin to our Sliding Window Clock Diffusion.
However, unlike these works, we propose efficient vectorized training and KV cache sampling.
Furthermore, we complement our sampling algorithm with the Cache Grab samplers, which significantly boost throughput.
Finally, both \citet{chen2024diffusion} and \citet{ruhe2024rolling} examined continuous data, such as video frames, while we have adapted this method to language and derived each component of the variational bound in this setting.


\section{Discussion \& Conclusion}
In this work, we introduced a semi-autoregressive formulation for continuous diffusion language modeling that relies on position-dependent noise schedules that determine which tokens are actively being denoised at any given diffusion step.
We complement this parameterization with efficient training recipes and KV cache sampling and call this framework Clock Diffusion.
We demonstrated two instantiations of Clock Diffusion that enable block and sliding window SAR generation.
Additionally, we leveraged the sliding window formulation to develop accelerated sampling techniques that adapt confidence-threshold and self-speculative decoding from discrete diffusion.
We introduced models that use Clock Diffusion with Gaussian noise on token embeddings, ClockDLMs, and trained them on unconditional and conditional language modeling tasks.
We attained SoTA language modeling performance measured in both perplexity and downstream benchmark metrics.
We consistently outperformed full sequence continuous language models on both performance and efficiency, demonstrating the usefulness of our SAR parameterization and KV cache enabled sampling.

\textbf{Limitations}~
The improved language modeling performance of our method comes at the cost of decoding flexibility, which may be beneficial in certain domains that lack a more strict left-to-right bias.
Additionally, we have presently only explored relatively smaller scale models.

\textbf{Future Directions}~
In future efforts, we intend to address the limitations of our work with extensions to any-order sliding window implementations.
We also plan to test our method at larger scales and explore the efficacy of leveraging pre-trained AR and discrete diffusion models.
Finally, we intend to experiment with hybrid discrete and continuous parameterizations.

\section*{Reproducibility Statement}
We provide the full derivation for the variational objective employed by our method in Appendix \ref{appsec:clock_nelbo} and the models we trained are explicitly defined in Appendix \ref{appsec:clock_dlms}.
All relevant training details and training and evaluation hyperparameter settings are explicitly laid out in Appendices \ref{appsec:training} and \ref{appsec:exp_setup}.

\section*{AI Use Statement}

In this work, we used generative AI tools, specifically large language models (LLMs), to assist in the derivation of the mathematical formulation presented in Section \ref{sec:clock} and Appendix \ref{appsec:clock_nelbo}.
Generative AI was used in developing code to run experiments and in producing the tables and figures that report our results.
We also used LLMs to summarize existing literature, edit the formatting of this work, and format references.
We further used LLMs to assist in the writing of proofs, propose or refine hypotheses, design or provide feedback on research methodology or experiments, implement methods, and interpret experimental results.
We have not used generative AI tools to generate synthetic datasets, assist with translation, or clean and reformat datasets.
We take responsibility for the final content of this work,
including text, claims or artifacts produced with the aid of generative AI.

\bibliographystyle{plainnat-etal}
\bibliography{refs}

\begin{thebibliography}{65}
\providecommand{\natexlab}[1]{#1}
\providecommand{\url}[1]{\texttt{#1}}
\expandafter\ifx\csname urlstyle\endcsname\relax
  \providecommand{\doi}[1]{doi: #1}\else
  \providecommand{\doi}{doi: \begingroup \urlstyle{rm}\Url}\fi

\bibitem[Agarwal et~al.(2026)Agarwal, Shah, Lee, Yoo, Huang, Hong, Raghunathan,
  Kim, and Boffi]{agarwal2026posterior}
Manan Agarwal, Sheel Shah, Chanhyuk Lee, Jaehoon Yoo, Jerry Huang, Seunghoon
  Hong, Aditi Raghunathan, Jinwoo Kim, and Nicholas~M Boffi.
\newblock Posterior refinement: Fast language generation via any-order flow
  maps.
\newblock \emph{arXiv preprint arXiv:2606.24773}, 2026.

\bibitem[Allal et~al.(2024)Allal, Lozhkov, Bakouch, von Werra, and
  Wolf]{allal2024SmolLM}
Loubna~Ben Allal, Anton Lozhkov, Elie Bakouch, Leandro von Werra, and Thomas
  Wolf.
\newblock Smollm - blazingly fast and remarkably powerful, 2024.

\bibitem[Arriola et~al.(2026)Arriola and Kuleshov]{arriola2026set}
Marianne Arriola and Volodymyr Kuleshov.
\newblock Set diffusion: Interpolating token orderings between autoregression
  and diffusion for fast and flexible decoding.
\newblock \emph{arXiv preprint arXiv:2607.01775}, 2026.

\bibitem[Arriola et~al.(2025{\natexlab{a}})Arriola, Gokaslan, Chiu, Yang, Qi,
  Han, Sahoo, and Kuleshov]{arriola2025block}
Marianne Arriola, Aaron Gokaslan, Justin~T Chiu, Zhihan Yang, Zhixuan Qi, Jiaqi
  Han, Subham~Sekhar Sahoo, and Volodymyr Kuleshov.
\newblock Block diffusion: Interpolating between autoregressive and diffusion
  language models.
\newblock \emph{arXiv preprint arXiv:2503.09573}, 2025{\natexlab{a}}.

\bibitem[Arriola et~al.(2025{\natexlab{b}})Arriola, Schiff, Phung, Gokaslan,
  and Kuleshov]{arriola2026encoder}
Marianne Arriola, Yair Schiff, Hao Phung, Aaron Gokaslan, and Volodymyr
  Kuleshov.
\newblock Encoder-decoder diffusion language models for efficient training and
  inference.
\newblock \emph{Advances in Neural Information Processing Systems},
  38:\penalty0 15612--15644, 2025{\natexlab{b}}.

\bibitem[Austin et~al.(2021)Austin, Johnson, Ho, Tarlow, and Van
  Den~Berg]{austin2021structured}
Jacob Austin, Daniel~D Johnson, Jonathan Ho, Daniel Tarlow, and Rianne Van
  Den~Berg.
\newblock Structured denoising diffusion models in discrete state-spaces.
\newblock \emph{Advances in Neural Information Processing Systems},
  34:\penalty0 17981--17993, 2021.

\bibitem[Campbell et~al.(2022)Campbell, Benton, De~Bortoli, Rainforth,
  Deligiannidis, and Doucet]{campbell2022continuous}
Andrew Campbell, Joe Benton, Valentin De~Bortoli, Thomas Rainforth, George
  Deligiannidis, and Arnaud Doucet.
\newblock A continuous time framework for discrete denoising models.
\newblock \emph{Advances in Neural Information Processing Systems},
  35:\penalty0 28266--28279, 2022.

\bibitem[Chen et~al.(2024)Chen, Mart{\'\i}~Mons{\'o}, Du, Simchowitz, Tedrake,
  and Sitzmann]{chen2024diffusion}
Boyuan Chen, Diego Mart{\'\i}~Mons{\'o}, Yilun Du, Max Simchowitz, Russ
  Tedrake, and Vincent Sitzmann.
\newblock Diffusion forcing: Next-token prediction meets full-sequence
  diffusion.
\newblock \emph{Advances in Neural Information Processing Systems},
  37:\penalty0 24081--24125, 2024.

\bibitem[Chen et~al.(2023)Chen, Borgeaud, Irving, Lespiau, Sifre, and
  Jumper]{chen2023accelerating}
Charlie Chen, Sebastian Borgeaud, Geoffrey Irving, Jean-Baptiste Lespiau,
  Laurent Sifre, and John Jumper.
\newblock Accelerating large language model decoding with speculative sampling.
\newblock \emph{arXiv preprint arXiv:2302.01318}, 2023.

\bibitem[Chen et~al.(2022)Chen, Zhang, and Hinton]{chen2022analog}
Ting Chen, Ruixiang Zhang, and Geoffrey Hinton.
\newblock Analog bits: Generating discrete data using diffusion models with
  self-conditioning.
\newblock \emph{arXiv preprint arXiv:2208.04202}, 2022.

\bibitem[Chen et~al.(2026)Chen, Liang, Sui, Guo, Cheng, You, and
  Liu]{chen2026langflow}
Yuxin Chen, Chumeng Liang, Hangke Sui, Ruihan Guo, Chaoran Cheng, Jiaxuan You,
  and Ge~Liu.
\newblock Langflow: Continuous diffusion rivals discrete in language modeling.
\newblock \emph{arXiv preprint arXiv:2604.11748}, 2026.

\bibitem[Cobbe et~al.(2021)Cobbe, Kosaraju, Bavarian, Chen, Jun, Kaiser,
  Plappert, Tworek, Hilton, Nakano, et~al.]{cobbe2021training}
Karl Cobbe, Vineet Kosaraju, Mohammad Bavarian, Mark Chen, Heewoo Jun, Lukasz
  Kaiser, Matthias Plappert, Jerry Tworek, Jacob Hilton, Reiichiro Nakano,
  et~al.
\newblock Training verifiers to solve math word problems.
\newblock \emph{arXiv preprint arXiv:2110.14168}, 2021.

\bibitem[Davis et~al.(2026)Davis, Filippova, Ablin, Turrisi, Shidani, Cuturi,
  and B{\'e}thune]{davis2026scaling}
Oscar Davis, Anastasiia Filippova, Pierre Ablin, Victor Turrisi, Amitis
  Shidani, Marco Cuturi, and Louis B{\'e}thune.
\newblock Scaling categorical flow maps.
\newblock \emph{arXiv preprint arXiv:2605.07820}, 2026.

\bibitem[Deschenaux et~al.(2026)Deschenaux and
  Gulcehre]{deschenaux2026language}
Justin Deschenaux and Caglar Gulcehre.
\newblock Language modeling with hyperspherical flows.
\newblock \emph{arXiv preprint arXiv:2605.11125}, 2026.

\bibitem[Dieleman et~al.(2022)Dieleman, Sartran, Roshannai, Savinov, Ganin,
  Richemond, Doucet, Strudel, Dyer, Durkan, et~al.]{dieleman2022continuous}
Sander Dieleman, Laurent Sartran, Arman Roshannai, Nikolay Savinov, Yaroslav
  Ganin, Pierre~H Richemond, Arnaud Doucet, Robin Strudel, Chris Dyer, Conor
  Durkan, et~al.
\newblock Continuous diffusion for categorical data.
\newblock \emph{arXiv preprint arXiv:2211.15089}, 2022.

\bibitem[Falcon et~al.(2019)Falcon and {The PyTorch Lightning
  team}]{Falcon_PyTorch_Lightning_2019}
William Falcon and {The PyTorch Lightning team}.
\newblock {PyTorch Lightning}, March 2019.
\newblock URL \url{https://github.com/Lightning-AI/lightning}.

\bibitem[Fu et~al.(2026)Fu, Whalen, Garg, Wu, Khadkevich, Oswald, Xie, Egert,
  Sreenivas, Diao, et~al.]{fu2026nemotron}
Yonggan Fu, Lexington Whalen, Abhinav Garg, Chengyue Wu, Maksim Khadkevich,
  Nicolai Oswald, Enze Xie, Daniel Egert, Sharath~Turuvekere Sreenivas, Shizhe
  Diao, et~al.
\newblock Nemotron-labs-diffusion: A tri-mode language model unifying
  autoregressive, diffusion, and self-speculation decoding.
\newblock \emph{arXiv preprint arXiv:2607.05722}, 2026.

\bibitem[Gao et~al.(2021)Gao, Tow, Biderman, Black, DiPofi, Foster, Golding,
  Hsu, McDonell, Muennighoff, et~al.]{gao2021framework}
Leo Gao, Jonathan Tow, Stella Biderman, Sid Black, Anthony DiPofi, Charles
  Foster, Laurence Golding, Jeffrey Hsu, Kyle McDonell, Niklas Muennighoff,
  et~al.
\newblock A framework for few-shot language model evaluation.
\newblock \emph{Zenodo}, 2021.

\bibitem[Gat et~al.(2025)Gat, Ben-Hamu, Havasi, Haziza, Reizenstein, Synnaeve,
  Lopez-Paz, Karrer, and Lipman]{gat2025set}
Itai Gat, Heli Ben-Hamu, Marton Havasi, Daniel Haziza, Jeremy Reizenstein,
  Gabriel Synnaeve, David Lopez-Paz, Brian Karrer, and Yaron Lipman.
\newblock Set block decoding is a language model inference accelerator.
\newblock \emph{arXiv preprint arXiv:2509.04185}, 2025.

\bibitem[Gokaslan et~al.(2019)Gokaslan and Cohen]{Gokaslan2019OpenWeb}
Aaron Gokaslan and Vanya Cohen.
\newblock Openwebtext corpus.
\newblock \url{http://Skylion007.github.io/OpenWebTextCorpus}, 2019.

\bibitem[Gulrajani et~al.(2023)Gulrajani and
  Hashimoto]{gulrajani2023likelihood}
Ishaan Gulrajani and Tatsunori~B Hashimoto.
\newblock Likelihood-based diffusion language models.
\newblock \emph{Advances in Neural Information Processing Systems},
  36:\penalty0 16693--16715, 2023.

\bibitem[Han et~al.(2022)Han, Kumar, and Tsvetkov]{han2022ssd}
Xiaochuang Han, Sachin Kumar, and Yulia Tsvetkov.
\newblock Ssd-lm: Semi-autoregressive simplex-based diffusion language model
  for text generation and modular control.
\newblock \emph{arXiv preprint arXiv:2210.17432}, 2022.

\bibitem[Harris et~al.(2020)Harris, Millman, van~der Walt, Gommers, Virtanen,
  Cournapeau, Wieser, Taylor, Berg, Smith, Kern, Picus, Hoyer, van Kerkwijk,
  Brett, Haldane, del R{\'{i}}o, Wiebe, Peterson, G{\'{e}}rard-Marchant,
  Sheppard, Reddy, Weckesser, Abbasi, Gohlke, and Oliphant]{harris2020array}
Charles~R. Harris, K.~Jarrod Millman, St{\'{e}}fan~J. van~der Walt, Ralf
  Gommers, Pauli Virtanen, David Cournapeau, Eric Wieser, Julian Taylor,
  Sebastian Berg, Nathaniel~J. Smith, Robert Kern, Matti Picus, Stephan Hoyer,
  Marten~H. van Kerkwijk, Matthew Brett, Allan Haldane, Jaime~Fern{\'{a}}ndez
  del R{\'{i}}o, Mark Wiebe, Pearu Peterson, Pierre G{\'{e}}rard-Marchant,
  Kevin Sheppard, Tyler Reddy, Warren Weckesser, Hameer Abbasi, Christoph
  Gohlke, and Travis~E. Oliphant.
\newblock Array programming with {NumPy}.
\newblock \emph{Nature}, 585\penalty0 (7825):\penalty0 357--362, September
  2020.
\newblock \doi{10.1038/s41586-020-2649-2}.
\newblock URL \url{https://doi.org/10.1038/s41586-020-2649-2}.

\bibitem[Hermann et~al.(2015)Hermann, Ko{\v{c}}isk{\'y}, Grefenstette,
  Espeholt, Kay, Suleyman, and Blunsom]{DBLP:conf/nips/HermannKGEKSB15}
Karl~Moritz Hermann, Tom{\'a}{\v{s}} Ko{\v{c}}isk{\'y}, Edward Grefenstette,
  Lasse Espeholt, Will Kay, Mustafa Suleyman, and Phil Blunsom.
\newblock Teaching machines to read and comprehend.
\newblock In \emph{NIPS}, pages 1693--1701, 2015.
\newblock URL
  \url{http://papers.nips.cc/paper/5945-teaching-machines-to-read-and-comprehend}.

\bibitem[Ho et~al.(2020)Ho, Jain, and Abbeel]{ho2020denoising}
Jonathan Ho, Ajay Jain, and Pieter Abbeel.
\newblock Denoising diffusion probabilistic models.
\newblock \emph{Advances in neural information processing systems},
  33:\penalty0 6840--6851, 2020.

\bibitem[Hu et~al.(2026)Hu, Qiu, Lu, Zhao, Li, Kim, Andreas, and He]{hu2026elf}
Keya Hu, Linlu Qiu, Yiyang Lu, Hanhong Zhao, Tianhong Li, Yoon Kim, Jacob
  Andreas, and Kaiming He.
\newblock Elf: Embedded language flows.
\newblock \emph{arXiv preprint arXiv:2605.10938}, 2026.

\bibitem[Hunter(2007)]{Hunter:2007}
J.~D. Hunter.
\newblock Matplotlib: A 2d graphics environment.
\newblock \emph{Computing in Science \& Engineering}, 9\penalty0 (3):\penalty0
  90--95, 2007.
\newblock \doi{10.1109/MCSE.2007.55}.

\bibitem[Kingma et~al.(2021)Kingma, Salimans, Poole, and
  Ho]{kingma2021variational}
Diederik Kingma, Tim Salimans, Ben Poole, and Jonathan Ho.
\newblock Variational diffusion models.
\newblock \emph{Advances in neural information processing systems},
  34:\penalty0 21696--21707, 2021.

\bibitem[Kingma et~al.(2014)Kingma and Ba]{kingma2014adam}
Diederik~P Kingma and Jimmy Ba.
\newblock Adam: A method for stochastic optimization.
\newblock \emph{arXiv preprint arXiv:1412.6980}, 2014.

\bibitem[Lee et~al.(2026)Lee, Yoo, Agarwal, Shah, Huang, Raghunathan, Hong,
  Boffi, and Kim]{lee2026flow}
Chanhyuk Lee, Jaehoon Yoo, Manan Agarwal, Sheel Shah, Jerry Huang, Aditi
  Raghunathan, Seunghoon Hong, Nicholas~M Boffi, and Jinwoo Kim.
\newblock Flow map language models: One-step language modeling via continuous
  denoising.
\newblock \emph{arXiv preprint arXiv:2602.16813}, 2026.

\bibitem[Leviathan et~al.(2023)Leviathan, Kalman, and
  Matias]{leviathan2023fast}
Yaniv Leviathan, Matan Kalman, and Yossi Matias.
\newblock Fast inference from transformers via speculative decoding.
\newblock In \emph{International conference on machine learning}, pages
  19274--19286. PMLR, 2023.

\bibitem[Li et~al.(2022)Li, Thickstun, Gulrajani, Liang, and
  Hashimoto]{li2022diffusion}
Xiang Li, John Thickstun, Ishaan Gulrajani, Percy~S Liang, and Tatsunori~B
  Hashimoto.
\newblock Diffusion-lm improves controllable text generation.
\newblock \emph{Advances in Neural Information Processing Systems},
  35:\penalty0 4328--4343, 2022.

\bibitem[Lin(2004)]{lin2004rouge}
Chin-Yew Lin.
\newblock Rouge: A package for automatic evaluation of summaries.
\newblock In \emph{Text summarization branches out}, pages 74--81, 2004.

\bibitem[Liu et~al.(2023)Liu, Bubeck, Eldan, Kulkarni, Li, Nguyen, Ward, and
  Zhang]{liu2023tinygsm}
Bingbin Liu, Sebastien Bubeck, Ronen Eldan, Janardhan Kulkarni, Yuanzhi Li, Anh
  Nguyen, Rachel Ward, and Yi~Zhang.
\newblock Tinygsm: achieving> 80\% on gsm8k with small language models.
\newblock \emph{arXiv preprint arXiv:2312.09241}, 2023.

\bibitem[Loshchilov et~al.(2017)Loshchilov and Hutter]{loshchilov2017decoupled}
Ilya Loshchilov and Frank Hutter.
\newblock Decoupled weight decay regularization.
\newblock \emph{arXiv preprint arXiv:1711.05101}, 2017.

\bibitem[Lou et~al.(2024)Lou, Meng, and Ermon]{loudiscrete}
Aaron Lou, Chenlin Meng, and Stefano Ermon.
\newblock Discrete diffusion modeling by estimating the ratios of the data
  distribution.
\newblock In \emph{Forty-first International Conference on Machine Learning},
  2024.

\bibitem[Nie et~al.(2025)Nie, Zhu, You, Zhang, Ou, Hu, Zhou, Lin, Wen, and
  Li]{nie2025large}
Shen Nie, Fengqi Zhu, Zebin You, Xiaolu Zhang, Jingyang Ou, Jun Hu, Jun Zhou,
  Yankai Lin, Ji-Rong Wen, and Chongxuan Li.
\newblock Large language diffusion models.
\newblock \emph{arXiv preprint arXiv:2502.09992}, 2025.

\bibitem[pandas~development team(2020)]{reback2020pandas}
The pandas~development team.
\newblock pandas-dev/pandas: Pandas, February 2020.
\newblock URL \url{https://doi.org/10.5281/zenodo.3509134}.

\bibitem[Paszke et~al.(2019)Paszke, Gross, Massa, Lerer, Bradbury, Chanan,
  Killeen, Lin, Gimelshein, Antiga, Desmaison, Kopf, Yang, DeVito, Raison,
  Tejani, Chilamkurthy, Steiner, Fang, Bai, and
  Chintala]{Paszke_PyTorch_An_Imperative_2019}
Adam Paszke, Sam Gross, Francisco Massa, Adam Lerer, James Bradbury, Gregory
  Chanan, Trevor Killeen, Zeming Lin, Natalia Gimelshein, Luca Antiga, Alban
  Desmaison, Andreas Kopf, Edward Yang, Zachary DeVito, Martin Raison, Alykhan
  Tejani, Sasank Chilamkurthy, Benoit Steiner, Lu~Fang, Junjie Bai, and Soumith
  Chintala.
\newblock {PyTorch: An Imperative Style, High-Performance Deep Learning
  Library}.
\newblock In H.~Wallach, H.~Larochelle, A.~Beygelzimer, F.~d'Alché Buc,
  E.~Fox, and R.~Garnett, editors, \emph{Advances in Neural Information
  Processing Systems 32}, pages 8024--8035. Curran Associates, Inc., 2019.

\bibitem[Peebles et~al.(2023)Peebles and Xie]{peebles2023scalable}
William Peebles and Saining Xie.
\newblock Scalable diffusion models with transformers.
\newblock In \emph{Proceedings of the IEEE/CVF International Conference on
  Computer Vision}, pages 4195--4205, 2023.

\bibitem[Pillutla(2021)]{mauve_github}
Krishna Pillutla.
\newblock mauve.
\newblock \url{https://github.com/krishnap25/mauve}, 2021.
\newblock Accessed: 2025-05-14.

\bibitem[Pillutla et~al.(2021)Pillutla, Swayamdipta, Zellers, Thickstun,
  Welleck, Choi, and Harchaoui]{pillutla2021mauve}
Krishna Pillutla, Swabha Swayamdipta, Rowan Zellers, John Thickstun, Sean
  Welleck, Yejin Choi, and Zaid Harchaoui.
\newblock Mauve: Measuring the gap between neural text and human text using
  divergence frontiers.
\newblock \emph{Advances in Neural Information Processing Systems},
  34:\penalty0 4816--4828, 2021.

\bibitem[Potaptchik et~al.(2026)Potaptchik, Yim, Saravanan, Holderrieth,
  Vanden-Eijnden, and Albergo]{potaptchik2026discrete}
Peter Potaptchik, Jason Yim, Adhi Saravanan, Peter Holderrieth, Eric
  Vanden-Eijnden, and Michael~S Albergo.
\newblock Discrete flow maps.
\newblock \emph{arXiv preprint arXiv:2604.09784}, 2026.

\bibitem[Pynadath et~al.(2025)Pynadath, Shi, and Zhang]{pynadath2025candi}
Patrick Pynadath, Jiaxin Shi, and Ruqi Zhang.
\newblock Candi: Hybrid discrete-continuous diffusion models.
\newblock \emph{arXiv preprint arXiv:2510.22510}, 2025.

\bibitem[Radford et~al.(2019)Radford, Wu, Child, Luan, Amodei, Sutskever,
  et~al.]{radford2019language}
Alec Radford, Jeffrey Wu, Rewon Child, David Luan, Dario Amodei, Ilya
  Sutskever, et~al.
\newblock Language models are unsupervised multitask learners.
\newblock \emph{OpenAI blog}, 1\penalty0 (8):\penalty0 9, 2019.

\bibitem[Roos et~al.(2026)Roos, Davis, Eijkelboom, Bronstein, Welling, Ceylan,
  Ambrogioni, and van~de Meent]{roos2026categorical}
Daan Roos, Oscar Davis, Floor Eijkelboom, Michael Bronstein, Max Welling,
  Ismail~Ilkan Ceylan, Luca Ambrogioni, and Jan-Willem van~de Meent.
\newblock Categorical flow maps.
\newblock \emph{arXiv preprint arXiv:2602.12233}, 2026.

\bibitem[Ruhe et~al.(2024)Ruhe, Heek, Salimans, and Hoogeboom]{ruhe2024rolling}
David Ruhe, Jonathan Heek, Tim Salimans, and Emiel Hoogeboom.
\newblock Rolling diffusion models.
\newblock \emph{arXiv preprint arXiv:2402.09470}, 2024.

\bibitem[Sahoo et~al.(2024{\natexlab{a}})Sahoo, Arriola, Schiff, Gokaslan,
  Marroquin, Chiu, Rush, and Kuleshov]{sahoo2024simple}
Subham~Sekhar Sahoo, Marianne Arriola, Yair Schiff, Aaron Gokaslan,
  Edgar~Mariano Marroquin, Justin~T Chiu, Alexander~M Rush, and Volodymyr
  Kuleshov.
\newblock Simple and effective masked diffusion language models.
\newblock In \emph{The Thirty-eighth Annual Conference on Neural Information
  Processing Systems}, 2024{\natexlab{a}}.
\newblock URL \url{https://openreview.net/forum?id=L4uaAR4ArM}.

\bibitem[Sahoo et~al.(2024{\natexlab{b}})Sahoo, Gokaslan, Sa, and
  Kuleshov]{sahoo2024diffusion}
Subham~Sekhar Sahoo, Aaron Gokaslan, Christopher~De Sa, and Volodymyr Kuleshov.
\newblock Diffusion models with learned adaptive noise.
\newblock In \emph{The Thirty-eighth Annual Conference on Neural Information
  Processing Systems}, 2024{\natexlab{b}}.
\newblock URL \url{https://openreview.net/forum?id=loMa99A4p8}.

\bibitem[Sahoo et~al.(2025)Sahoo, Deschenaux, Gokaslan, Wang, Chiu, and
  Kuleshov]{sahoo2025diffusion}
Subham~Sekhar Sahoo, Justin Deschenaux, Aaron Gokaslan, Guanghan Wang, Justin~T
  Chiu, and Volodymyr Kuleshov.
\newblock The diffusion duality.
\newblock In \emph{ICLR 2025 Workshop on Deep Generative Model in Machine
  Learning: Theory, Principle and Efficacy}, 2025.

\bibitem[Schiff et~al.(2024)Schiff, Sahoo, Phung, Wang, Boshar, Dalla-torre,
  de~Almeida, Rush, Pierrot, and Kuleshov]{schiff2024simple}
Yair Schiff, Subham~Sekhar Sahoo, Hao Phung, Guanghan Wang, Sam Boshar, Hugo
  Dalla-torre, Bernardo~P de~Almeida, Alexander Rush, Thomas Pierrot, and
  Volodymyr Kuleshov.
\newblock Simple guidance mechanisms for discrete diffusion models.
\newblock \emph{arXiv preprint arXiv:2412.10193}, 2024.

\bibitem[See et~al.(2017)See, Liu, and Manning]{see-etal-2017-get}
Abigail See, Peter~J. Liu, and Christopher~D. Manning.
\newblock Get to the point: Summarization with pointer-generator networks.
\newblock In \emph{Proceedings of the 55th Annual Meeting of the Association
  for Computational Linguistics (Volume 1: Long Papers)}, pages 1073--1083,
  Vancouver, Canada, July 2017. Association for Computational Linguistics.
\newblock \doi{10.18653/v1/P17-1099}.
\newblock URL \url{https://www.aclweb.org/anthology/P17-1099}.

\bibitem[Sohl-Dickstein et~al.(2015)Sohl-Dickstein, Weiss, Maheswaranathan, and
  Ganguli]{sohl2015deep}
Jascha Sohl-Dickstein, Eric Weiss, Niru Maheswaranathan, and Surya Ganguli.
\newblock Deep unsupervised learning using nonequilibrium thermodynamics.
\newblock In \emph{International conference on machine learning}, pages
  2256--2265. PMLR, 2015.

\bibitem[Song et~al.(2019)Song and Ermon]{song2019generative}
Yang Song and Stefano Ermon.
\newblock Generative modeling by estimating gradients of the data distribution.
\newblock \emph{Advances in neural information processing systems}, 32, 2019.

\bibitem[Song et~al.(2020)Song, Sohl-Dickstein, Kingma, Kumar, Ermon, and
  Poole]{song2020score}
Yang Song, Jascha Sohl-Dickstein, Diederik~P Kingma, Abhishek Kumar, Stefano
  Ermon, and Ben Poole.
\newblock Score-based generative modeling through stochastic differential
  equations.
\newblock \emph{arXiv preprint arXiv:2011.13456}, 2020.

\bibitem[Team et~al.(2024)Team, Riviere, Pathak, Sessa, Hardin, Bhupatiraju,
  Hussenot, Mesnard, Shahriari, Ram{\'e}, et~al.]{team2024gemma}
Gemma Team, Morgane Riviere, Shreya Pathak, Pier~Giuseppe Sessa, Cassidy
  Hardin, Surya Bhupatiraju, L{\'e}onard Hussenot, Thomas Mesnard, Bobak
  Shahriari, Alexandre Ram{\'e}, et~al.
\newblock Gemma 2: Improving open language models at a practical size.
\newblock \emph{arXiv preprint arXiv:2408.00118}, 2024.

\bibitem[Team(2025)]{qwen3technicalreport}
Qwen Team.
\newblock Qwen3 technical report, 2025.
\newblock URL \url{https://arxiv.org/abs/2505.09388}.

\bibitem[Vaswani et~al.(2017)Vaswani, Shazeer, Parmar, Uszkoreit, Jones, Gomez,
  Kaiser, and Polosukhin]{vaswani2017attention}
Ashish Vaswani, Noam Shazeer, Niki Parmar, Jakob Uszkoreit, Llion Jones,
  Aidan~N. Gomez, Lukasz Kaiser, and Illia Polosukhin.
\newblock Attention is all you need.
\newblock \emph{Advances in Neural Information Processing Systems}, 2017.

\bibitem[Warner et~al.(2025)Warner, Chaffin, Clavi{\'e}, Weller, Hallstr{\"o}m,
  Taghadouini, Gallagher, Biswas, Ladhak, Aarsen, et~al.]{warner2025smarter}
Benjamin Warner, Antoine Chaffin, Benjamin Clavi{\'e}, Orion Weller, Oskar
  Hallstr{\"o}m, Said Taghadouini, Alexis Gallagher, Raja Biswas, Faisal
  Ladhak, Tom Aarsen, et~al.
\newblock Smarter, better, faster, longer: A modern bidirectional encoder for
  fast, memory efficient, and long context finetuning and inference.
\newblock In \emph{Proceedings of the 63rd annual meeting of the association
  for computational linguistics (volume 1: Long papers)}, pages 2526--2547,
  2025.

\bibitem[Wolf et~al.(2019)Wolf, Debut, Sanh, Chaumond, Delangue, Moi, Cistac,
  Rault, Louf, Funtowicz, et~al.]{wolf2019huggingface}
Thomas Wolf, Lysandre Debut, Victor Sanh, Julien Chaumond, Clement Delangue,
  Anthony Moi, Pierric Cistac, Tim Rault, R{\'e}mi Louf, Morgan Funtowicz,
  et~al.
\newblock Huggingface's transformers: State-of-the-art natural language
  processing.
\newblock \emph{arXiv preprint arXiv:1910.03771}, 2019.

\bibitem[Wu et~al.(2025)Wu, Zhang, Xue, Liu, Diao, Zhu, Luo, Han, and
  Xie]{wu2025fast}
Chengyue Wu, Hao Zhang, Shuchen Xue, Zhijian Liu, Shizhe Diao, Ligeng Zhu, Ping
  Luo, Song Han, and Enze Xie.
\newblock Fast-dllm: Training-free acceleration of diffusion llm by enabling kv
  cache and parallel decoding.
\newblock \emph{arXiv preprint arXiv:2505.22618}, 2025.

\bibitem[Wu et~al.(2023)Wu, Fan, Liu, Zheng, Gong, Shen, Jiao, Li, Wei, Guo,
  Duan, and Chen]{wu2023ar}
Tong Wu, Zhihao Fan, Xiao Liu, Hai-Tao Zheng, Yeyun Gong, Yelong Shen, Jian
  Jiao, Juntao Li, Zhongyu Wei, Jian Guo, Nan Duan, and Weizhu Chen.
\newblock Ar-diffusion: Auto-regressive diffusion model for text generation.
\newblock \emph{Advances in Neural Information Processing Systems},
  36:\penalty0 39957--39974, 2023.

\bibitem[Yadan(2019)]{Yadan2019Hydra}
Omry Yadan.
\newblock Hydra - a framework for elegantly configuring complex applications.
\newblock Github, 2019.
\newblock URL \url{https://github.com/facebookresearch/hydra}.

\bibitem[Yang et~al.(2026)Yang, Guo, Zhang, Sahoo, Chen, Vahdat, Mardani, and
  Thickstun]{yang2026continuous}
Zhihan Yang, Wei Guo, Shuibai Zhang, Subham~Sekhar Sahoo, Yongxin Chen, Arash
  Vahdat, Morteza Mardani, and John Thickstun.
\newblock Continuous diffusion scales competitively with discrete diffusion for
  language.
\newblock \emph{arXiv preprint arXiv:2605.18530}, 2026.

\bibitem[Zheng et~al.(2026)Zheng, Gong, Zhang, Chen, Gu, Zhou, Jaitly, and
  Zhang]{zheng2026continuously}
Huangjie Zheng, Shansan Gong, Ruixiang Zhang, Tianrong Chen, Jiatao Gu,
  Mingyuan Zhou, Navdeep Jaitly, and Yizhe Zhang.
\newblock Continuously augmented discrete diffusion model for categorical
  generative modeling.
\newblock In \emph{International Conference on Learning Representations},
  volume 2026, pages 34212--34242, 2026.

\end{thebibliography}

\newpage

\appendix

\setcounter{tocdepth}{2}
\tableofcontents
\newpage

\section{Clock Diffusion Variational Bound}\label{appsec:clock_nelbo}
In this section, we derive the Clock Diffusion variational bound.
We reuse elements from the derivation in \citet{ruhe2024rolling}, which demonstrates how splitting each latent sequence $\ztL$ into active (\textit{awake}) and inactive (\textit{snoozed}) positions simplifies the diffusion loss component of the variational objective.

In the finite-time setting, we have $T \in \mathbb{N}$ latent sequences $\zzerooneL := (\z_{t(i)}^{1:L})_{i=0}^{T-1}$, where we 
let $t: \{0, \ldots, T-1\} \rightarrow [0, 1]$ and $0 = t(0) < t(1) < \ldots < t(T-1) = 1,$ and $s(i) := t(i-1)$, for $i\in\{1,\ldots,T-1\}$.

We start by explicitly stating our assumed true reverse process:
\begin{align}\label{eq:diff_joint_forward}
    q(\zzerooneL \mid \xL) := q(\zoneL \mid \xL) \textstyle\prod_{i=1}^{T-1} q(\zsiL \mid \ztiL, \xL).
\end{align}
To simplify the notation, in what follows, we drop the dependence of $s, t$ on $i$.

Our aim is to optimize the maximum likelihood, or equivalently, the minimum negative log-likelihood, objective.
We apply the standard importance weighting and Jensen's inequality techniques to this objective to derive the diffusion variational bound \citep{sohl2015deep}:
\begin{align}\label{eq:diff_nelbo}
    &\E_{q_{\text{data}}(\xL)}\left[- \log p_\theta(\xL)\right] = \E_{q_{\text{data}}(\xL)}\left[- \log \int_{\zzerooneL} p_\theta(\xL, \zzerooneL) d\zzerooneL\right] \nonumber\\
    &= \E_{q_{\text{data}}(\xL)}\left[- \log \int_{\zzerooneL} \frac{p_\theta(\xL, \zzerooneL)}{q(\zzerooneL\mid \xL)} q(\zzerooneL \mid \xL) d\zzerooneL\right] \nonumber \\
    &\leq \E_{q_{\text{data}}(\xL)}\left[\log q(\zzerooneL\mid\xL) - \log p_\theta(\xL, \zzerooneL) \right] \nonumber \\
    &= \E_{q_{\text{data}}(\xL)}\left[\lossprior + \lossrecons + \lossdiffdisc\right].
\end{align}
Using the factorizations from (\ref{eq:diff_joint_forward}) and those from Section \ref{subsec:clock_param}, the prior loss $\lossprior$, reconstruction loss $\lossrecons$, and finite-time diffusion loss $\lossdiffdisc$ are defined as follows:
\begin{align}
    \lossprior &:= \KL[q(\zoneL \mid \xL) || p_\theta(\zoneL)] \label{eq:prior_loss} \\
    \gL^T_{\text{recons}} &:= \sum_{i=1}^{T-1}\E_{q(\zs^{\Ast} \mid \xL)} \sum_{\ell \in \Cst} -\log p_\theta(\xl \mid \zsAst, \xCt) \label{eq:recon_loss} \\
    \lossdiffdisc &:= \sum_{i=1}^{T-1} \E_{q(\ztL \mid \xL)} \KL\left[q(\zsL \mid\ztL, \xL) || p_\theta(\zsL \mid \ztL, \xCt) \right]. \label{eq:diff_loss_disc}
\end{align}

\textbf{Simplifying the Diffusion Loss}~
We first note that we parameterize the awake transition as a product over positions, with each factor conditioned on all awake latents and previously revealed tokens:
\begin{align*}
    p_\theta(\zs^{\Ast} \mid \zt^{\Ast}, \xCt) := \prod_{\ell \in \Ast} p_\theta(\zsl \mid \zt^{\Ast}, \xCt).
\end{align*}
Together with the deterministic transitions at snoozed positions, this factorization lets us simplify each term in (\ref{eq:diff_loss_disc}):
\begin{align}\label{eq:clock_simplified_kl}
    &\KL\left[q(\zsL \mid \ztL, \xL) || p_\theta(\zsL \mid \ztL, \xCt) \right] = \E_{q(\zsL \mid \ztL, \xL)}\left[-\log \frac{p_{\theta}(\zsL \mid \ztL, \xCt)}{q(\zsL \mid \ztL, \xL)}\right] \nonumber \\
    &= \E_{q(\zsL \mid \ztL, \xL)}\left[\sum_{\ell \in \Ast}-\log \frac{p_{\theta}(\zsl \mid \ztAst, \xCt)}{q(\zsl \mid \ztAst, \xL)} - \underbrace{\sum_{\ell \in \Zst} \log \frac{p_{\theta}(\zsl \mid \ztl)}{q(\zsl \mid \ztl, \xl)}}_{=0}\right] \nonumber \\
    &= \sum_{\ell \in \Ast}\E_{q(\zsl \mid \ztl, \xl)}\left[ -\log \frac{p_{\theta}(\zsl \mid \ztAst, \xCt)}{q(\zsl \mid \ztl, \xl)}\right].
\end{align}
Plugging (\ref{eq:clock_simplified_kl}) into (\ref{eq:diff_loss_disc}), we get:
\begin{align}\label{eq:diff_loss_disc_clock}
    \gL^{T, \text{Clock}}_{\text{diffusion}} = \sum_{i=1}^{T-1}\E_{q(\ztL \mid \xL)} \sum_{\ell \in \Ast} \KL[q(\zsl \mid \ztl, \xl) || p_\theta(\zsl \mid \ztAst, \xCt)].
\end{align}

\section{Clock Diffusion Language Models}\label{appsec:clock_dlms}
In this section we build on our framework to create a family of models that use Gaussian noise applied to token embeddings and leverage the parameterization and training recipes from RePlaid \citep{yang2026continuous} in conjunction with Clock Diffusion.
When using the sliding window formulation, we call this family of models Clock Diffusion Language Models (ClockDLMs).
We also define block versions that we denote as B-ClockDLMs.

\subsection{Clock Diffusion with Gaussian Noise}\label{appsubsec:clock_gaussian}
When adding Gaussian noise, we have the following functional form of the forward corruption process and true posterior \citep{sohl2015deep}:
\begin{align}
    q(\ztl \mid \xl) = \N(\at \xl, \st^2\mathbf{I}) \quad \text{and} \quad q(\zsl \mid \ztl, \xl) = \gN\left( \tfrac{\ats\ss^2}{\st^2}\ztl + \tfrac{\as\sts^2}{\st^2}\xl, \tfrac{\sts^2\ss^2}{\st^2}\mathbf{I} \right),
\end{align}
where $\ats := \at / \as, \sts^2 := \st^2 - \ats^2\ss^2$, and $\at := \alpha(t)$ and $\st := \sigma(t)$ are monotonically decreasing and increasing functions in $t$, respectively, that dictate the amount of signal and noise, respectively, present in a given latent variable.
Thus we have $\zzeroL$ as the least noisy latent sequence and $\zoneL$ as the most noisy.
An important quantity of interest is the signal-to-noise ratio, which we define as $\snr(t) := \at^2 / \st^2$.

In Clock Diffusion, we adjust all of the noise schedule quantities to be position dependent, that is: $\alt := \alpha(\ult), \slt := \sigma(\ult),$ and $\snrl(t) := \snr(\ult) = \alpha(\ult)^2 / \sigma(\ult)^2$.

With this, each KL term in (\ref{eq:diff_loss_disc_clock}) becomes \citep{kingma2021variational}:
\begin{align}
    \KL[q(\zsl \mid \ztl, \xl) || p_\theta(\zsl \mid \ztAst, \xCt)] = \tfrac{1}{2}(\snrl(s) - \snrl(t)) \,\| \xl -  \xl_\theta(\ztAst, \xCt)\|_2^2.
\end{align}

Finally, we take as our prior $p_\theta(\zoneL) := \prod_{\ell=1}^L \gN(\mathbf{0}, \mathbf{I})$.
Using $\ul(1) = 1,$ for all $\ell$, the prior loss, as a KL between two Gaussian distributions, has a closed-form solution:
\begin{align}
    \lossprior = \frac{1}{2}\sum_{\ell=1}^{L} \left[ \alpha(1)^2 \|\xl\|_2^2 + d\left(\sigma(1)^2 - 1 - \log \sigma(1)^2\right)\right].
\end{align}

\subsection{Learned Embedding Parameterizations}\label{appsubsec:plaid_nelbo}
In our experiments, we choose to instantiate continuous diffusion for discrete data using Gaussian noise applied to token embeddings, as in \citet{li2022diffusion, dieleman2022continuous, gulrajani2023likelihood, lee2026flow, yang2026continuous}.
In particular, let $E \in \reals^{\vocabsize \times d}$ denote an embedding matrix that maps one-hot tokens and probability vectors in the simplex to $d$-dimensional token embeddings.
As described in Section \ref{sec:background}, we let $\xl \in \{0, 1\}^\vocabsize \cap \Delta^\vocabsize$ denote a one-hot vector and $\el := E^\top\xl \in \reals^{d}$ denote its token embedding.
We use the notation $\eL := E[\xL] := (E^\top\x^1, \ldots, E^\top\x^L) \in \reals^{L \times d}$ for a sequence of $L$ embeddings.
Following \citet{gulrajani2023likelihood, yang2026continuous}, we let $\xl_\theta \in \Delta^\vocabsize$ be the softmax output of a neural network that predicts logits over the vocabulary at position $\ell$ and our parameterized prediction of the clean token embeddings is $\eL_\theta := E[\xL_\theta]$, with $\el_\theta := E^\top\xl_\theta$, for all $\ell$.
We normalize each row of $E$ to unit $L^2$ norm.
Since $\xl$ is one-hot, $\|\el\|_2=1$.
The predicted embedding $\el_\theta$ is a convex combination of normalized embedding rows, so $\|\el_\theta\|_2\leq 1$.

Our equations from Appendix \ref{appsubsec:clock_gaussian} are updated as follows:
\begin{align}
    &q(\ztl \mid \xl) = \N(\at \el, \st^2\mathbf{I}), \qquad q(\zsl \mid \ztl, \xl) = \gN\left( \tfrac{\ats\ss^2}{\st^2}\ztl + \tfrac{\as\sts^2}{\st^2}\el, \tfrac{\sts^2\ss^2}{\st^2}\mathbf{I} \right), \\
    &\KL[q(\zsl \mid \ztl, \xl) || p_\theta(\zsl \mid \ztAst, \xCt)] = \tfrac{1}{2}(\snrl(s) - \snrl(t)) \,\| \el -  \el_\theta\|_2^2, \\
    &\text{and} \qquad \lossprior = \frac{1}{2}\sum_{\ell=1}^{L} \left[ \alpha(1)^2 + d\left(\sigma(1)^2 - 1 - \log \sigma(1)^2\right)\right].
\end{align}
Note that $\zsl, \ztl \in \reals^d$ and that $\mathbf{I}$ is a $d \times d$ identity matrix.

Finally, the total optimizable parameters consist of both neural network parameters $\theta$ and the learned embedding matrix $E$.

\subsection{Variational Clock Diffusion: Continuous-time Limit}
As in previous works \citep{kingma2021variational, sahoo2024simple, sahoo2024diffusion}, we can explore the variational lower bound in the continuous-time limit of the diffusion process, $T \rightarrow \infty$.
We now examine each of the loss terms from (\ref{eq:diff_nelbo}) in this limit.

\textbf{Prior Loss}~
First, note that $\lossprior$ is unaffected by this limit, since it only depends on the noise schedule values at the noisy endpoint $\alpha(1)$ and $\sigma(1)$.

\textbf{Reconstruction Loss}~
For the reconstruction loss, we show that there exists a finite grid on which the reconstruction loss is invariant under further refinement.

Define each position’s completion time as
$\tau_\ell := \max \{ t : \ult = 0 \},$ i.e., the time at which position $\ell$ completes denoising as global time decreases from $1$ to $0$.
Now, for any time $t \in [0, 1)$ define: 
\begin{align}
    \gA_{t^+} := \lim_{\varepsilon\downarrow0} \gA_{t\,|\,t+\varepsilon} \qquad \text{and} \quad \gC_{t^+} := \lim_{\varepsilon\downarrow0} \gC_{t+\varepsilon}.
\end{align}
These limits exist because these sets are nested as $\varepsilon$ decreases and contain only finitely many positions, i.e., they are subsets of $\{1, \ldots, L\}$; consequently, they are constant for all sufficiently small positive $\varepsilon$.

Suppose we have a grid of $T$ latents such that for all $\ell \in \{1, \ldots, L\}$, $t(i) = \tau_\ell$, for some $i \in \{0, \ldots, T-1\}$.
Then for any refinements of this finite grid containing every completion time, with the maximum interval length tending to zero, the reconstruction loss will converge to a stable value that is equal to the following:
\begin{align}
    \gL^\infty_{\text{recons}} = \sum_{\ell=1}^{L} \mathbb E_{ q(\mathbf z^{\mathcal A_{\tau_\ell^+}}_{\tau_\ell} \mid\mathbf x^{1:L}) } \left[ -\log p_\theta\!\left( \mathbf x^\ell \;\middle|\; \mathbf z_{\tau_\ell}^{\mathcal A_{\tau_\ell^+}}, \mathbf x^{\mathcal C_{\tau_\ell^+}} \right) \right].
\end{align}

Intuitively this loss is simply a sum of the per-token cross-entropy losses for our distribution over the token $\xl$ given any active tokens and any clean token that previously completed denoising at the time when position $\ell$ finishes denoising.

\textbf{Diffusion Loss}~
For the diffusion loss, our derivation follows directly from \citet{kingma2021variational} and \citet{ruhe2024rolling}, and we have:
\begin{align}
    \gL^{\infty, \text{Clock}}_{\text{diffusion}} := \int_{t=0}^{t=1}\left[\E_{q(\ztL \mid \xL)} \sum_{\ell \in \gA_{t^+}} -\tfrac{1}{2}\snr^\prime(\ult) \, \dot{u}^\ell(t) \,\| \el -  \el_\theta(\zt^{\gA_{t^+}}, \xCt)\|_2^2\right] dt,
\end{align}
where $\snr^\prime$ and $\dot{u}^\ell$ denote the time derivatives of $\snr$ and $\ul$, respectively.
We assume that the local clocks are absolutely continuous and the base SNR schedule is continuously differentiable for these quantities to exist, almost surely for $t \in [0, 1]$.

\subsection{Inherited RePlaid Components}\label{appsubsec:training_replaid}
In addition to using Gaussian noise on token embeddings, ClockDLMs leverage several of the training techniques from RePlaid \citep{yang2026continuous}, a recent promising continuous DLM.

\textbf{Variance-Preserving Latents and $\gamma$ Conditioning}~
We adopt the variance-preserving formulation, i.e., $\at^2 + \st^2 = 1$, and we use the negative log-SNR parameterization of the noise schedule, $\gamma_t := \gamma(t) := -\log \snr(t)$, so that
\begin{align}\label{eq:alpha_sigma_gamma}
    \at = \sqrt{\mathrm{sigmoid}(-\gamma_t)}, \qquad \st = \sqrt{1 - \at^2} = \sqrt{\mathrm{sigmoid}(\gamma_t)}.
\end{align}
We provide $\gamma_t$ as the noise conditioning variable to the denoising network.
Unlike RePlaid, $\gamma_t$ varies per position for our models, so we use $\gamma_t^{1:L} := \big(\gamma(t^{\ell})\big)_{\ell=1}^{L}$ to denote per-position noise conditioning.

\textbf{Learned Noise Schedule}~
As in \citet{yang2026continuous}, $\gamma_t$ is a learnable monotone function of $t$ that is trained jointly with the model to reduce the variance of the variational objective, and we use their efficient variance-reduction estimator (see Figure 8 in \citet{yang2026continuous}): the gradient reaching the schedule's shape parameters is rescaled so that the schedule minimizes the second moment, rather than the mean, of the diffusion loss.
We consider two monotone parameterizations of $\gamma_t$. The first is the monotone network of \citet{kingma2021variational} used by RePlaid, with positively constrained weights.
The second, which we found more stable on our conditional tasks, controls the schedule through its slope.
Specifically, fix $K$ equally spaced times spanning $[0,1]$ and learn a value of $\log m(t)$ at each one.
We linearly interpolate these values between neighboring times and exponentiate the result to obtain a positive function $m(t)$.
We then define
\begin{align}
    \gamma_t = \gamma_0 + (\gamma_1-\gamma_0) \frac{\int_0^t m(u)\,du}{\int_0^1 m(u)\,du}.
\end{align}

The endpoints $\gamma_0, \gamma_1$ can also be learned jointly with the shape.
For some settings we fix these endpoints and for some we allow them to be learned, as in RePlaid.
These schedule settings are reported in Appendix \ref{appsec:exp_setup}.

\textbf{Embedding Bottleneck}~
Following RePlaid, we represent each token by a learned embedding of dimension $d=16$ and apply Gaussian diffusion in this embedding space.
We normalize each row of the embedding matrix $E$ to unit $L^2$ norm.
Before passing a corrupted embedding $\ztl$ to the Transformer, we divide it by $\sqrt{(\alt)^2/d+(\slt)^2}$.
This keeps its expected squared norm equal to $d$ across noise levels.
We then apply a learned linear projection without bias to map the $d$-dimensional latent to the Transformer's hidden dimension.
Clean context embeddings are multiplied by $\sqrt d$ and passed through the same projection.

We use separate optimizer parameter groups for $E$ and the input projections, as specified in Appendix \ref{appsec:exp_setup}.

\textbf{Self-Conditioning}~
We employ self-conditioning \citep{chen2022analog}, where during inference, the denoising model is also conditioned on the output from the previous denoising forward model evaluation.
During training, we apply self-conditioning to a deterministically selected quarter of the examples in each global batch.
For these examples, we first run the denoiser without self-conditioning under \texttt{no\_grad} and compute its posterior-mean clean embedding $\eL_\theta=E[\xL_\theta]$, without re-normalizing its length.
We then use this estimate to condition a second forward pass on the same corrupted input, diffusion times, and attention mask, and compute the training loss from that second pass.
For these training examples, we stop gradients through the input embeddings (both corrupted token embeddings $\ztL$ and clean context embeddings) and the schedule quantities.
The second pass still trains the denoiser and its input projections.

The self-conditioning input is zero on the clean half of the $2L$ canvas and on examples not selected for self-conditioning.
On batch samples that receive self-conditioning, we supply the estimate only to the noisy half.
For conditional tasks, any fixed conditioning tokens present in that half use their known clean embeddings instead of model predictions.

Validation uses the same two-pass procedure for every example.

At inference, generated positions start with zero self-conditioning and subsequent denoising steps reuse the preceding clean embedding estimate without an additional bootstrap pass.

We implement self-conditioning as in RePlaid by multiplying the self-conditioning embedding by \(\sqrt d\), applying a separate bias-free linear projection to the Transformer width, and adding it to the projected input.
This projection is initialized to zero.

\textbf{Batch Splitting}~
During training, evaluating both diffusion and reconstruction loss terms on every example would require two backbone forwards per example, plus any self-conditioning bootstrap forwards.
Instead, as in RePlaid (see Algorithm 1 in \citet{yang2026continuous}), during training, we split each global batch into disjoint subsets, evaluating the diffusion term on one subset and the reconstruction term on the other.
We allocate the split as follows: with $n$ examples in the global batch, we assign
\begin{align}\label{eq:batch_split}
    n_{\mathrm{recons}} = \mathrm{round}\!\left(n \cdot \frac{\hat\sigma_{\mathrm{recons}}}{\hat\sigma_{\mathrm{recons}} + \hat\sigma_{\mathrm{diffusion}}}\right)
\end{align}
examples to the reconstruction term, clipped to $[1, n-1]$, where $\hat\sigma_{\text{recons}}$ and $\hat\sigma_{\text{diffusion}}$ are standard deviations estimated from de-biased exponential moving averages (decay $0.997$) of the first and second loss moments of reconstruction and diffusion loss terms, respectively.
Each subset's contribution is rescaled by the inverse of its sampling fraction so that the resulting estimate of the full objective remains unbiased.

\textbf{Output Logit Prior}~
We also use the `output logit prior' parameterization proposed in \citet{gulrajani2023likelihood} and adopted in \citet{yang2026continuous}: rather than applying softmax directly to the neural network's logits, we add to these logits the analytic Gaussian log-likelihood of the current latent under each vocabulary embedding.
Concretely, let $h_\theta^\ell \in \reals^{\vocabsize}$ denote the raw logits emitted by the denoiser at position $\ell$.
We form the prior logits
\begin{align}\label{eq:output_prior}
    \pi_t^{\ell}[v] := \log \N\!\left(\ztl;\, \alt E_{v, :},\, (\slt)^2 \mathbf{I}\right)
    \;\stackrel{c}{=}\;
    \frac{\alt}{(\slt)^2}\big\langle \ztl,\, E_{v, :} \big\rangle
    - \frac{(\alt)^2}{2(\slt)^2}\big\|E_{v, :}\big\|_2^2,
    \qquad v \in \{1, \ldots, \vocabsize\},
\end{align}
where $\stackrel{c}{=}$ denotes equality up to an additive term that is constant in $v$ and therefore absorbed by the softmax.
Because we $L^2$-normalize the embedding rows, the second term in (\ref{eq:output_prior}) is constant across the vocabulary and can be dropped, so the prior reduces to the scaled inner product $\pi^\ell_t = \frac{\alt}{(\slt)^2}\, E \ztl$.

\textbf{Soft-capping}~
One notable departure from RePlaid's formulation is that we apply soft-capping \citep{team2024gemma} to our logits.
Thus, combined with the output logit prior from the paragraph above, our final softmax output takes the following form:
\begin{align}\label{eq:residual_logits}
    \xl_\theta := \softmax\!\left(\kappa \tanh\!\left(h_\theta^{\ell} / \kappa\right) + \pi_t^{\ell}\right), \qquad \kappa = 30.
\end{align}

\subsection{Block Clock Diffusion Language Models}\label{appsubsec:block_clock_dlms}
Here we present a block instantiation of Clock Diffusion.
As noted in Section \ref{subsec:special_clocks}, we need to specify the per-position phase, and then the local clocks, training mask, and sampling follow accordingly.
Specifically, for a sequence of $L$ tokens, we can split the sequence into $B$ non-overlapping blocks of $S := L/B$ tokens.
Letting $b(\ell) := 1 + \lfloor(\ell - 1) / S\rfloor$ be the block index of position $\ell$, we have:
\begin{align}\label{eq:block_clock}
    \varphi_{\textsc{B}}(\ell) := B - b(\ell) \implies u_{\textsc{B}}^\ell(t) := \operatorname{clip}_{[0,1]}\left( B(t - 1) + b(\ell) \right).
\end{align}
$\varphi_{\textsc{B}}(\ell)$ can be interpreted as the number of blocks that come after $\ell$'s own block.
Under this formulation, the phase gap for two tokens simply becomes their difference in block index:
\begin{align*}
    \Gv(\ell, \ellp) = (B - b(\ellp)) - (B - b(\ell)) = b(\ell) - b(\ellp)
    \begin{cases}
    \geq 1 &\ellp \text{ decodes earlier than } \ell, \\
    = 0 &\ell, \ellp \text{ are in the same block}, \\
    \leq -1 &\ellp \text{ decodes later than } \ell.
    \end{cases}
\end{align*}

Consequently, our training mask becomes the one depicted in Figure \ref{fig:attention_masks}.
For sampling, $r = B$, so Algorithm \ref{alg:clock_sampling} runs for $T = BN$ model calls, with one block awake at any time.
Every $N$ steps, an entire block completes in lockstep and is added to the KV cache.
With this formulation, vectorized training and sampling coincide with the methods of \citet{arriola2025block}.

Now, using the block clocks defined in (\ref{eq:block_clock}), we show what each loss term becomes for the block instantiation of ClockDLMs.
First, for a sequence $\xL$, for each $b \in \left\{1, \ldots, B\right\}$, let $ \x^{b} := \x^{(b-1)S+1:bS}$ and $\x^{<b} := \x^{1:(b-1)S}$ denote block $b$ and its preceding clean context, respectively.

Under the block clocks, for
$ t\in\left(\frac{B-b}{B},\frac{B-b+1}{B}\right),$ the active and completed sets are:
\begin{align}
    \mathcal A_{t^+} = \{(b-1)S+1,\ldots,bS\}, \qquad \mathcal C_t = \{1,\ldots,(b-1)S\}.
\end{align}

All positions in the active block share local time
$u=B(t-1)+b,$ which has $\dot u^\ell(t)=B.$

\textbf{Prior Loss}~
The prior loss is unaffected by the block instantiation, and we have
\begin{align}
    \mathcal L_{\mathrm{prior}} = \KL\left[q(\zoneL \mid \xL) || p_\theta(\zoneL) \right] = \sum_{b=1}^{B} \KL\left[ q(\z_1^{b} \mid \x^{b}) || p_\theta(\z_1^{b}) \right]. 
\end{align}
The equality follows from the factorization of both $q(\zoneL \mid \xL)$ and $p_\theta(\zoneL)$ across blocks and the additivity of KL divergence.

\textbf{Reconstruction Loss}~
For reconstruction, every position in block $b$ completes at global time $(B-b)/B$.
At this boundary, the incoming active set is precisely block $b$, its local time is zero, and its clean context consists of the preceding blocks.
Consequently,
\begin{align}
    \gL_{\mathrm{recons}}^{\infty,\mathrm{Block}} = \sum_{b=1}^{B} \E_{q(\z^b_{(B-b)/B} \mid \x^{b})} \left[ -\log p_\theta\!\left( \x^{b} \mid \z^b_{(B-b)/B}, \x^{<b} \right) \right].    
\end{align}

\textbf{Diffusion Loss}~
For the diffusion loss, partition the global-time integral into the \(B\) active-block intervals and change variables within each interval:

$$ u=B(t-1)+b, \qquad du=B\,dt. $$

The factor \(B\) from the clock derivative therefore cancels the factor \(1/B\) from the integration measure, giving:
\begin{align}
    \mathcal L_{\mathrm{diffusion}}^{\infty,\mathrm{Block}} = \sum_{b=1}^{B} \int_0^1 -\frac{1}{2} \snr^\prime(u) \E_{q(\z_{(u+B-b)/B}^b \mid \x^{b})} \left[ \left\| \e^{b} - \e_\theta^{b} \left(\z_{(u+B-b)/B}^b,\x^{<b}\right) \right\|_F^2 \right]du,
\end{align}
where $\|\cdot\|_F$ represents the Frobenius norm.

\textbf{Comparing to BD3LM}~
BD3LM \citep{arriola2025block} begins from a different perspective, namely that we first factorize the generative model into blocks:
\begin{align}
    p_\theta(\xL) = \textstyle\prod_{b=1}^{B} p_\theta \left( \x^{b} \mid \x^{<b} \right),
\end{align}
and then apply diffusion within each block, using the same priors, reverse kernels, and reconstruction decoders.
Our Clock Diffusion perspective recovers the BD3LM approach where masked diffusion is replaced with Gaussian corruption.

\subsection{Sliding Window Clock Diffusion Language Models}\label{appsubsec:sw_clock_dlms}
Using the sliding window clocks defined in (\ref{eq:window_clock}) in Section \ref{subsec:special_clocks}, we now derive the corresponding loss terms.
Recall that the local clocks are
\begin{align}
    u_\textsc{SW}^\ell(t) := \operatorname{clip}_{[0,1]}\left(
        \frac{(L+S-1)t+\ell-L}{S}
    \right),
\end{align}
where $S$ is the maximum active-window size.
We write $\x^{<\ell} := \x^{1:\ell-1}$ for the clean prefix preceding position $\ell$.

\textbf{Prior Loss}~
Every position still satisfies $\ul(1)=1$.
Thus, the forward endpoint distribution and the standard Gaussian prior are identical to those of Block Clock Diffusion, and the prior loss is unchanged:
\begin{align}
    \lossprior
    = \KL\left[
        q(\zoneL \mid \xL)
        \,\middle\|\,
        p_\theta(\zoneL)
    \right].
\end{align}

\textbf{Reconstruction Loss}~
Position $\ell$ completes denoising at global time
\begin{align}
    \tau_\ell = \frac{L-\ell}{L+S-1}.
\end{align}
At this boundary, the incoming active set and previously completed positions are
\begin{align}
    \gA_{\tau_\ell^+}
    &= \left\{\ell,\ldots,\min(\ell+S-1,L)\right\},
    \qquad
    \gC_{\tau_\ell^+}
    = \left\{1,\ldots,\ell-1\right\}.
\end{align}
The local times within this incoming window satisfy
\begin{align}
    u^j(\tau_\ell)
    = \frac{j-\ell}{S},
    \qquad j \in \gA_{\tau_\ell^+}.
\end{align}
Thus, position $\ell$ has reached local time zero, while subsequent active positions remain at progressively higher noise levels.
Substituting these sets into the Clock Diffusion reconstruction loss gives
\begin{align}
    \gL_{\mathrm{recons}}^{\infty,\mathrm{SW}}
    = \sum_{\ell=1}^{L}
    \E_{q(\z_{\tau_\ell}^{\gA_{\tau_\ell^+}} \mid \xL)}
    \left[
        -\log p_\theta\!\left(
            \xl
            \mid
            \z_{\tau_\ell}^{\gA_{\tau_\ell^+}},
            \x^{<\ell}
        \right)
    \right].
\end{align}
In contrast to Block Clock Diffusion, where reconstruction jointly decodes a block whose positions have all reached local time zero, here, each reconstruction factor decodes one token, conditioned on its clean prefix and an incoming window containing that token and up to $S-1$ partially noisy subsequent positions.

\textbf{Diffusion Loss}~
Position $\ell$ is actively denoising over the global-time interval
\begin{align}
    t \in \left(
        \frac{L-\ell}{L+S-1},
        \frac{L+S-\ell}{L+S-1}
    \right),
\end{align}
where its clock derivative is
\begin{align}
    \dot u^\ell(t) = \frac{L+S-1}{S}.
\end{align}
Substituting this derivative into the general continuous-time diffusion loss gives
\begin{align}
    \gL_{\mathrm{diffusion}}^{\infty,\mathrm{SW}}
    &=
    \frac{L+S-1}{S}
    \int_0^1
    \E_{q(\ztL \mid \xL)}
    \Bigg[
        \sum_{\ell \in \gA_{t^+}}
        -\frac{1}{2}\snr^\prime(\ult)
        \left\|
            \el
            -
            \el_\theta\!\left(
                \zt^{\gA_{t^+}}, \xCt
            \right)
        \right\|_2^2
    \Bigg]dt.
\end{align}
The prediction error is evaluated using the sliding window and clean context present along each position's denoising trajectory.

To express this loss in local time, as in the block derivation, we exchange the integral and finite sum and change variables separately for each position:
\begin{align}
    u &= \frac{(L+S-1)t+\ell-L}{S},
    \qquad
    dt = \frac{S}{L+S-1}\,du.
\end{align}
Each position's active interval maps to $u \in [0,1]$.
The clock derivative therefore cancels against the integration measure.

\section{Additional Training \& Evaluation Details for Clock Diffusion Language Models}\label{appsec:training}

\begin{figure}
\begin{minipage}[t]{0.46\textwidth}
    \vspace{0pt}
    \begin{algorithm}[H]
        \small
        \caption{Block ClockDLM Training}
        \label{alg:training_block}
        \begin{algorithmic}[1]
        \Require data $\xL$, embedder $E$, noise schedule $\gamma$, forward process $q$, $B$ blocks of size $S := L/B$, $\textsc{Denoiser}$ with parameters $\theta$, loss $\gL$
        \State $t^{b} \sim \mathcal{U}[0, 1], \forall b \in \{1, \dots, B\}$ \Comment{one clock per block}
        \State $t^{\ell} \gets t^{b(\ell)}, \forall \ell \in \{1, \dots, L\}$
        \State $\eL \gets E[\xL]$
        \State $\ztL \sim q\big(\cdot \mid \eL, t^{1:L}\big)$
        \State $\textsc{logits} \gets \textsc{Denoiser}\big(\eL \oplus \ztL,\; \gamma_{0}^{1:L} \oplus \gamma_{t}^{1:L},\; \gM_{\textsc{B}}\big)$
        \State Take a gradient step on $\nabla_{E, \theta} \gL(\textsc{logits}; E, \theta)$
        \end{algorithmic}
    \end{algorithm}
\end{minipage}
\hfill
\begin{minipage}[t]{0.53\textwidth}
    \vspace{0pt}
    \begin{algorithm}[H]
        \small
        \caption{Sliding Window ClockDLM Training}
        \label{alg:training_sw}
        \begin{algorithmic}[1]
        \Require data $\xL$, embedder $E$, noise schedule $\gamma$, forward process $q$, window size $S$, groups $\gW_b := \{(b-1)S+1, \ldots, bS\}$, $\textsc{Denoiser}$ with parameters $\theta$, loss $\gL$
        \State $t^{\ell} \sim \mathcal{U}[0, 1], \forall \ell \in \{1, \dots, L\}$ \Comment{one clock per position}
        \State $(t^{\ell})_{\ell \in \gW_b} \gets \textsc{SortAscending}\big((t^{\ell})_{\ell \in \gW_b}\big), \forall b$
        \State $\eL \gets E[\xL]$
        \State $\ztL \sim q\big(\cdot \mid \eL, t^{1:L}\big)$
        \State $\textsc{logits} \gets \textsc{Denoiser}\big(\eL \oplus \ztL,\; \gamma_{0}^{1:L} \oplus \gamma_{t}^{1:L},\; \gM_{\textsc{SW}}\big)$
        \State Take a gradient step on $\nabla_{E, \theta} \gL(\textsc{logits}; E, \theta)$
        \end{algorithmic}
    \end{algorithm}
\end{minipage}
\end{figure}

Throughout this section, we use $\oplus$ to denote concatenation along the sequence axis and $\gM_{\textsc{B}}, \gM_{\textsc{SW}}$ for the block and sliding window training masks of Figure \ref{fig:attention_masks}, respectively.

Note that the diffusion loss training steps presented in Algorithms \ref{alg:training_block} and \ref{alg:training_sw} do not depict self-conditioning.

\subsection{Block Training}\label{appsubsec:training_block}
When using the local clocks from (\ref{eq:block_clock}) that enable block SAR continuous diffusion, our training algorithm closely resembles that proposed in \citet{arriola2025block} for block discrete diffusion.
For completeness, in Algorithm \ref{alg:training_block}, we present the procedure for computing the diffusion loss.
Every position in a block shares a single diffusion time, so we draw $B$ times per sequence, one per block, and broadcast each to the $S$ positions of its block.
The mask $\gM_{\textsc{B}}$ is the one depicted in Figure \ref{fig:attention_masks}: clean-to-clean attention is \emph{block} causal, a noisy query attends to clean keys strictly preceding the start of its own block, and noisy positions attend bidirectionally within their block.
Note that for the time conditioning on the clean half of the canvas we use the clean endpoint $\gamma_0$.

\subsection{Sliding Window Training \& Evaluation}\label{appsubsec:training_sw}
In Algorithm \ref{alg:training_sw}, we present the method for efficiently computing diffusion loss at each position for Sliding Window Clock Diffusion.
As in block training, we divide the noisy latent sequence into disjoint groups of $S$-width token windows.
Each window corresponds to a different awake set.
However, the Sliding Window Clock Diffusion training algorithm differs from that of Block Clock Diffusion in two ways: 1) the vectorized training attention mask has a different clean-to-clean attention pattern, and 2) diffusion times need to be unique per position and are not shared within an active window.

\textbf{Modified Attention Mask}~
Concretely, $\gM_{\textsc{SW}}$ is identical to $\gM_{\textsc{B}}$ except that clean-to-clean attention becomes \emph{token} causal; the noisy-to-clean and noisy-to-noisy quadrants are unchanged, so noisy positions still attend to one another within a width-$S$ window and to clean keys strictly preceding that window.

\textbf{Non-Uniform Time per Active Window}~
In block training, every position of a block shares one diffusion time, so a single draw per block suffices.
In sliding window generation, each position in the active window of $S$ tokens is at a different time in its denoising trajectory.
The schedule in (\ref{eq:window_clock}) would space those times exactly $1/S$ apart, but we deliberately train on \emph{random} spacings instead.
We do so to enable Cache Grab samplers (Section \ref{subsec:sampling_acceleration}): since these samplers commit a variable number of positions per step, the window advances by a variable stride and adjacent positions can end up separated in time by non-fixed-width intervals.
We therefore draw $L$ times $t^{1:L} \sim \mathcal{U}[0,1]^L$, one per position, and sort them in ascending order within each of the disjoint length-$S$ groups $\gW_b := \{(b-1)S+1, \ldots, bS\}$, for $b \in \{1, \ldots, L/S\}$.
Sorting within groups of $S$ tokens preserves the property that within a sliding window local time increases from left to right.

\textbf{Proxy Reconstruction Loss for Training}~
The sliding window reconstruction term derived above,
\begin{align}\label{eq:sw_recons_recall}
    \gL_{\mathrm{recons}}^{\infty,\mathrm{SW}}
    = \sum_{\ell=1}^{L}
    \E_{q(\z_{\tau_\ell}^{\gA_{\tau_\ell^+}} \mid \xL)}
    \left[-\log p_\theta\!\left(\xl \mid \z_{\tau_\ell}^{\gA_{\tau_\ell^+}}, \x^{<\ell}\right)\right],
\end{align}
asks, for each position $\ell$, that the $S-1$ positions after $\ell$ be held at the \emph{staggered} local times $u^{j}(\tau_\ell) = (j - \ell)/S$ while $\ell$ itself sits at local time zero.
Each of the $L$ terms in the summation in (\ref{eq:sw_recons_recall}) therefore imposes a different noise pattern, and evaluating all of them exactly would require $L$ forward passes per sequence.

\emph{At training time}, we optimize a cheap proxy instead.
We place \emph{every} position at the clean endpoint simultaneously, i.e., we set $u^{\ell} = 0$ for all $\ell$, corrupt the whole sequence at $\big(\alpha(0), \sigma(0)\big)$, run a forward pass, plus a bootstrap forward on self-conditioned rows, through the same $2L$ canvas and mask $\gM_{\textsc{SW}}$, and take the token-level cross-entropy $-\log p_\theta\big(\xl \mid \z_{0}^{1:L}, \x^{<\ell}\big)$ at every position.
This is the standard clean-endpoint decoder term of a variational diffusion model, and it is evaluated only on the reconstruction subset of the batch selected by (\ref{eq:batch_split}).
The proxy is evaluated under a strictly cleaner context than the one the sampler provides and does not by itself yield a valid bound on $\gL_{\mathrm{recons}}^{\infty,\mathrm{SW}}$; we use it purely as a training signal for the clean endpoint.

\textbf{Reconstruction Loss at Validation}~
\emph{At validation time}, we replace the proxy with an unbiased Monte Carlo estimate of (\ref{eq:sw_recons_recall}).
For an offset $\rho\in\{0,\ldots,S-1\}$, we partition the sequence into disjoint $S$-wide groups shifted by $\rho$ and write $j(\ell):=(\ell-\rho)\bmod S$ for the offset of position $\ell$ within its group.
We shift the attention mask with these groups, so each noisy query attends to noisy positions in its own group and clean positions strictly before the group's start.
We then corrupt each position at its own local time
\begin{align}\label{eq:commit_ladder}
    u^{\ell} = \frac{j(\ell)}{S},
\end{align}
which reproduces exactly one of the incoming-window patterns from (\ref{eq:sw_recons_recall}): the head of each group (i.e., the position $\ell$ such that $j(\ell) = 0$) sits at local time zero while subsequent positions in that group sit at progressively higher noise levels.
Only the group heads contribute to the cross-entropy loss, and their contribution is multiplied by $S$:
\begin{align}\label{eq:ht_recons}
    \widehat{\gL}_{\mathrm{recons}}^{\infty,\mathrm{SW}}
    := S \sum_{\ell=1}^{L} \1\!\left[j(\ell) = 0\right] \left(-\log p_\theta\!\left(\xl \mid \z^{1:L}, \x^{<\ell}\right)\right).
\end{align}
Note that at sequence boundaries, we retain only in-range positions and include a reconstruction target only if the group head lies within the sequence.

Because $\rho$ is uniform, every position in $\{1, \ldots, L\}$ is a group head with probability $1/S$, so multiplying by $S$ exactly offsets the sub-sampling and $\E_{\rho}\big[\widehat{\gL}_{\mathrm{recons}}^{\infty,\mathrm{SW}}\big] = \gL_{\mathrm{recons}}^{\infty,\mathrm{SW}}$ for every sequence; averaging over validation batches and devices therefore converges to the exact $L$-term sum.

\textbf{Diffusion Loss at Validation}~
The diffusion term requires different noise levels than those used for reconstruction, so we evaluate it separately using a new corrupted input.
We keep the same window partition and draw one random number $U\sim\mathcal U[0,1]$ for each window in each sequence.
We assign each token the local time
\begin{align}
    u^\ell=\frac{j(\ell)+U}{S}.
\end{align}
Within each window, adjacent positions therefore have diffusion times separated by $1/S$.
For example, a full window with $S=4$ has times $U/4,(1+U)/4,(2+U)/4,(3+U)/4$.
If the window offset $\rho$ is sampled uniformly, each token is equally likely to occupy any of the $S$ positions within a window.
Its local time is consequently uniform on $[0,1]$, as required for estimating the diffusion integral.
We evaluate the diffusion loss at every token, using the weighting from the variational bound.

We use self-conditioning for every validation example: for each loss term, a first model pass produces the full posterior-mean clean embeddings.
We feed this prediction into a second pass on the same corrupted input and compute the loss from that second pass.
With the single Monte Carlo sample used in our reported evaluations, computing both reconstruction and diffusion therefore requires four model passes per batch.

\section{Additional Clock Diffusion Sampling Details}\label{appsec:sampling}

\textbf{Denoising Budgets}~
We briefly comment on what the total number of denoising steps looks like under each of the local clock formulations for block and sliding window clocks.

For block generation, from (\ref{eq:block_clock}), we have $r = B$, so Algorithm \ref{alg:clock_sampling} runs for $T = BN$ model calls, with one block awake at any time.
Every $N$ steps, an entire block completes in lockstep and is added to the KV cache.
So on average we decode $S$ tokens every $N$ steps.

For sliding window generation, from (\ref{eq:window_clock}), $r = (L+S-1)/S$, so Algorithm \ref{alg:clock_sampling} runs for $T = (L+S-1)N/S$ model calls and commits one position every $N/S$ calls, rather than $S$ positions every $N$ as in the block instantiation.
Thus, other than at the end of the sequence, both instantiations commit $S$ tokens every $N$ model calls on average.

\textbf{Budgets Below the Window Size: $N < S$}~
The sliding window clocks in (\ref{eq:window_clock}) offset adjacent positions by $1/S$, while each model call advances every clock by $1/N$.
When $N = S$, one position completes per call and Algorithm \ref{alg:clock_sampling} runs for $T = rN = L + S - 1$ calls.
When $N < S$, $rN$ is generally not an integer, so we instead group positions into cohorts of $S/N$ consecutive positions that share one clock.
One cohort is admitted per model call and committed together $N$ calls later.
The window therefore still holds $S$ positions but advances $S/N$ at a time, and sampling takes $\lceil LN/S \rceil + N - 1$ model calls.


\textbf{Fused Caching and Denoising}~
As noted in Section \ref{subsec:clock_sampling}, as written, Algorithm \ref{alg:clock_sampling} requires an additional model forward to cache decoded tokens.
This would be particularly unfavorable for sliding window models that commit a new token to the KV cache on every decoding iteration.
We instead \textit{fuse} this caching operation with the subsequent denoising model forward pass by concatenating the newly decoded tokens with the incoming noisy latent active tokens: $\x^{\Cst} \oplus \zs^{\Ast}$.
To process this concatenated sequence we then require a custom attention mask for the subsequent denoiser model forward.
We depict this mask for both block and sliding window generation in Figure \ref{fig:attention_masks_sampling}.

\begin{figure}[ht]
    \centering
%
%
%
%
%
%
%
\begingroup%
\definecolor{cgMaskBlue}{HTML}{A8C6EE}%
\definecolor{cgMaskGreen}{HTML}{BBDDA8}%
\definecolor{cgMaskOrange}{HTML}{F7C89B}%
\definecolor{cgMaskCut}{HTML}{C76835}%
\definecolor{cgMaskEmpty}{HTML}{ECECEC}%
\definecolor{cgMaskRule}{HTML}{55606E}%
\definecolor{cgMaskInk}{HTML}{25232A}%
\definecolor{cgMaskMuted}{HTML}{667085}%
\def\smP{4}\def\smA{2}%
\begin{tikzpicture}[
  x=0.42cm, y=0.42cm,
  mask label/.style={font=\scriptsize, text=cgMaskInk, inner sep=0.8pt},
  group label/.style={font=\scriptsize, text=cgMaskMuted, inner sep=1pt},
  title/.style={font=\small, text=cgMaskInk, inner sep=1pt},
  legend text/.style={font=\scriptsize, text=cgMaskInk, inner sep=0pt, align=left},
]
%
\newcommand{\smpanel}[5]{%
  \pgfmathtruncatemacro{\nC}{#2}%
  \pgfmathtruncatemacro{\nR}{\nC+\smA}%
  \pgfmathtruncatemacro{\nK}{\smP+\nC+\smA}%
  \foreach \q in {1,...,\nR}{%
    \foreach \k in {1,...,\nK}{%
      \pgfmathtruncatemacro{\qc}{\q<=\nC ? 1 : 0}%
      \pgfmathtruncatemacro{\kg}{\k<=\smP ? 0 : (\k<=\smP+\nC ? 1 : 2)}%
      \pgfmathtruncatemacro{\kc}{\k-\smP}%
      \pgfmathtruncatemacro{\st}{\qc==1 ? (\kg==2 ? 0 : ((\kg==1 && #3==1 && \kc>\q) ? 0 : 1))
                                         : (\kg==2 ? 3 : 2)}%
      \ifcase\st\relax \colorlet{smCur}{cgMaskEmpty}\or \colorlet{smCur}{cgMaskBlue}%
      \or \colorlet{smCur}{cgMaskGreen}\or \colorlet{smCur}{cgMaskOrange}\fi
      \path[fill=smCur, draw=white, line width=0.35pt]
        ({#1+\k-1},{-\q+1}) rectangle ++(1,-1);%
    }%
  }%
  \draw[cgMaskRule, line width=0.7pt] ({#1+\smP},0) -- ({#1+\smP},-\nR);
  \draw[cgMaskRule, line width=0.7pt] ({#1+\smP+\nC},0) -- ({#1+\smP+\nC},-\nR);
  \draw[cgMaskRule, line width=0.7pt] (#1,-\nC) -- ({#1+\nK},-\nC);
  \draw[cgMaskRule, line width=0.7pt] (#1,0) rectangle ({#1+\nK},-\nR);
  \foreach \i in {1,...,\nC}{%
    \pgfmathtruncatemacro{\pos}{\smP+\i}%
    \node[mask label, anchor=east] at ({#1-0.2},{-\i+0.5}) {$\mathbf{x}^{\pos}$};}
  \pgfmathtruncatemacro{\nX}{\smP+\nC}%
  \foreach \k in {1,...,\nX}{%
    \node[mask label, anchor=north] at ({#1+\k-0.5},{-\nR-0.1}) {$\mathbf{x}^{\k}$};}
  \foreach \sub [count=\i] in {#5}{%
    \pgfmathtruncatemacro{\pos}{\smP+\nC+\i}%
    \node[mask label, anchor=east] at ({#1-0.2},{-\nC-\i+0.5}) {$\mathbf{z}_{\sub}^{\pos}$};
    \node[mask label, anchor=north] at ({#1+\smP+\nC+\i-0.5},{-\nR-0.1}) {$\mathbf{z}_{\sub}^{\pos}$};}
  \draw[decorate, decoration={brace, amplitude=2.5pt}, cgMaskMuted, line width=0.4pt]
    ({#1+0.08},0.25) -- ({#1+\smP-0.08},0.25)
    node[group label, midway, above=2pt] {$\KV$};
  \draw[decorate, decoration={brace, amplitude=2.5pt}, cgMaskMuted, line width=0.4pt]
    ({#1+\smP+0.08},0.25) -- ({#1+\smP+\nC-0.08},0.25)
    node[group label, midway, above=2pt] {$\Cst$};
  \draw[decorate, decoration={brace, amplitude=2.5pt}, cgMaskMuted, line width=0.4pt]
    ({#1+\smP+\nC+0.08},0.25) -- ({#1+\nK-0.08},0.25)
    node[group label, midway, above=2pt] {$\Ast$};
  \node[title, anchor=base] at ({#1+\nK/2},1.95) {#4};
  \node[group label, font=\scriptsize\itshape] at ({#1+\nK/2},{-\nR-1.65}) {Keys};
}%
\smpanel{0}{2}{0}{Block}{1,1}
\smpanel{12.2}{1}{1}{Sliding Window}{s,1}
\node[group label, font=\scriptsize\itshape] at (-2.3,-2) [rotate=90] {Queries};
%
\newcommand{\smlegend}[3]{%
  \path[fill=#2, draw=cgMaskRule, line width=0.3pt] (21.4,{#1}) rectangle ++(1,-1);
  \node[legend text, anchor=west] at (22.8,{#1-0.5}) {#3};}
\smlegend{0.6}{cgMaskBlue}{Clean-to-clean}
\smlegend{-0.9}{cgMaskGreen}{Noisy-to-clean}
\smlegend{-2.4}{cgMaskOrange}{Noisy-to-noisy}
\node[group label, font=\scriptsize\itshape, anchor=north west] at (21.3,-3.9) {$S=2$};
\end{tikzpicture}%
\endgroup%
    \caption{Fused caching and denoising attention masks for (\textit{Left}) block and (\textit{Right}) sliding window generation.
    Queries are the newly committed tokens $\x^{\Cst}$ followed by the active noisy tokens $\zs^{\Ast}$; keys are the cache $\KV$ followed by the same tokens.
    Committed tokens never attend to noisy ones, so the keys and values they write to the cache are exactly those a separate caching forward would produce.
    In this example, we assume the KV cache already has 4 tokens in it.
    For the block example, we are adding two newly decoded tokens and therefore starting to denoise 2 new samples from the prior on the subsequent denoising pass.
    For the sliding window example, we are adding one newly decoded token.
    The window on the subsequent pass consists of the token that was previously in the active window as well as a newly admitted sample from the prior.}
    \label{fig:attention_masks_sampling}
\end{figure}

\section{Cache Grab Sampler Algorithms}\label{appsec:cache_grab}    
\subsection{Cache Grab-Confidence}\label{appsubsec:cache_grab_conf}
\textbf{Sampling in Window Form}~
Algorithm \ref{alg:clock_sampling} is written with closed-form clocks: it loops over a fixed grid of $T = rN$ global times and reads the active set off the clocks $\ult$.
Cache Grab samplers decouple committing from the clocks: a position can be committed before its clock reaches zero, so neither the active sets nor the number of model calls follow the closed-form clocks any more.
To detail the Cache Grab sampler algorithms more clearly, we therefore rewrite sampling in terms of the state the implementation keeps.
Namely, we keep track of an ordered window $\gA$ of at most $S$ active positions.
For each position $\ell \in \gA$, we track its `age' $k^\ell$, which corresponds to the number of denoising steps it has received.
We then define the local clock for that position as $u^\ell = 1 - k^\ell/N$.
Algorithm \ref{alg:clock_sampling_window} presents sampling with this new perspective.
Without the Cache Grab samplers and assuming $N = S$, Algorithm \ref{alg:clock_sampling_window} is the same as Algorithm \ref{alg:clock_sampling} for sliding window ClockDLMs.

\begin{algorithm}[H]
    \small
    \caption{ClockDLM Sampling in Window Form (Sliding Window, $N = S$)}
    \label{alg:clock_sampling_window}
    \begin{algorithmic}[1]
    \Require prompt, length $L$, window size $S$, per-position budget $N = S$, prior $q_{\mathrm{prior}}$, \textsc{Denoiser}, $\textsc{Sampler}$, commit rule $\textsc{Commit}$
    \State $\KV \gets \textsc{Cache}(\text{prompt}, \emptyset)$
    \State $\gA \gets \emptyset$ \Comment{$\gA$: tracks active positions, left to right}
    \State $f \gets 0$ \Comment{tracks number of model calls}
    \State $p \gets 1$ \Comment{next position to admit}
    \While{$\gA \neq \emptyset$ or $p \leq L$}
        \While{$|\gA| < \min(f+1, S)$ and $p \leq L$} \Comment{refill; capacity grows by one per step until it reaches $S$}
            \State $\gA \gets \gA \cup \{p\}$
            \State $\z^p \sim q_{\mathrm{prior}}$
            \State $k^p \gets 0$
            \State $p \gets p + 1$
        \EndWhile
        \State $\textsc{logits}^{\gA} \gets \textsc{Denoiser}(\z^{\gA}, u^{\gA}, \KV)$ \Comment{$u^\ell = 1 - k^\ell / N$}
        \State $\z^{\gA} \gets \textsc{Sampler}(\textsc{logits}^{\gA})$
        \State $k^\ell \gets k^\ell + 1$ for all $\ell \in \gA$
        \State $\gC \gets \textsc{Commit}(\gA, \textsc{logits}^{\gA})$ \Comment{under standard decoding, this amounts to: $\{\ell \in \gA \mid k^\ell = N\}$}
        \State $\x^{\gC} \gets \argmax(\textsc{logits}^{\gC})$
        \State $\KV \gets \textsc{Cache}(\x^{\gC}, \KV)$
        \State $\gA \gets \gA \setminus \gC$
        \State $f \gets f + 1$
    \EndWhile
    \State \Return $\xL$
    \end{algorithmic}
\end{algorithm}

Cache Grab-Confidence replaces the commit rule of Algorithm \ref{alg:clock_sampling_window} with Algorithm \ref{alg:cache_grab_conf}.
A position is eligible to be committed if its clock has run out, as in standard decoding, or if the probability the model assigns to its $\argmax$ token is at least a threshold $\eta$.
The committed set is the longest run of eligible positions starting from the left end of the window, so even if some position $\ell$ clears the $\eta$ threshold, it may not be committed if a different position $\ell' < \ell$ does not meet the confidence threshold and its local clock has not run out (see, for example, Figure \ref{fig:cache_grab_sampling}, \textit{Left}).
After we commit the leftmost tokens, the remaining positions keep their latents and ages and shift left, and the next iteration refills the window with samples from the prior.
With $\eta = 1$ the confidence test is disabled and the standard sampler is recovered.

\begin{algorithm}[H]
    \small
    \caption{Cache Grab-Confidence Commit Rule}
    \label{alg:cache_grab_conf}
    \begin{algorithmic}[1]
    \Function{Commit}{$\gA = (\ell_1, \ldots, \ell_m)$, $\textsc{logits}^{\gA}$; threshold $\eta$}
        \For{$i = 1, \ldots, m$}
            \State $e_i \gets \1\!\left[k^{\ell_i} = N\right] \vee \1\!\left[\eta < 1 \wedge \max_v \operatorname{softmax}(\textsc{logits}^{\ell_i})_v \geq \eta\right]$ \Comment{eligible}
        \EndFor
        \State $j \gets \max\{\, i \in \{0, \ldots, m\} \mid e_1 = \cdots = e_i = 1 \,\}$ \Comment{longest eligible prefix}
        \State \Return $\{\ell_1, \ldots, \ell_j\}$
    \EndFunction
    \end{algorithmic}
\end{algorithm}

\textbf{Adapting Cache Grab-Confidence for Block Decoding}~
B-ClockDLMs denoise a whole block in lockstep and commit it at once, so the rule is applied to the block as a unit.
After each model call, if every position being generated in the current block reaches the threshold $\eta$, the block is committed; otherwise the block takes its next denoising step, and after its $N$-th call it is committed.
The committed block is written to the cache in the next block's first model call, with full attention within the block (Figure \ref{fig:attention_masks_sampling}, \textit{Left}), and the next block starts with samples from the prior.

\subsection{Cache Grab-SSD}\label{appsubsec:cache_grab_ssd}
In Cache Grab-SSD, each iteration makes two model calls.
The first is the usual denoising call over the window, from which every active position receives a draft token.
The second call runs the model in AR mode over the drafts, attending causally among the draft tokens.
At every position in the draft window, the AR head provides its prediction given the cache and draft tokens to its left.
The sampler commits the longest prefix of the window on which the draft and the AR prediction agree, together with the AR prediction at the first position where they disagree.
Thus every iteration commits between $1$ and $|\gA|$ tokens.
A position's clock running out never commits it: a position that reaches $u^\ell = 0$ stays there and is drafted again until it is committed, as an accepted draft or as the AR correction.
This maintains the invariant that the produced sequence matches the one we would see under AR decoding alone.

\textbf{Keeping Rejected Positions}~
Self-speculative decoding in discrete diffusion discards the positions it rejects.
Cache Grab-SSD instead keeps every position it does not commit, together with its denoising progress.

\textbf{Staggering Noise in the New Active Window}~
At the start of each iteration, the window is refilled to $S$ positions.
Newly admitted positions start from the prior, at $u^\ell = 1$.
Positions kept from earlier iterations advance by one step per iteration, $u^\ell \gets \max(u^\ell - 1/N, 0)$.
To maintain staggered noise in the active window, if the newly admitted positions are not committed via SSD, their local clocks are spread evenly over the gap between the newest position kept from earlier and $1 - 1/N$.
This ensures that the window stays ordered from clean on the left to noisy on the right.
Concretely, if $u_b$ is the new time of the newest position kept from earlier ($u_b = 0$, if no position from earlier remains in the window), the $i$-th of $n$ surviving new positions, counted from the left, is placed at $u^\ell = u_b + \tfrac{i}{n}(1 - 1/N - u_b)$; the rightmost therefore lands at $1 - 1/N$.

\section{Additional Experimental Setup Details}\label{appsec:exp_setup}

\textbf{Optimization}~
All models trained in our experiments use AdamW \citep{kingma2014adam, loshchilov2017decoupled}.
We clip gradients to a maximum norm of $1.0$ and use a constant learning rate after linear warmup.
For RePlaid and ClockDLMs (as well as B-ClockDLMs), the learnable embedding matrix has a learning rate of $10^{-2}$, while the latent, self-conditioning and time-conditioning input projections have a learning rate of $3\times10^{-4}$; these parameter groups have zero weight decay.
We maintain an exponential moving average (EMA) of the weights with decay $0.9999$ and use these EMA weights for evaluation.
The remaining optimizer settings, batch sizes and training budgets are listed in the task-specific tables.

\textbf{Numerical Precision}~
All models trained in our experiments, including MDLM, BD3LM and SetDLM, use \texttt{bfloat16} autocast for the Transformer blocks and output head, including the vocabulary projection.
Layer normalization and RMS normalization compute their statistics in \texttt{float32}.
Validation loss sums and token counts are accumulated and reduced across devices in \texttt{float64} before computing perplexity.
TF32 is enabled for eligible \texttt{float32} matrix multiplications.

For ClockDLMs and our RePlaid baselines, embedding normalization, latent corruption, the latent and self-conditioning input projections, time embeddings, posterior-mean embedding computations and the analytic output logit prior use \texttt{float32}.
Their learned noise schedules are evaluated in \texttt{float64}.
For their output logit prior in (\ref{eq:output_prior}), we clip $(\slt)^2$ to a minimum of $10^{-8}$.

\textbf{Evaluation Protocol}~
Our evaluations run on a single node with eight \texttt{H100-80GB} GPUs, using seed $=1$.
Generation uses batch size one per GPU; likelihood evaluation uses the batch size specified below.
Our NFE metrics account for all model forwards, including prompt ingestion, cache updates, final readout calls, and AR verification calls for Cache Grab-SSD.
Sampling algorithms, denoising budgets and stopping rules are specified below for each task.

\textbf{Distinguishing Cache Grab-SSD}
All of the results for ClockDLM represent standard diffusion.
Any results reported for Cache Grab-Confidence use these models as well.
To enable Cache Grab-SSD, we \textbf{retrain} separate models with the AR loss enabled.
Only results that are explicitly marked as using Cache Grab-SSD, as well as those in Table \ref{tab:gsm8k_ssd_ablation}, use models trained with joint AR and diffusion loss.

\subsection{OpenWebText}\label{appsubsec:exp_setup_owt}

\textbf{Data}~
We train on OpenWebText \citep{Gokaslan2019OpenWeb}.\footnote{\url{https://huggingface.co/datasets/Skylion007/openwebtext}}
Following \citet{sahoo2024simple, arriola2025block}, the last $100{,}000$ documents of the \texttt{train} split are held out for validation and the remainder is used for training.
Documents are tokenized with the GPT-2 tokenizer \citep{radford2019language}.
We append \texttt{[EOS]} to each document and concatenate the tokenized documents within each pre-processing batch.
The stream is split into chunks of $L-2$ tokens, each wrapped with \texttt{[BOS]} and \texttt{[EOS]} to form a sequence of length $L=1024$.
Incomplete chunks at the end of each pre-processing batch are discarded; the resulting sequences contain no padding.

\textbf{Training}~
For our retrained RePlaid model and ClockDLMs we learn both the noise schedule $\gamma_t$ and the endpoints $\gamma_0$ and $\gamma_1$, following the setup in \citet{yang2026continuous}.
We initialize the endpoints to $\gamma_0 = -5.5$ and $\gamma_1 = 3$.
We use the same monotone MLP as in \citet{kingma2021variational, yang2026continuous} for the learned schedule.
Full training details are provided in Table \ref{tab:owt_setup}.

\textbf{Likelihood Evaluation}~
We report variational perplexity upper bounds on the full held-out validation split, with validation batch size $32$ per GPU.
RePlaid and Block ClockDLMs use the ordinary validation negative log-likelihood bound, while Sliding Window ClockDLMs use the rolling estimator described in Appendix \ref{appsubsec:training_sw}.
We evaluate reconstruction and diffusion separately, using self-conditioning for every validation example.
All evaluations use the $1{,}000{,}000$-step checkpoint and EMA weights.

\textbf{Sampling Evaluation}~
To produce the generative quality results in Appendix \ref{appsubsec:exp_results_mauve}, we use a DDPM sampler.
We also explore nucleus (top-$p$) filtering: at every denoiser call, we keep the smallest set of most probable tokens whose cumulative probability is at least $p$, re-normalize over them, and use the expected embedding under this truncated distribution as the clean-embedding estimate in the DDPM update.

\begin{table}[ht]
    \centering
    \small
    \caption{Experimental configuration for OpenWebText.}
    \setlength{\tabcolsep}{4pt}
    \begin{tabular}{ll}
        \toprule
        \multicolumn{2}{l}{\textit{Data}} \\
        Tokenizer / vocabulary & \texttt{gpt2} / $50{,}258$ \\
        Context length $L$ & $1024$ \\
        Packing & \texttt{[EOS]} per document; \texttt{[BOS]}/\texttt{[EOS]} per packed block \\
        Validation split & Last $100{,}000$ documents \\
        \midrule
        \multicolumn{2}{l}{\textit{Architecture}} \\
        Backbone & DiT \citep{peebles2023scalable} with adaptive layer norm \\
        Layers / hidden / heads & $12$ / $768$ / $12$ (head dim.\ $64$) \\
        Conditioning dim.\ / dropout / MLP ratio & $128$ / $0.1$ / $4$ \\
        Embedding bottleneck $d$ & $16$ ($L_2$-normalized rows, untied output head) \\
        \midrule
        \multicolumn{2}{l}{\textit{Optimization}} \\
        Gradient updates & $1{,}000{,}000$ \\
        Global / per-device batch size & $512$ / $8$ (no gradient accumulation) \\
        GPUs & $64$ ($8$ nodes $\times$ $8$ \texttt{H100-80GB}) \\
        Optimizer & AdamW, $(\beta_1, \beta_2) = (0.9, 0.999)$, $\epsilon = 10^{-8}$, wd $0$ \\
        Learning rate / warmup & $3\times 10^{-4}$, constant after $2{,}500$ linear-warmup steps \\
        Embedding / projection LR & $10^{-2}$ / $3 \times 10^{-4}$ (wd $0$) \\
        Gradient clipping / precision / EMA & $1.0$ / \texttt{bf16} / $0.9999$ \\
        Time sampling & Antithetic, stratified over the global batch \\
        \midrule
        \multicolumn{2}{l}{\textit{\method}} \\
        Window sizes $S$ & $\{4, 8, 16, 32, 64\}$ \\

        $\gamma_t$ parameterization & Monotone MLP \citep{kingma2021variational}, hidden width $1024$ \\
        $[\gamma_{\min}, \gamma_{\max}]$ (initialization) & $[-5.5, 3.0]$, both endpoints learned \\
        Schedule LRs (shape / endpoints) & $10^{-2}$ (wd $0$) / $10^{-2}$ (wd $0.1$) \\
        \bottomrule
    \end{tabular}
    \label{tab:owt_setup}
\end{table}

\subsection{Math Reasoning Benchmark}\label{appsubsec:exp_setup_gsm8k}

\textbf{Data}~
We train on TinyGSM \citep{liu2023tinygsm},\footnote{\url{https://huggingface.co/datasets/TinyGSM/TinyGSM}} a synthetic corpus of grade-school math questions paired with Python programs whose execution produces the answer.
Each example is tokenized as \texttt{[BOS]} $\oplus$ question $\oplus$ \texttt{\textbackslash n} $\oplus$ code $\oplus$ \texttt{[EOS]}; the prompt (\texttt{[BOS]}, question and \texttt{\textbackslash n}) is provided as clean conditioning context and diffusion is applied to the code continuation.
Examples exceeding the $L = 512$ canvas are discarded and shorter examples are right-padded.
For the diffusion models, we include padded positions in the loss, as we found that this led to improved performance.
We evaluate zero-shot on the full GSM8K \citep{cobbe2021training} test set ($1{,}319$ examples).\footnote{\url{https://github.com/openai/grade-school-math}}

\textbf{Training}~
For our retrained RePlaid model and ClockDLMs we learn the noise schedule $\gamma_t$ and keep endpoints fixed at $\gamma_0 = -1$ and $\gamma_1 = 6$.
We use the monotone log-slope schedule model described in Appendix \ref{appsubsec:training_replaid}.
Full training details are provided in Table \ref{tab:gsm8k_setup}.

\textbf{Evaluation}~
Each configuration decodes all $1{,}319$ test problems once with seed $1$ and batch size one per device.
When applicable, we apply early stopping on \texttt{[EOS]}.
For example, full sequence MDLM and RePlaid cannot benefit from this early stopping.
We follow the evaluation protocol from \citet{deschenaux2026language}.
A generated output is scored by extracting the emitted program and executing it.
The returned value is compared against the reference answer with a tolerance of $10^{-3}$.
We report pass@$1$ accuracy.
All evaluations use the $250{,}000$-step checkpoint with EMA weights.

\textbf{Sampling}~
RePlaid, Block ClockDLMs and Sliding Window ClockDLMs use DDPM ancestral sampling with vocabulary top-$1$ at every denoising update, meaning that we sample $\zs \sim q(\zs \mid \zt, \x = \argmax(\x_\theta))$, i.e., each DDPM update therefore uses the embedding of the highest-probability token as its clean-embedding estimate.
The results in Table \ref{tab:gsm8k_accuracy} reflect $N=S$ for our models.
For Block and Sliding Window ClockDLMs we sweep the per-position budget $N\in\{S,S/2,\ldots,2\}$ to produce the accuracy--efficiency trade-offs in Figure \ref{fig:nfe_pareto}.
RePlaid is decoded full-sequence with $T\in\{2,4,\ldots,1024\}$.
The SW-SetDLM curves in Figure \ref{fig:nfe_pareto} instead sweep the confidence threshold.
SAR MDLM decoding refers to the sampling employed by models such as LLaDA \citep{nie2025large}, where at each denoising iteration, only positions from the leftmost block of size $S$ are eligible for decoding.
For BD3LM and MDLM, we use confidence-based decoding, where the top-$k$ most confident masked tokens are generated on every iteration.
We control the denoising budget by varying $k$.
Cache Grab-Confidence evaluations use $N=S$, with the early-commit threshold swept over $\{0.99,0.95,0.9,0.85,0.8,0.75,0.7\}$.
Cache Grab-SSD runs at $N=S$.
For self-conditioning, we use the top-$1$ embedding from the previous denoising iteration.

\begin{table}[ht]
    \centering
    \small
    \caption{Experimental configuration for TinyGSM training and GSM8K evaluation.}
    \setlength{\tabcolsep}{4pt}
    \begin{tabular}{ll}
        \toprule
        \multicolumn{2}{l}{\textit{Data}} \\
        Tokenizer / vocabulary & \texttt{HuggingFaceTB/SmolLM-135M} / $49{,}153$ \\
        Context length $L$ & $512$ (right-padded, overlong examples filtered) \\
        Packing & None; one example per sequence \\
        Prompt / target & \texttt{[BOS]}$\oplus$question$\oplus$\texttt{\textbackslash n} / code$\oplus$\texttt{[EOS]} \\
        Validation split & $1\%$ deterministic holdout of TinyGSM train \\
        \midrule
        \multicolumn{2}{l}{\textit{Architecture}} \\
        Backbone & DiT, identical to Table \ref{tab:owt_setup} \\
        Layers / hidden / heads & $12$ / $768$ / $12$ \\
        Embedding bottleneck $d$ & $16$ \\
        \midrule
        \multicolumn{2}{l}{\textit{Optimization}} \\
        Gradient updates & $250{,}000$ (all reported checkpoints) \\
        Global / per-device batch size & $512$ / $16$ (no gradient accumulation) \\
        GPUs & $32$ ($4$ nodes $\times$ $8$ \texttt{H100-80GB}) \\
        Optimizer & AdamW, $(\beta_1, \beta_2) = (0.9, 0.999)$, $\epsilon = 10^{-8}$, wd $0$ \\
        Learning rate / warmup & $3 \times 10^{-4}$, constant after $2{,}500$ linear-warmup steps \\
        Gradient clipping / precision / EMA & $1.0$ / \texttt{bf16} / $0.9999$ \\
        Time sampling & Antithetic, stratified over the global batch (except SetDLM) \\
        Validation / checkpoint interval & Every $10{,}000$ steps ($500$ validation batches) \\
        \midrule
        \multicolumn{2}{l}{\textit{\method}} \\
        Window sizes $S$ & $\{4, 8, 16, 32, 64\}$ \\
        $\gamma_t$ parameterization & Monotone log-slope, $32$ knots, shape only \\
        $[\gamma_{\min}, \gamma_{\max}]$ & $[-1, 6]$, endpoints frozen \\
        Schedule shape LR & $10^{-4}$ (wd $0$) \\
        Cache Grab-SSD models & AR head tied to the diffusion output layer, loss weight $1.0$ \\
        \midrule
        \multicolumn{2}{l}{\textit{Evaluation}} \\
        Test set & GSM8K test, all $1{,}319$ examples, seed $1$ \\
        Sampler & DDPM, top-$1$ at each update, \texttt{[EOS]} stopping \\
        Denoising budget & $N \in \{S, S/2, \ldots, 2\}$ \\
        Confidence thresholds & $\{0.99, 0.95, 0.9, 0.85, 0.8, 0.75, 0.7\}$ at $N = S$ \\
        \bottomrule
    \end{tabular}
    \label{tab:gsm8k_setup}
\end{table}

\subsection{Summarization}\label{appsubsec:exp_setup_cnndm}

\textbf{Data}~
We use CNN/DailyMail v3.0.0 \citep{DBLP:conf/nips/HermannKGEKSB15, see-etal-2017-get}.\footnote{\url{https://huggingface.co/datasets/abisee/cnn_dailymail}}
Following \citet{arriola2026set}, the source region is \texttt{[BOS]} $\oplus$ article $\oplus$ \texttt{[EOS]}, independently truncated to $512$ tokens, and the target region is \texttt{Summary: } $\oplus$ highlights $\oplus$ \texttt{[EOS]}, independently truncated to $180$ tokens; both regions reserve their final position for \texttt{[EOS]} and are concatenated and right-padded to the $L = 768$ max sequence length.
The article region is provided as clean conditioning context.
We only noise and compute the diffusion loss and the reconstruction loss on the target region, and padded positions are excluded from both the loss and the attention mask.
We select the checkpoint with the lowest validation loss for evaluation.

\textbf{Training}~
For our retrained RePlaid model and ClockDLMs we learn the noise schedule $\gamma_t$ and keep endpoints fixed at $\gamma_0 = -5$ and $\gamma_1 = 3$.
We use the monotone log-slope schedule model described in Appendix \ref{appsubsec:training_replaid}.
Full training details are provided in Table \ref{tab:cnndm_setup}.

\textbf{Evaluation}~
Each configuration generates summaries for all $11{,}490$ test articles once with seed $1$ and batch size one per device.
We adopt the decoding recipe of \citet{arriola2026set}: at most $180$ new tokens are generated, with a minimum of $50$ non-\texttt{[EOS]} tokens generated.
We apply an exponential-decay length penalty with a factor of $1.1$ that starts after $80$ tokens, and a repetition penalty swept over $\{1.0, 1.2\}$.
Logits processors are applied in a fixed order: log-softmax, length penalty, repetition penalty, renormalization.
Where applicable, we use early stopping on the $\texttt{[EOS]}$ token.
We report ROUGE-1/2/L \citep{lin2004rouge} computed with the TorchMetrics implementation without stemming and averaged over examples.

\textbf{Sampling}~
We use the same sampling setup as in the GSM8K experiment.
RePlaid, Block ClockDLMs and Sliding Window ClockDLMs use DDPM ancestral sampling with vocabulary top-$1$ at every denoising update (as in the GSM8K setup).
Fixed conditioning tokens use their known clean embeddings wherever self-conditioning is supplied.
The per-position budget is swept over $N\in\{S,S/2,\ldots,2\}$ and the Cache Grab-Confidence threshold sweep is run at $N=S$; RePlaid is decoded full-sequence.

\begin{table}[ht]
    \centering
    \small
    \caption{Experimental configuration for CNN/DailyMail.}
    \setlength{\tabcolsep}{4pt}
    \begin{tabular}{ll}
        \toprule
        \multicolumn{2}{l}{\textit{Data}} \\
        Tokenizer / vocabulary & \texttt{Qwen/Qwen3-0.6B-Base} / $151{,}669$ \\
        Context length $L$ & $768$ ($512$ source $+$ $180$ target, right-padded) \\
        Target prefix & \texttt{"Summary: "} \\
        Supervision & Target region only; padding neither noised nor scored \\
        \midrule
        \multicolumn{2}{l}{\textit{Architecture}} \\
        Backbone & Qwen3-style decoder (architecture only; trained from scratch) \\
        Layers / hidden / SwiGLU intermediate & $28$ / $256$ / $768$ \\
        Query / key-value heads, head dim. & $16$ / $8$, $128$ \\
        Normalization / positions & RMSNorm ($\epsilon = 10^{-6}$, per-head Q/K norm) / RoPE $\theta = 10^{6}$ \\
        Conditioning dim.\ / biases / tying & $128$ / none / untied \\
        Embedding bottleneck $d$ & $16$ \\
        \midrule
        \multicolumn{2}{l}{\textit{Optimization}} \\
        Gradient updates & $\leq 500{,}000$; best-validation checkpoint selected \\
        Global / per-device batch size & $128$ / $4$ \\
        GPUs & $32$ \texttt{H100-80GB} \\
        Optimizer & AdamW, $(\beta_1, \beta_2) = (0.9, 0.98)$, $\epsilon = 10^{-8}$, wd $10^{-5}$ \\
        Learning rate / warmup & $3 \times 10^{-4}$, constant after $1{,}000$ linear-warmup steps \\
        Gradient clipping / precision / EMA & $1.0$ / \texttt{bf16} / $0.9999$ \\
        Time sampling & Uniform (antithetic sampling disabled) \\
        Validation / checkpoint interval & Every $5{,}000$ steps \\
        \midrule
        \multicolumn{2}{l}{\textit{\method}} \\
        Window sizes $S$ & $\{4, 8, 16, 32, 64\}$ \\
        Endpoint terms & $\lossrecons$ optimized (weight $1.0$); $\lossprior$ evaluated only \\
        $\gamma_t$ parameterization & Monotone log-slope, $32$ knots, shape only, frozen endpoints \\
        $[\gamma_{\min}, \gamma_{\max}]$ & $[-5, 3]$, endpoints frozen \\
        \midrule
        \multicolumn{2}{l}{\textit{Evaluation}} \\
        Test set & CNN/DailyMail test, all $11{,}490$ examples, seed $1$ \\
        Max.\ / min.\ generated tokens & $180$ / $50$ non-\texttt{[EOS]} \\
        Length penalty (start, factor) & $(80, 1.1)$ \\
        Repetition penalty & $\{1.0, 1.2\}$ \\
        Metric & ROUGE-1/2/L, TorchMetrics, no stemming, example mean \\
        \bottomrule
    \end{tabular}
    \label{tab:cnndm_setup}
\end{table}

\section{Additional Experimental Results}\label{appsec:exp_results}

\subsection{Open-Ended Generation}\label{appsubsec:exp_results_mauve}
Using our ClockDLM ($S=64$) model and the retrained RePlaid from the OpenWebText experiment, we evaluate the quality of generated samples.
Specifically, we follow the setup from \citet{yang2026continuous}, where we generate 5,120 sequences and use ModernBERT-large \citep{warner2025smarter} to compute MAUVE scores \citep{pillutla2021mauve, mauve_github}.
For both models we sweep over using a repetition penalty with factor 1.2 and applying top-$p$ nucleus sampling to the softmax output of the denoising network with $p \in \{0.9, 0.95\}$.
For RePlaid, we consistently find the best results using no nucleus sampling and no repetition penalty.
We found ClockDLM to work consistently better when using the repetition penalty.
However, our model was more sensitive to the top-$p$ parameter with improved results for $p=0.95$ when using NFEs $> L$ and improved results for $p=0.9$ when using an NFE budget $< L$.

\begin{figure}
    \centering
    \includegraphics[width=0.5\linewidth]{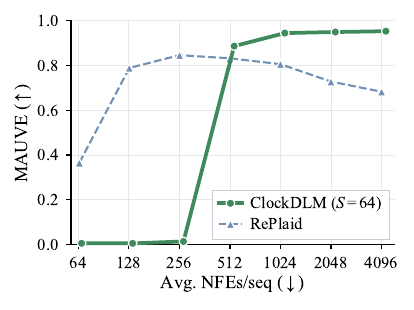}
    \caption{MAUVE scores vs.\ NFEs/seq for sequences generated from models trained on OWT.}
    \label{fig:owt_mauve}
\end{figure}

In Figure \ref{fig:owt_mauve}, we display the best results for each model.
Notably, in higher-NFE regimes $\sim512+$, ClockDLM outperforms RePlaid, whose performance plateaus and even dips after a certain budget.
However, for this task of open-ended generation in the low-NFE regime, our model struggles.
This can be attributed to the fact that in this regime we apply a comparatively much smaller denoising budget per token.
For example, with a total denoising budget of $\sim256$, ClockDLM will apply $\sim16$ denoising steps per token, whereas RePlaid, because it is a full sequence model, uses this same total decoding budget for each token, applying $256$ denoising steps to each token.

\subsection{Cache Grab-Confidence for ClockDLMs \& B-ClockDLMs on GSM8K}\label{appsubsec:exp_results_conf_gsm8k}
Table \ref{tab:gsm8k_accuracy_cache_grab_conf} reports the full Cache Grab-Confidence threshold sweep on GSM8K for ClockDLMs and B-ClockDLMs at $S \in \{32, 64\}$.
For ClockDLMs, lowering the threshold to $\eta = 0.7$ reduces NFEs/seq by $3.2\times$ at $S = 32$ with no loss in accuracy, and by $2.8\times$ at $S = 64$ at a cost of 2.0 points.
B-ClockDLMs, which commit a block only once every position in it clears the threshold, see smaller reductions of $2.7\times$ at $S = 32$ and $2.1\times$ at $S = 64$, with accuracy within one point of standard decoding in both cases.
Lowering the threshold produces the expected efficiency gains of lowering NFEs/seq.
However, interestingly, we do not see a clear monotonic trend in accuracy reduction as the threshold is reduced.
This is likely due to the nature of this task's evaluation metric: the correctness of the executed program that is generated can be significantly altered by even small decoding deviations.

\begin{table}[ht]
    \centering
    \small
    \caption{Cache Grab-Confidence on GSM8K.
    Zero-shot pass@$1$ accuracy (Acc \%; $\uparrow$) on the GSM8K test set and NFEs/seq for ClockDLMs and B-ClockDLMs trained on TinyGSM, sweeping the early-commit threshold $\eta$ at $N = S$.
    NFE values are means $\pm$ sample standard deviations.
    \textit{Standard} rows decode without early commits.}
    \setlength{\tabcolsep}{3pt}
    \begin{tabular}{lllcccc}
        \toprule
         & & & \multicolumn{2}{c}{$S = 32$} & \multicolumn{2}{c}{$S = 64$} \\
         \cmidrule(lr){4-5}\cmidrule(lr){6-7}
         & & & NFEs/seq. ($\downarrow$) & Acc. \% ($\uparrow$) & NFEs/seq. ($\downarrow$) & Acc. \% ($\uparrow$) \\
         & & & \tiny{Mean$_{\pm\text{SD}}$} & \tiny{$0$-shot, pass@$1$} & \tiny{Mean$_{\pm\text{SD}}$} & \tiny{$0$-shot, pass@$1$} \\
         \midrule
         \multicolumn{7}{l}{\textit{B-ClockDLM}} \\
         & Standard &  & 190.1$_{\pm\text{47.3}}$ & 54.3 & 220.1$_{\pm\text{44.3}}$ & 49.1 \\
         & + Cache Grab-Conf. & \tiny($\eta=0.99$) & 134.8$_{\pm\text{37.6}}$ & 51.6 & 182.3$_{\pm\text{45.1}}$ & 48.5 \\
         &  & \tiny($\eta=0.95$) & 117.5$_{\pm\text{39.8}}$ & 53.8 & 161.8$_{\pm\text{47.0}}$ & 47.2 \\
         &  & \tiny($\eta=0.9$) & 105.1$_{\pm\text{38.3}}$ & 53.0 & 146.5$_{\pm\text{49.1}}$ & 48.0 \\
         &  & \tiny($\eta=0.85$) & 95.6$_{\pm\text{37.4}}$ & \textbf{54.7} & 135.8$_{\pm\text{49.6}}$ & 49.1 \\
         &  & \tiny($\eta=0.8$) & 86.8$_{\pm\text{36.6}}$ & 53.6 & 125.8$_{\pm\text{50.2}}$ & \textbf{49.4} \\
         &  & \tiny($\eta=0.75$) & 79.0$_{\pm\text{35.7}}$ & 51.6 & 113.1$_{\pm\text{48.1}}$ & 48.6 \\
         &  & \tiny($\eta=0.7$) & 71.1$_{\pm\text{33.7}}$ & 53.4 & 103.6$_{\pm\text{48.3}}$ & 48.4 \\
         \midrule
         \multicolumn{7}{l}{\textit{ClockDLM}} \\
         & Standard &  & 188.1$_{\pm\text{44.8}}$ & 59.6 & 219.1$_{\pm\text{43.4}}$ & \textbf{55.6} \\
         & + Cache Grab-Conf. & \tiny($\eta=0.99$) & 123.0$_{\pm\text{35.1}}$ & 57.7 & 161.5$_{\pm\text{41.5}}$ & 51.6 \\
         &  & \tiny($\eta=0.95$) & 105.6$_{\pm\text{36.0}}$ & 57.6 & 139.3$_{\pm\text{41.0}}$ & 52.7 \\
         &  & \tiny($\eta=0.9$) & 92.8$_{\pm\text{34.5}}$ & 58.3 & 123.3$_{\pm\text{40.1}}$ & 54.9 \\
         &  & \tiny($\eta=0.85$) & 82.3$_{\pm\text{32.0}}$ & 58.8 & 112.2$_{\pm\text{40.4}}$ & 55.0 \\
         &  & \tiny($\eta=0.8$) & 74.6$_{\pm\text{31.0}}$ & \textbf{60.0} & 99.9$_{\pm\text{40.0}}$ & 53.0 \\
         &  & \tiny($\eta=0.75$) & 67.1$_{\pm\text{29.8}}$ & 59.3 & 88.7$_{\pm\text{38.1}}$ & 55.0 \\
         &  & \tiny($\eta=0.7$) & 59.5$_{\pm\text{27.6}}$ & \textbf{60.0} & 78.4$_{\pm\text{35.6}}$ & 53.6 \\
        \bottomrule
    \end{tabular}
    \label{tab:gsm8k_accuracy_cache_grab_conf}
\end{table}

\subsection{Effect of SSD Training}\label{appsubsec:exp_results_ssd}
Table \ref{tab:gsm8k_ssd_ablation} isolates the effect of the joint AR and diffusion training objective used to train models that enable Cache Grab-SSD.
All models are decoded with the standard diffusion sampler, i.e., without SSD.
Joint training improves accuracy at every window size, by 0.9 to 5.7 points, while leaving NFEs/seq essentially unchanged.
At $S \in \{4, 8\}$, the jointly trained models (66.6\% and 67.2\%) even exceed the AR baseline in Table \ref{tab:gsm8k_accuracy} (64.3\%).

\begin{table}[ht]
    \centering
    \small
    \caption{Effect of joint AR and diffusion loss training.
    Zero-shot pass@$1$ on all 1,319 GSM8K test examples for Sliding Window ClockDLMs trained on TinyGSM, decoded with the standard diffusion sampler in both cases.
    Adding the AR loss leaves the denoising budget unchanged but improves accuracy at every active window size $S$.
    NFEs are means with $\pm$ sample standard deviations in subscripts.}
    \setlength{\tabcolsep}{5pt}
    \begin{tabular}{lccccc}
        \toprule
        & \multicolumn{2}{c}{Diffusion loss only} & \multicolumn{2}{c}{Joint AR $+$ diffusion loss} & \\
        \cmidrule(lr){2-3}\cmidrule(lr){4-5}
        $S$ & NFEs/seq. ($\downarrow$) & Acc. \% ($\uparrow$) & NFEs/seq. ($\downarrow$) & Acc. \% ($\uparrow$) & $\Delta$ \\
        \midrule
        4 & 161.5$_{\pm\text{46.5}}$ & 65.7 & 160.4$_{\pm\text{45.6}}$ & \textbf{66.6} & $+$0.9 \\
        8 & 165.0$_{\pm\text{45.8}}$ & 63.8 & 164.5$_{\pm\text{45.6}}$ & \textbf{67.2} & $+$3.4 \\
        16 & 172.6$_{\pm\text{44.9}}$ & 60.0 & 173.7$_{\pm\text{47.0}}$ & \textbf{63.8} & $+$3.8 \\
        32 & 188.1$_{\pm\text{44.8}}$ & 59.6 & 187.4$_{\pm\text{44.4}}$ & \textbf{63.5} & $+$3.9 \\
        64 & 219.1$_{\pm\text{43.4}}$ & 55.6 & 219.5$_{\pm\text{44.3}}$ & \textbf{61.3} & $+$5.7 \\
        \bottomrule
    \end{tabular}
    \label{tab:gsm8k_ssd_ablation}
\end{table}

\subsection{Generated Sample Examples}\label{appsubsec:exp_results_examples}
\begingroup
\newtcolorbox{generatedsample}[1]{breakable,colback=black!1,colframe=black!35,boxrule=0.4pt,arc=2pt,left=6pt,right=6pt,top=5pt,bottom=5pt,colbacktitle=black!8,coltitle=black,fonttitle=\bfseries\small,title={#1},fontupper=\fontsize{8.5}{10}\selectfont,before skip=7pt,after skip=5pt}
\setlength{\parindent}{0pt}
\setlength{\parskip}{2pt}
\captionsetup{font=small}
We present selected generations from the saved evaluation runs, using the model checkpoints and sampling procedures described above.
Examples are screened for non-explicit content; the GSM8K examples are additionally checked for correct program logic and final answers.
Generated wording is preserved, and the CNN/DailyMail comparisons use the same articles for both models.
All runs use EMA weights and base seed $1$.

\textbf{OpenWebText}~
We show complete $1{,}024$-token generations from two configurations in Figure \ref{fig:owt_mauve}: RePlaid with $T=256$, no nucleus truncation and no repetition penalty, and ClockDLM with $S=N=64$, top-$p=0.95$ and repetition penalty $1.2$.
Both use DDPM at temperature $1$ and reuse the preceding clean-embedding estimate for self-conditioning: the full posterior mean for RePlaid and the nucleus-truncated posterior mean for ClockDLM.
The respective numbers of executed model forwards are $257$ and $1{,}087$ per sequence.
\clearpage

\begin{generatedsample}{RePlaid --- OpenWebText}
\textless{}\textbar{}endoftext\textbar{}\textgreater{} few days.
Then seabirds who had died on the surface threw off the rafts and carcasses.
Seabirds started diving for food But the rushing cold water worsened the situation and released the polar chicks, as they apparently did sometimes.
The fact that it was possible in such cold waters for plankton chicks to be preserved, says Sabulis, shows that Antarctic animals were not getting along well.
In fact, the female plankton, Cepolina, itself died just after its arrival, because of freezing during the warm winter.
Later on, heavy meltwaters flooded the Columbia River on the northern coast of Antarctica, and began to wash away the chicks.
Along with them were the bodies of two clams, Dionosaurus and Vesiformes, which had hatched.
"Scientists did not think," says Sabulis, that Cepolina was made of this different mass and could therefore be used for cod.
However, with some clams that remained alive, the warmer water suggests that their deaths were caused by traffic in the river.
Other clams remain in the lake, along with many juveniles, and Sabulis says they suggest that the body from Cepolina was probably used for milking up seabirds (or river pancreas).
"Milking on such a large volume of land — consisting largely of fish," says Sabulis — "makes it even harder for salmon and rivermen to maintain a better economic standard of living than minks, humpback turtles and fish." [3]

See also [ edit ]\textless{}\textbar{}endoftext\textbar{}\textgreater{}Donald Trump — an unconventional presidential candidate — rallied a group of business people in the state's capital last week.
Though mostly miners and traditional coal industry workers, dozens of Trump supporters were ranging from mine owners to port builders and employee government workers.
In recent weeks, Trump has shown the heavy weight of business and commerce, spotlighting government regulations and thrusting job creation into a top national priority.

On the grounds of Idaho Lutheran Brothers High School Chapel in Boise, Republican candidate Donald Trump announces his intention to win the Oregon Republican nomination for president onApril 16, 2016.
(Photo: The Oregonian)

"A lot of the miners he's spoken against are men," said Greg Summer, a Trump supporter who attended a meeting here.
"They don't get any healthcare.
They get no family."

Trump's 'teaching message' puts Westminster in jeopardy

County public sector workers said they value their candidate's authenticity.

"We think he is a great voice for people, both for good working conditions and small business and applaud his insistence on giving them time off work," Rob Solomon, national director of Local 1148, a public sector union in Gadsden, Ore., said in an email.

But others on the business side of the stump expressed a greater acceptance of Trump's message.
Both decent working conditions and small businesses are important, said Jim Frankel, the owner of a brick-and-mortar outdoor store in Lincoln Village.
He also serves as a vice chairman of the union-backed group Coalition for American Jobs and acknowledges that those issues "belong to some of the small business owners" who support Trump.
But good working conditions remain, he said, "and certainly we accept his message."

Critics see huge risks in a Trump battleground economy, especially compared with the businessman outlining explicitly his plan to expand job creation.
Frank Tucker, an adviser and consultant at BRCA Consulting LLC, a coal-industry watchdog, has criticized that approach as generally unswerving.
Trump's first jobs plan shows a thrashy approach and could be a significant slog, he said.

"The coal recovery is arguably more important not only because of his small business investments, but in part because of his repudiation of core conservative policies," Tucker said.
"Trump is arguably the most popular union and small business candidate in the race and has consistently proposed developing infrastructure projects that would lock in any permanent coal jobs."

Other supporters argue that politics is about big business.
Jeff Martinez, a Trump supporter in Glenport, Wash., said he has supported coal policies championed by Hillary Clinton and wants to see her "doing this our own way." When he asked if Democratic nominee Clinton supported coal workers in communities such as Rampunda Range, Ramos said, "Hillary is very big on personal money."

County business owners say Trump is the only candidate for president emphasizing conservative government rules that can actually spur job creation.

Why Americans most needed backing

At the Trump rally, miners from King counties, Spokane, Wash., Vernon and others in Oregon were waving small signs and posters related to their work.
Some were unions, wearing a tinted coat and a painted hat on a baseball bat.
Some had been inspired by "Manchita Run," the popular calendar from 1900s in Oregon.

Last year, when labor activists in Boise urged the Idaho state legislature to strike down a law on miners' rights, a Walmart store in Mont\textless{}\textbar{}endoftext\textbar{}\textgreater{}

\end{generatedsample}

\captionof{figure}{Full OpenWebText generation from RePlaid, saved sequence index $0$, without nucleus truncation or a repetition penalty.}
\label{fig:sample_owt_replaid_0}
\clearpage

\begin{generatedsample}{ClockDLM --- OpenWebText}
\textless{}\textbar{}endoftext\textbar{}\textgreater{} round up employees who worked abroad to write the rules.

The rules he quietly signed were for these in 2010, under the tweaking of the Competition Act of 2005, a law for companies that deal with international trade that went into one constitutional term in 2007 and was repealed in 2014.

The enthusiasm for limits on the vulnerable often reminds me of the time you look for a train from the opposite end, leaving the station in time to witness the aspirations of your mates, friend your parents, examine the new gear, focus your eyes on the approaching fireworks.
When someone wants to embrace the challenge of the job, nothing about these rules will do.\textless{}\textbar{}endoftext\textbar{}\textgreater{}This was a dream of mine, Erica’s queue broadcaster that can broadcast messages to a channel or a console, and can push notifications whenever necessary.
Sorry if you missed this and see below:

So, let’s continue down this road!

The Objective

I created this “Condition Scaling Script”.
All it has to do is tell you you may have the value processing defined, and you can type in the message or scroll it with your finger.

For example, you might type this:

yourMessage.hasValue(0, 0);

Which is not only to your message, but your Messenger…

What’s Next?

In this post, I’ll be explaining the concepts of interface and access in a nutshell by telling you what I managed.
If your knowledge becomes too limited, it might be time to dive deeper and explore some tips that to make the path easier.

To Get You Started With It

You need to set document.keyText.hasValue :

class Window \{ string idText; string prevText; window.setIgnoringText(idText); \}

When your Window gets too big, like this one, prevText will take this string, and then try to copy the contents you read onto the current session.

And you should also set an prevText value of zero, so the clipboard can only hold the text it currently has.

Knowing this allows you to use it globally.
Just like when you click the Bookmarks button everywhere you go, you can easily get the list of all messages.
That’s similar to the waterfall on the splash screen, and you can easily create a variety of different kinds of events to override displaying the content if needed.

The Internet Explorer is burdened with a screen view option.
This allows you to fearlessly know where your messages come from in more than one screen view, and then not geolocate your friends.
Great.
Right now, the program can replay itself 2 million words at a time, and to date it has been broadcasting messages in multiple screen views, as one action and for multiple users, at the same time window it is reading.

When your Window is too big, it will copy all messages in the Internet Explorer Frame, which causes it globally.

How to Use It

Since the program returns the list of messages consumed by the user, it can change the screen top display width to its own depurion.
And you can use a method like store() , which copies the messages emitted without being viewed.

More beautiful crummy interface.

Conclusion

I got beautiful results sticking to the “Condition Scaling” script alone, all the way until I finally executed it, a bunch of data from many people come and went, and hours spent on calls, processing messages across many different mouse sessions.
And finally, once it’s finished, everyone has had a blast using it.

This is definitely a milestone for lots of other designers, who need to put together a useful program, and have feature-rich interactions that they can’t seem to afford, let alone writing a beautiful piece of code.
And I hope that many of you will love that.\textless{}\textbar{}endoftext\textbar{}\textgreater{}DemTV Live 05/05/2017 (Update 8)

Live Online Android streaming service has a third major update that has been recently pushed to DemTV Live (Update 4).
This update covers several new features.
The main part of the update is the ability to stream streams of any seasons via the official Netflix or Google Play TV sites.
The purpose of the streaming builds upon the current DemTV Live website, is to allow viewers to watch their most-anticipated seasons in 4K HD.
Here is from the description of the update:

“High Quality™ Streaming (Premium Premium) is designed to deliver gorgeous photos and an affordable way to experience almost anything.
The new High Frame Streaming (HDDR™) interface can accommodate up to 1 gigabyte of raw or native video.
Additional frame sizes provide extra Turbo HD Memory capabilities sufficient to handle up to 6 gigabytes in native video to deliver 100\% high quality and immersive video.
This change builds to DemTV Live’s future strategy of high image quality and epic video delivery\textless{}\textbar{}endoftext\textbar{}\textgreater{}

\end{generatedsample}

\captionof{figure}{Full OpenWebText generation from ClockDLM, saved sequence index $1$, with top-$p=0.95$ and repetition penalty $1.2$.}
\label{fig:sample_owt_clockdlm_1}
\clearpage





\textbf{GSM8K}~
These ClockDLM examples use $S=N=32$, DDPM with top-$1$ clean-token predictions and token-embedding self-conditioning, and the $250{,}000$-step EMA checkpoint.
The displayed results are those recorded by the evaluation program.

\begin{generatedsample}{ClockDLM --- GSM8K ($S=N=32$, $198$ NFEs)}
\textbf{Question}\par
Janet’s ducks lay 16 eggs per day.
She eats three for breakfast every morning and bakes muffins for her friends every day with four.
She sells the remainder at the farmers' market daily for \$2 per fresh duck egg.
How much in dollars does she make every day at the farmers' market?

\medskip
\textbf{Generated program}\par
\begin{Verbatim}
def simple_math_problem() -> int:
    """
    Janet’s ducks lay 16 eggs per day.
    She eats three for breakfast every morning and bakes muffins for her friends every day with four.
    She sells the remainder at the farmers' market daily for $2 per fresh duck egg.
    How much in dollars does she make every day at the farmers' market?
    """
    total_eggs = 16
    breakfast_eggs = 3
    muffin_eggs = 4
    fresh_eggs = total_eggs - breakfast_eggs - muffin_eggs
    price_per_egg = 2
    total_revenue = fresh_eggs * price_per_egg
    result = total_revenue

    return result
\end{Verbatim}
\medskip
\textbf{Recorded execution result:} $18$.
\textbf{Reference answer:} $18$.
\end{generatedsample}

\captionof{figure}{ClockDLM generation for GSM8K test example $0$.
The program correctly computes $(16-3-4)\times 2=18$.}
\label{fig:sample_gsm8k_0}

\begin{generatedsample}{ClockDLM --- GSM8K ($S=N=32$, $118$ NFEs)}
\textbf{Question}\par
A robe takes 2 bolts of blue fiber and half that much white fiber.
How many bolts in total does it take?

\medskip
\textbf{Generated program}\par
\begin{Verbatim}
def simple_math_problem() -> int:
    '''
    A robe takes 2 bolts of blue fiber and half that much white fiber.
    blocking How many bolts in total does it take?
    '''
    blue_fiber = 2
    white_fiber = 0.5 * blue_fiber
    total_bolts = blue_fiber + white_fiber
    result = total_bolts
    return result
\end{Verbatim}
\medskip
\textbf{Recorded execution result:} $3$.
\textbf{Reference answer:} $3$.
\end{generatedsample}

\captionof{figure}{ClockDLM generation for GSM8K test example $1$.
The program correctly computes $2+0.5\times 2=3$.}
\label{fig:sample_gsm8k_1}

\clearpage

\textbf{CNN/DailyMail}~
RePlaid uses $T=1024$ and ClockDLM uses $S=N=32$, both with repetition penalty $1.2$ and the remaining decoding settings in Appendix \ref{appsubsec:exp_setup_cnndm}.
We show an excerpt of the article input, followed by the reference and both complete generated summaries.

\begin{generatedsample}{CNN/DailyMail --- Large Hadron Collider}
\textbf{Article excerpt}\par
(CNN)The world's biggest and most powerful physics experiment is taking place as you read this.
The Large Hadron Collider (LHC), a particle accelerator and the largest machine in the world, is ready for action following a two-year shutdown.
After problems that delayed the restart in March, scientists at the European Organization for Nuclear Research (CERN) completed final tests, enabling the first beams to start circulating Sunday inside the LHC's 17 mile (27 km) ring. \ldots

\medskip
\textbf{Reference summary}\par
The Large Hadron Collider (LHC) begins again after a two-year shutdown .\par The restart was delayed in March .

\medskip
\textbf{RePlaid ($T=1024$, $1{,}025$ NFEs)}\par
Up to 600 million particles per second at the Large Hadron Collider ready for action .\par The device was join for 10 hours, traveling more than 6 billion miles (more than 10 billion km)\par LHC is the largest device in 1975, in 2001, 2008, 2001 and 2001.
level

\medskip
\textbf{ClockDLM ($S=N=32$, $121$ NFEs)}\par
the U.S.
particle accelerator is ready for action following a two-year shutdown .\par A proton in the LHC makes 11,245 circuits per second .\par The purpose is to recreate the conditions that existed moments after the "Big Bang"\par But some still say it's not complete, but doesn't know that the universe is only matter .\par The project has been developed to replicate the origin of the universe based on the particle system-like
\end{generatedsample}

\captionof{figure}{Paired summaries for CNN/DailyMail test example $32$.
Both models receive the same article input, and their generated summaries are shown without corrections.}
\label{fig:sample_cnndm_32}

\medskip

\begin{generatedsample}{CNN/DailyMail --- Lunar eclipse}
\textbf{Article excerpt}\par
(CNN)Sky watchers in western North America are in for a treat: a nearly five-minute total lunar eclipse this morning.
Here's how it's unfolding: .
It started at 3:16 a.m.
Pacific Daylight Time, when the moon began moving into Earth's shadow.
For the next hour and 45 minutes, that shadow will move across the moon and engulf it at 4:58 a.m.
Pacific Time.
The total eclipse will only last four minutes and 43 seconds, and NASA says that makes it the shortest one of the century. \ldots

\medskip
\textbf{Reference summary}\par
The total eclipse will only last 4 minutes and 43 seconds .\par People west of the Mississippi River will have the best view .\par Parts of South America, India, China and Russia also will see the eclipse .

\medskip
\textbf{RePlaid ($T=1024$, $1{,}025$ NFEs)}\par
The lunar eclipse will only last four minutes and 43 seconds, when it was 3:16 a.m.\par Break is third in a series of four in a row, known as a "Arrad"\par The first was on April 15, 2014 .\par The second was in September 2014, the next is Saturday and there will be one more, on September 28 .

\medskip
\textbf{ClockDLM ($S=N=32$, $119$ NFEs)}\par
The eclipse will take for four minutes and 43 seconds .\par It is the shortest one of the century, but sunrise will see the East Coast .\par The first was on April 15, 2014, and the second was on September 28 .\par The moon that you're up may not be good, but it's dark we can't have seen it, but it seems like it will give us a
\end{generatedsample}

\captionof{figure}{Paired summaries for CNN/DailyMail test example $34$.
Both models receive the same article input, and their generated summaries are shown without corrections.}
\label{fig:sample_cnndm_34}

\clearpage

\endgroup

\section{Assets}\label{appsec:assets}
Table~\ref{tab:assets_data} lists datasets used in this work.
Table~\ref{tab:assets_models} lists tokenizers, model architectures, pre-trained models, and codebases used in this work.
Table~\ref{tab:assets_software} lists relevant software packages used in this work.
For each asset, we detail its corresponding license.

\begin{table}[ht]
    \centering
    \small
    \caption{Datasets and corresponding licenses.
    For OpenWebText, CC0 applies to the dataset packaging; source articles retain their original rights.}
    \setlength{\tabcolsep}{5pt}
    \begin{tabular}{@{}llp{0.42\textwidth}@{}}
        \toprule
        Dataset & License & URL \\
        \midrule
        OpenWebText {\scriptsize \citep{Gokaslan2019OpenWeb}} & CC0 1.0 & {\scriptsize \url{https://huggingface.co/datasets/Skylion007/openwebtext}} \\
        TinyGSM {\scriptsize \citep{liu2023tinygsm}} & MIT & {\scriptsize \url{https://huggingface.co/datasets/TinyGSM/TinyGSM}} \\
        GSM8K {\scriptsize \citep{cobbe2021training}} & MIT & {\scriptsize \url{https://github.com/openai/grade-school-math}} \\
        CNN/DailyMail {\scriptsize \citep{DBLP:conf/nips/HermannKGEKSB15, see-etal-2017-get}} & Apache 2.0 & {\scriptsize \url{https://huggingface.co/datasets/abisee/cnn_dailymail}} \\
        \bottomrule
    \end{tabular}
    \label{tab:assets_data}
\end{table}

\begin{table}[ht]
    \centering
    \small
    \caption{Tokenizers, architecture definitions, external code, and their corresponding licenses.}
    \setlength{\tabcolsep}{4pt}
    \begin{tabular}{@{}lllp{0.33\textwidth}@{}}
        \toprule
        Asset & Use & License & URL \\
        \midrule
        GPT-2 \scriptsize\citep{radford2019language} & Tokenizer & MIT & {\scriptsize \url{https://huggingface.co/openai-community/gpt2}} \\
        SmolLM-135M \scriptsize\citep{allal2024SmolLM} & Tokenizer & Apache 2.0 & {\scriptsize \url{https://huggingface.co/HuggingFaceTB/SmolLM-135M}} \\
        Qwen3-0.6B-Base \scriptsize\citep{qwen3technicalreport} & Tokenizer, architecture & Apache 2.0 & {\scriptsize \url{https://huggingface.co/Qwen/Qwen3-0.6B-Base}} \\
        ModernBERT \scriptsize\citep{warner2025smarter} & MAUVE evaluation & Apache 2.0 & \scriptsize{\url{https://huggingface.co/answerdotai/ModernBERT-large}} \\
        $\mathbb{S}$-FLM \scriptsize\citep{deschenaux2026language} & Baseline models, code & Apache 2.0 & \scriptsize\url{https://huggingface.co/jdeschena/s-flm} \\
        FMLM$+$ \scriptsize\citep{agarwal2026posterior} & Baseline models, code & Apache 2.0 & \scriptsize\url{https://github.com/MananAg007/posterior-refinement} \\
        \bottomrule
    \end{tabular}
    \label{tab:assets_models}
\end{table}

\begin{table}[ht]
    \centering
    \small
    \caption{Software and corresponding licenses.}
    \setlength{\tabcolsep}{6pt}
    \begin{tabular}{@{}ll@{}}
        \toprule
        Library & License \\
        \midrule
        HuggingFace {\scriptsize \citep{wolf2019huggingface}} & Apache 2.0 \\
        Hydra {\scriptsize \citep{Yadan2019Hydra}} & MIT \\
        Language Model Evaluation Harness {\scriptsize \citep{gao2021framework}} & MIT \\
        Matplotlib {\scriptsize \citep{Hunter:2007}} & Matplotlib License \\
        MAUVE {\scriptsize \citep{mauve_github}} & GNU General Public License, Version 3 \\
        NumPy {\scriptsize \citep{harris2020array}} & BSD 3-Clause \\
        OmegaConf & BSD 3-Clause \\
        Pandas {\scriptsize \citep{reback2020pandas}} & BSD 3-Clause \\
        PyTorch {\scriptsize \citep{Paszke_PyTorch_An_Imperative_2019}} & BSD 3-Clause \\
        PyTorch Lightning {\scriptsize \citep{Falcon_PyTorch_Lightning_2019}} & Apache 2.0 \\
        TorchMetrics & Apache 2.0 \\
        \bottomrule
    \end{tabular}
    \label{tab:assets_software}
\end{table}

\end{document}